\documentclass{article}

\usepackage{placeins}
\usepackage{microtype}
\usepackage{graphicx}
\usepackage{subcaption}
\usepackage{booktabs} 
\usepackage{xspace}
\usepackage{xcolor}
\usepackage{dblfloatfix}
\usepackage{hyperref}
\usepackage[table]{xcolor}

\usepackage[accepted]{icml2026}

\usepackage{amsmath}
\usepackage{amssymb}
\usepackage{mathtools}
\usepackage{amsthm}

\usepackage{multirow}
\usepackage[table,xcdraw]{xcolor}
\usepackage{enumitem}
\usepackage{float}

\usepackage[capitalize,noabbrev]{cleveref}

\definecolor{Gray}{gray}{0.85}
\definecolor{Yellow}{rgb}{0.5,0.5,0.5}
\definecolor{ForestGreen}{rgb}{0.13,0.55,0.13}

\newcommand{\permute}{\textsc{Permute}\xspace}
\newcommand{\git}{\textsc{Git-Rebasin}\xspace}
\newcommand{\wa}{\textsc{WAvg.}\xspace}
\newcommand{\name}{\textsc{GnnMerge}\xspace}
\newcommand{\nameff}{\textsc{GffMerge}\xspace}
\newcommand{\surgery}{\textsc{Surgery}\xspace}
\newcommand{\zipit}{\textsc{ZipIt!}\xspace}

\newcommand{\gnns}{\textsc{Gnn}s\xspace}
\newcommand{\gnn}{\textsc{Gnn}\xspace}
\newcommand{\node}{\textsc{NodeFormer}\xspace}
\newcommand{\gat}{\textsc{Gat}\xspace}
\newcommand{\sage}{\textsc{GraphSage}\xspace}
\newcommand{\gcn}{\textsc{Gcn}\xspace}
\newcommand{\gin}{\textsc{Gin}\xspace}

\newcommand{\CG}{\mathcal{G}\xspace}

\newcommand{\CY}{\mathcal{Y}\xspace}

\newcommand{\CN}{\mathcal{N}\xspace}

\newcommand{\CV}{\mathcal{V}\xspace}
\newcommand{\CE}{\mathcal{E}\xspace}

\newcommand{\X}{\boldsymbol{X}\xspace}

\newcommand{\CZ}{\mathbf{Z}\xspace}
\newcommand{\CS}{\mathcal{S}\xspace}

\newcommand{\cW}{\mathbf{W}\xspace}

\newcommand{\cG}{\mathbf{G}\xspace}
\newcommand{\CL}{\mathcal{L}\xspace}
\newcommand{\cx}{\mathbf{x}\xspace}

\newcommand{\cg}{\mathbf{g}\xspace}

\newcommand{\ch}{\mathbf{h}\xspace}

\newcommand{\cz}{\mathbf{z}\xspace}

\newif\ifproofmode
\proofmodetrue 
\ifproofmode
    
\else
    
\fi

\setlist{nolistsep,leftmargin=*}

\newtheorem{defn}{\textbf{Definition}}

\newtheorem{prob}{\textbf{Problem}}

\definecolor{babypink}{rgb}{0.96, 0.76, 0.76}
\definecolor{top1}{HTML}{a5dc82}
\definecolor{top2}{HTML}{dff3d9}

\definecolor{c1}{HTML}{d5e8d4}
\definecolor{c1_1}{HTML}{82b366}
\definecolor{c2}{HTML}{ffe6cc}
\definecolor{c2_1}{HTML}{d79b01}
\definecolor{c3}{HTML}{dae8fc}
\definecolor{c3_1}{HTML}{6c8ebf}
\definecolor{text_grey}{HTML}{5e5e5e}

\newenvironment{review}
    {\color{black}} 

\icmltitlerunning{\nameff: Efficient Merging of Graph Neural Force Fields and Beyond}

\begin{document}

\twocolumn[
\icmltitle{\nameff: Efficient Merging of Graph Neural Force Fields and Beyond}

\icmlsetsymbol{equal}{*}

\begin{review}
\begin{icmlauthorlist}
\icmlauthor{Parth Verma}{cse}
\icmlauthor{Parv P. Singh}{ai}
\icmlauthor{Vipul Garg}{cse}
\icmlauthor{Ishita Thakre}{cse}
\icmlauthor{N. M. Anoop Krishnan}{ai,civil}
\icmlauthor{Sayan Ranu}{cse,ai}
\end{icmlauthorlist}

\icmlaffiliation{cse}{Department of Computer Science, Indian Institute of Technology Delhi, Hauz Khas, New Delhi, India}
\icmlaffiliation{ai}{Yardi School of Artificial Intelligence, Indian Institute of Technology Delhi, Hauz Khas, New Delhi, India}
\icmlaffiliation{civil}{Department of Civil Engineering, Indian Institute of Technology Delhi, Hauz Khas, New Delhi, India}

\icmlcorrespondingauthor{Sayan Ranu}{sayanranu@iitd.ac.in}
\end{review}

\icmlkeywords{Graph Neural Networks, Model Merging, Neural Force Fields, Molecular Dynamics}

\vskip 0.3in
]


\printAffiliationsAndNotice{}  

\begin{abstract}

Graph Neural Networks (\gnns) have revolutionized Neural Force Fields for atomistic simulations, achieving near-quantum accuracy at reduced cost, yet adapting these models to new chemical systems requires expensive retraining of foundation models. Inspired by model merging in vision and language processing, we introduce \nameff, the first principled framework for closed-form model merging in \gnn force fields, and \gnns in general. We exploit the linear structure of message-passing layers and formulate merging as a convex embedding-alignment problem with an analytical solution. Through the first systematic benchmarking of model merging for \gnn force fields, we show that existing methods designed for vision and language catastrophically fail on force field regression, while \nameff recovers performance approaching gold standard joint training. Across molecular (MD17, MD22), solid-state (LiPS20), and large-scale graph benchmarks, \nameff and \name (its generic \gnn counterpart) achieve 5-27$\times$ speedups while enabling modular composition of specialized models. Remarkably, our closed-form solution alone outperforms all baseline methods before fine-tuning and provides superior initialization for faster, data-efficient convergence.
\looseness=-1
\end{abstract}

\section{Introduction}
\label{sec:intro}
Graph Neural Network (\gnn) based force fields have become a central tool for modeling interatomic interactions with near-quantum accuracy while remaining computationally tractable~\cite{unke2021machine, batzner20223, Batatia2022mace,batatia_foundation_2025}. By operating directly on atomistic graphs, these models learn mappings from atomic configurations to potential energies and forces, enabling molecular dynamics, structure relaxation, and large-scale materials screening. Over the past few years, \gnn force fields have demonstrated strong empirical performance across chemistry and materials science benchmarks, and now underpin applications ranging from drug discovery and catalysis to battery materials and solid-state physics~\cite{Owen2024, Zhang2025, Park2021,stridernet}.

Recent progress has been driven by the emergence of large, general-purpose \emph{foundation models} for atomistic systems. Models such as M3GNet~\cite{Chen2022}, SevenNet~\cite{park_scalable_2024}, MACE~\cite{batatia_foundation_2025} and Orb~\cite{neumann2024Orbfastscalableneural} are trained on massive and diverse datasets, comprise millions of parameters, and consistently achieve state-of-the-art performance across broad chemical spaces~\cite{Riebesell2025}. To adapt these foundation models to specific applications, practitioners fine-tune them on targeted chemical systems, molecular families, or materials. While this strategy yields highly accurate specialized models, it creates a critical practical challenge: \emph{how can knowledge from multiple fine-tuned models be efficiently consolidated without expensive retraining?}

While foundation models have raised the performance ceiling, they have also introduced significant computational and practical bottlenecks. Fine-tuning large \gnn force fields demands substantial GPU resources and extended training times, driven by their million-scale parameter counts and the high computational cost of message passing on atomistic graphs. Moreover, in many practical settings, the specialized datasets used for fine-tuning evolve over time as new molecules, configurations, or chemical regimes are added, turning this already expensive procedure into a recurring and costly operation. Consequently, adapting a foundation model to each new molecular system or chemical family is often prohibitively expensive.

This motivates a natural and practically important question. Suppose one already has high-quality \gnn force fields fine-tuned separately on molecular systems $X$ and $Y$. If the goal is to obtain a force field specialized for the combined system $X \cup Y$, must one fine-tune a large foundation model again via joint training on the merged dataset, or can the existing fine-tuned models be \emph{merged} directly into a single model? Current standard practice relies on joint training over the combined dataset, which, while effective, is computationally expensive and often redundant. In contrast, model merging offers the promise of reusing fine-tuned force fields as modular \textit{building blocks}, enabling scalable specialization while significantly reducing training cost, energy consumption, and time-to-deployment.

\subsection{Related works and open challenges for \gnns.}
Model merging has recently emerged as a distinct research direction that studies how independently trained neural networks can be combined without access to training data~\cite{stoica2024zipit,ainsworth2023gitrebasin, yang2024adamerging,ilharco2023editingtaskarithmetic,emr-merging-DBLP:journals/corr/abs-2405-17461,lu2024twinmerging}. These methods typically rely on weight averaging, task arithmetic, permutation alignment, or lightweight optimization to reconcile differences across models. Despite their success in vision and language domains, these techniques have not been extended to \gnns in a principled manner.

The gap is even more pronounced for \gnn force fields. Unlike vision or language models, which are typically trained for discrete prediction tasks such as classification or next-token generation, \gnn force fields operate in a continuous regression regime, predicting energies and forces over high-dimensional atomic configurational spaces~\citep{https://doi.org/10.48550/arxiv.2207.09453,torrens_interatomic_2012, jacobs_practical_2025}. These predictions must remain numerically stable and physically consistent, with forces tightly coupled to energies through shared parameters~\citep{neumann2024Orbfastscalableneural,equiformer_v2} or differentiation~\citep{Batatia2022mace, batzner20223,Chen2022, park_scalable_2024}. As a result, small variation in internal representations can be amplified through force gradients, leading to large errors that compound over long molecular dynamics trajectories. Moreover, force fields must respect locality, smoothness, and transferability across atomic environments, properties that are not explicitly enforced by model merging objectives developed for discrete domains. Consequently, naive parameter merging is particularly brittle in this setting, which we will demonstrate in our empirical evaluation. Thus, \emph{model merging for \gnns remains an open problem, let alone for \gnn force fields}.

\subsection{Contributions}
In this work, we introduce \nameff, a principled framework for efficiently merging Graph Neural Network force fields. \nameff combines \textit{closed-form parameter merging} with lightweight, targeted adaptation to recover the performance of joint training at a fraction of the computational cost. Our contributions are summarized as follows:
\begin{itemize}
    \item \textbf{Analytical merging of \gnn force fields:} We identify linear parameter blocks within widely used \gnn force field architectures and derive a closed-form merging objective that minimizes activation mismatch with respect to the source models, eliminating the need for expensive gradient-based optimization.
    \item \textbf{Efficient recovery of joint-training performance:} We present the first benchmarking study for model merging in \gnns and empirically establish that current state-of-the-art methods are ineffective on \gnns. We demonstrate that sparse fine-tuning of only late interaction layers and force heads is sufficient to close the remaining performance gap, allowing \nameff to match joint-training accuracy of gold-standard train-from-scratch version, while achieving $5\times$ to $28\times$ training speedups.
    \item \textbf{Beyond \gnn force fields:} The proposed architecture is generic enough to accommodate generic \gnns such as \gat~\citep{velickovic2018graph}, \gcn~\citep{kipf2017semi-gcn}, \gin~\citep{xu2018how} and \sage~\citep{graphsage}. We demonstrate this generic version, referred to as \name, in Sec.~\ref{sec:gnnmerge}. The detailed benchmarking study establish that \name delivers superior accuracy to generic model merging baselines and achieves up to $\mathbf{1000\times}$ speed-up over retraining.
\end{itemize}

\section{Problem Formulation}
\label{sec:problem}
\begin{defn}[Atomistic Graph]
\textit{Let $\CG = (\CV, \CE, \X, \mathbf{r})$ denote an atomistic graph representing a molecular or material system. Here, $\CV$ denotes the set of atoms, $\CE \subseteq \CV \times \CV$ denotes interatomic edges defined by a neighborhood criterion (e.g., distance cutoff), $\X \in \mathbb{R}^{|\CV| \times d}$ denotes node attributes such as atomic species or types encoded as $d$-dimensional feature vectors, and $\mathbf{r} \in \mathbb{R}^{|\CV| \times 3}$ denotes the Cartesian coordinates of the atoms. The feature vector for atom $v \in \CV$ is denoted by $\cx_v$.}
\end{defn}

\gnn force fields operate on atomistic graphs to predict continuous physical quantities, most commonly the total potential energy of the system and the atomic forces. Unlike discrete graph prediction tasks such as node classification or link prediction, force field learning is a continuous regression problem defined over high-dimensional configuration spaces. We formalize the learning task as follows:
\begin{defn}[Learning a force field]
\textit{Let $\langle \mathbb{T}, \CY \rangle$ be a training dataset, where each element of $\mathbb{T}$ is an atomistic graph $\CG$ and $\CY = \{E, \mathbf{F}\}$ contains the corresponding ground-truth total energy $E \in \mathbb{R}$ and atomic forces $\mathbf{F} \in \mathbb{R}^{|\CV| \times 3}$. A \gnn force field with parameters $\Theta$ is trained to minimize a regression loss}
\vspace{-0.05in}
\begin{equation}
\CL(\CY, \Theta(\CG)) = \lambda_E \|E - \hat{E}\|_2^2 + \lambda_F \|\mathbf{F} - \hat{\mathbf{F}}\|_2^2,
\end{equation}
\textit{where $(\hat{E}, \hat{\mathbf{F}}) = \Theta(\CG)$ denote the predicted energy and forces, and $\lambda_E, \lambda_F$ balance their relative contributions.}
\end{defn}

In \gnn force fields, forces are obtained either directly as model outputs~\cite{neumann2024Orbfastscalableneural}) or implicitly as gradients of the predicted energy with respect to atomic positions~\cite{Batatia2022mace}. As a result, the learning objective tightly couples energy and force predictions and imposes strong smoothness and stability requirements on the learned representations.

We now define the problem of merging \gnn force fields.
\begin{prob}[Force Field Model Merging]\label{prob:merge-ff}
\textit{Given $n$ independently trained \gnn force field models $\Theta_1, \Theta_2, \dots, \Theta_n$, each specialized on a domain $\mathcal{D}_t$, the goal of model merging is to construct a single force field $\Theta_M$. For any input graph $\CG \in \mathcal{D}_{\text{union}}$, the predictions of $\Theta_M$ should closely match those of the expert model $\Theta_t$ corresponding to the domain of $\CG$. This is formulated as minimizing:}
\vspace{-0.05in}
\begin{equation}
\label{eq:mergeloss-ff}
\sum_{t=1}^n \mathbb{I}(\CG \in \mathcal{D}_t) \cdot \CL_t\big(\Theta_t(\CG), \Theta_M(\CG)\big),
\end{equation}
\textit{where $\mathbb{I}(\cdot)$ is the indicator function which is 1 if $\CG$ belongs to the domain $\mathcal{D}_t$ and 0 otherwise, and $\CL_t$ measures the predictive discrepancy (energy and forces) relative to the expert $\Theta_t$.}
\end{prob}

\comment{
In addition to minimizing the objective in Problem~\ref{prob:merge-ff}, effective force field model merging must satisfy the following desiderata:

\begin{itemize}
    \item \textbf{Computational efficiency:} The merging procedure should be substantially faster than retraining or jointly fine-tuning a force field on the union of all source datasets.
    \item \textbf{Model size preservation:} The merged force field $\Theta_M$ should not exceed the parameter count of the largest source model.
    \item \textbf{Physical fidelity:} The merged model should preserve the numerical stability and accuracy of energy and force predictions required for downstream molecular simulations.
\end{itemize}

We examine the problem of merging graph neural network potentials \textbf{fine-tuned from a common pre-trained backbone} on disjoint chemical domains.

\textbf{Definition 1 (Source Models).} Let $\{\mathcal{M}_1, \dots, \mathcal{M}_K\}$ be a set of $K$ \gnn force fields, referred to as \textit{source models}. We assume all models share a common architecture and initialization $\theta_{\text{init}}$ (e.g., a foundation model), but have been independently fine-tuned on distinct datasets $\mathcal{D}_k = \{(G_i, y_i)\}$, where $G_i$ denotes a molecular graph and $y_i$ denotes the ground-truth energy and forces.

\textbf{Definition 2 (Merging Task).} Our objective is to synthesize a unified model $\mathcal{M}^*$ with parameters $\theta^*$ such that it minimizes the prediction error on the union of all domains $\mathcal{D}_{\text{union}} \triangleq \bigcup_{k=1}^K \mathcal{D}_k$. The merging function $\Phi$ must satisfy:
\begin{equation}
    \theta^* = \Phi(\theta_1, \dots, \theta_K; \mathcal{D}_{\text{cal}})
\end{equation}
where $\mathcal{D}_{\text{cal}}$ is a set of calibration data utilized for aligning features. Crucially, we assume we cannot perform full joint training on $\mathcal{D}_{\text{union}}$ due to computational constraints.
}
\begin{figure}[t]
    \centering
\includegraphics[width=1.03\linewidth]{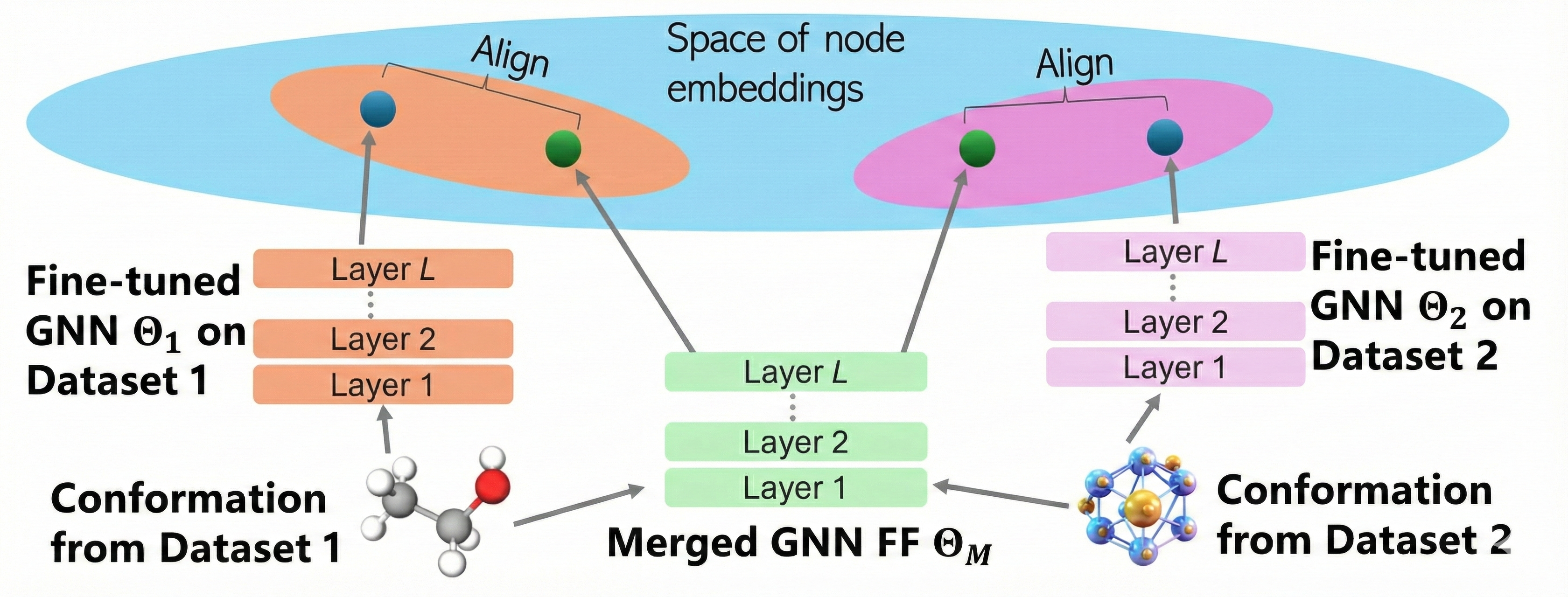}
    \caption{A visual depiction of the alignment objective in \nameff. The orange and purple ellipses represent the regions where the source \gnns $\Theta_1$ and $\Theta_2$ produce accurate predictions on Datasets 1 and 2 respectively. \nameff aims to learn a merged model $\Theta_M$ that embeds nodes closer to their \textit{original} embeddings, and thereby increasing the likelihood that the new embeddings fall within the ellipses.}
    \label{fig:objective}
    
\end{figure}

\section{\nameff: Proposed Methodology}
\label{sec:method}
\nameff leverages the insight that \gnn layers universally compute node embeddings, regardless of the specific task. In \gnn force fields, predicted energies and forces are deterministic, differentiable functions of these node embeddings and the underlying atomic geometry. Therefore, if the merged model can replicate the node embeddings generated by each base model, it can also replicate their outputs. To achieve this, we first define an optimization objective focused on preserving the node embeddings from the base models within the merged model. This objective is then relaxed to facilitate an analytical solution and enable various computational optimizations. 

\nameff proceeds in two stages. First, we derive a closed-form solution for merging the linear parameter blocks of independently fine-tuned force fields by explicitly matching their intermediate node embeddings, yielding a merged model that already lies in a favorable region of the weight space. Second, we perform targeted fine-tuning of only the final interaction layers and force heads for a small number of epochs. This lightweight refinement step corrects residual discrepancies introduced by merging, allowing the merged model to recover performance close to that of full joint training, while avoiding the cost of end-to-end optimization. The following subsections detail these steps.
\comment{
\subsection{Preliminaries and Notation}
We consider Message Passing Neural Networks (MPNNs) where the update rule for a node $v$ at layer $l$ can be generally abstracted as a linear projection of aggregated features followed by a non-linearity:
\begin{equation}
    h_v^{(l+1)} = \sigma \left( W^{(l)} \cdot \text{AGG}(\{h_u^{(l)} : u \in \mathcal{N}(v)\}) \right)
\end{equation}
Here, $h_v^{(l)} \in \mathbb{R}^d$ is the node feature vector, $W^{(l)} \in \mathbb{R}^{d \times d}$ is the learnable weight matrix, and $\mathcal{N}(v)$ denotes the neighborhood of $v$.
}

\subsection{Computation Framework of \gnn Force Fields}
\label{sec:gnnff}
\gnn-FFs compute node (atom) embeddings through a sequence of interaction layers that iteratively propagate information across the molecular graph while incorporating geometric features derived from interatomic distances. The $0^{\text{th}}$-layer embedding of atom $v \in \CV$ is initialized as $\mathbf{h}_v^0 = \cx_v$. At interaction layer $\ell$, each atom $v$ receives messages from its neighbors $\CN_v = \{u \in \CV \mid (u,v) \in \CE\}$. 
\vspace{-0.05in}
\begin{equation}
\label{eq:ff_message}
\mathbf{m}_{u \rightarrow v}^{\ell}
= \textsc{Msg}^{\ell}\!\left(\mathbf{h}_u^{\ell-1}, \, \phi(\mathbf{r}_{uv})\right),
\end{equation}
where $\mathbf{r}_{uv} = \mathbf{r}_u - \mathbf{r}_v$ and $\phi(\cdot)$ denotes a geometric encoding (e.g., distance-based radial features). The incoming messages are aggregated to form
\vspace{-0.05in}
\begin{equation}
\label{eq:ff_aggregate}
\mathbf{m}_v^{\ell}
= \textsc{Aggregate}^{\ell}\!\left(\{\!\!\{\mathbf{m}_{u \rightarrow v}^{\ell} \mid u \in \CN_v\}\!\!\}\right),
\end{equation}
where $\{\!\!\{\cdot\}\!\!\}$ denotes a multiset. The atom embedding is then updated as:
\vspace{-0.05in}
\begin{equation}
\label{eq:ff_update}
\mathbf{h}_v^{\ell}
= \textsc{Update}^{\ell}\!\left(\mathbf{h}_v^{\ell-1}, \mathbf{m}_v^{\ell}\right),
\end{equation}
where $\textsc{Update}^{\ell}$ typically consists of linear transformations followed by nonlinearities and normalization layers.

After $L$ interaction layers, the final atom embeddings $\{\mathbf{h}_v^L\}_{v \in \CV}$ are either directly mapped to per-atom energy contributions via an energy head $\epsilon_v = f_{\text{energy}}(\mathbf{h}_v^L)$, and the total potential energy of the system is obtained by a permutation-invariant aggregation, $E(\CG) = \sum_{v \in \CV} \epsilon_v$. Depending on the architecture, atomic forces are computed as the negative gradient of the energy with respect to atomic positions, $\mathbf{F}_v = - \nabla_{\mathbf{r}_v} E(\CG)$~\citep{Chen2022}, or directly from node embeddings, i.e., $\mathbf{F}_v = f_{\text{force}}(\mathbf{h}_v^L)$~\citep{neumann2024Orbfastscalableneural}.

\subsection{Merging through Node Embedding Alignment}
Under the node embedding alignment objective, we aim to align the intermediate atom embeddings produced by the merged model $\Theta_M$ with those produced by the corresponding source model $\Theta_i$. Specifically, for any input graph $\CG$ belonging to domain $\mathcal{D}_i$, we minimize the discrepancy between the atom embeddings generated by $\Theta_M$ and those generated by $\Theta_i$ at each interaction layer $\ell$. Formally, we consider the objective:
\vspace{-0.05in}
\begin{alignat}{2}
\nonumber
\textstyle
\overbrace{\sum_{i=1}^n \mathbb{I}(\CG \in \mathcal{D}_i)}^{\substack{\text{Select source} \\ \text{model}}} 
  \overbrace{\sum_{\ell=1}^L}^{\substack{\text{\gnn} \\ \text{layer}}} 
  \overbrace{\sum_{\forall v \in \CV}}^{\text{node}} 
&\big\|
\Theta_M^{\ell}\!\left(\ch_{v,M}^{\ell-1}, \CS_{v,M}^{\ell-1}, \mathbf{R}\right)\\
\label{eq:opt_ff}
&-
\Theta_i^{\ell}\!\left(\ch_{v,i}^{\ell-1}, \CS_{v,i}^{\ell-1}, \mathbf{R}\right)\big\|_2^2,
\end{alignat}
where $\mathbb{I}(\CG \in \mathcal{D}_i)$ is the indicator function active only when the input graph $\CG$ belongs to the domain of model $\Theta_i$. Here, $\ch_{v,i}^{\ell-1}$ denotes the embedding of atom $v$ at layer $\ell-1$ under model $\Theta_i$, and $\CS_{v,i}^{\ell-1}$ denotes the multiset of neighboring atom embeddings augmented with geometric features derived from $\mathbf{R}$ (analogously defined for $\Theta_M$). The operators $\Theta_i^{\ell}$ and $\Theta_M^{\ell}$ represent the transformations performed at interaction layer $\ell$.

The objective in Eq.~\ref{eq:opt_ff} seeks to determine the parameters of each interaction layer of the merged model such that it reproduces the intermediate atom embeddings of the source force fields. However, directly optimizing Eq.~\ref{eq:opt_ff} is computationally prohibitive, since the embedding at layer $\ell$ depends on the outputs of all preceding layers. This coupling necessitates backpropagation through the full network during merging, undermining the efficiency benefits of the approach. In the following, we introduce a mild relaxation of this objective enabling a closed-form solution for linear blocks, dramatically reducing its computational cost.\looseness=-1

\begin{figure}[t]
\centering
\includegraphics[width=1.01\linewidth]{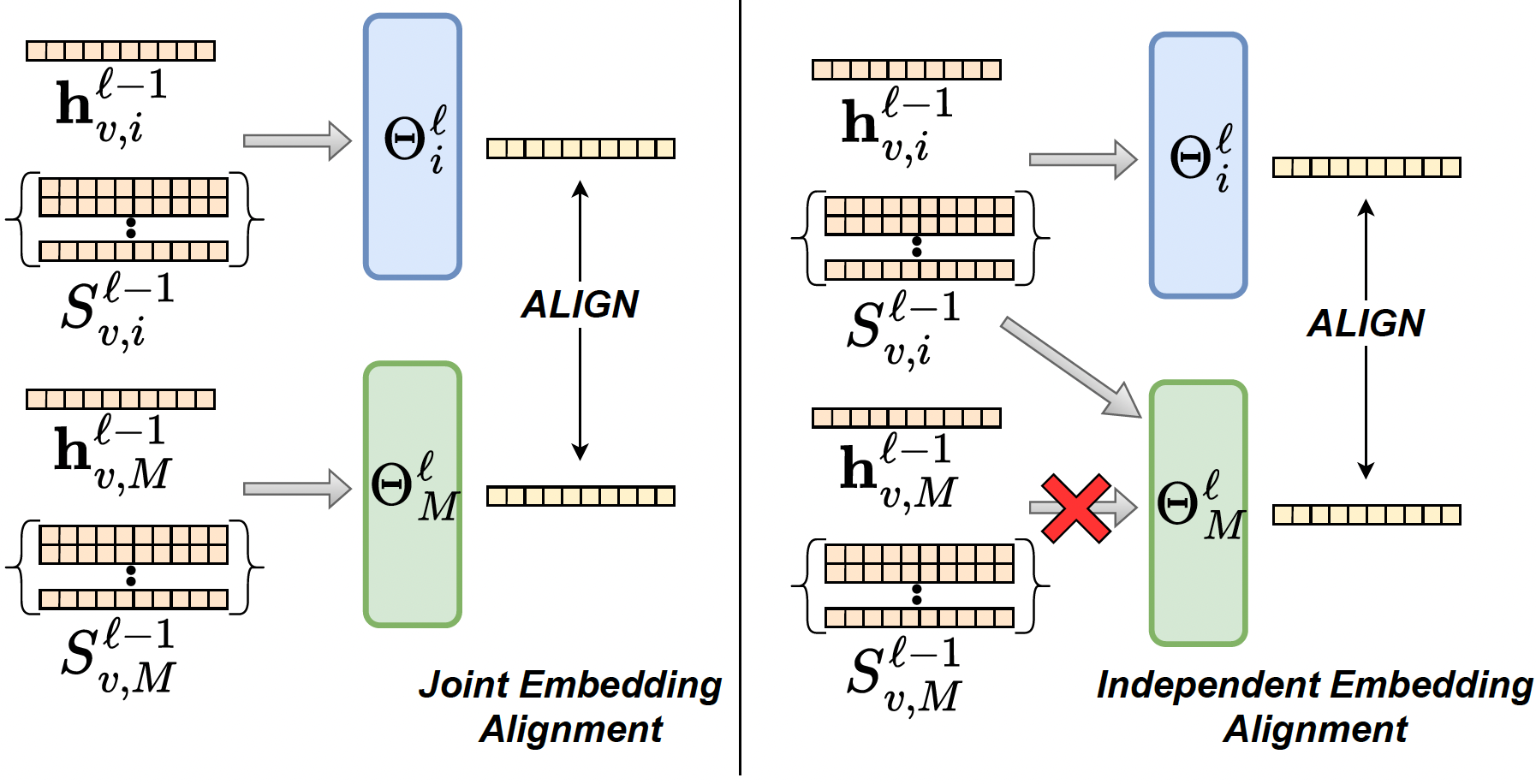}

    \caption{Illustrates the idea of independent layer-wise merging.}
    \label{fig:independent}
    
\end{figure}

\subsection{Closed-Form Linear Merging}
\label{sec:closed_form}

\textbf{Relaxing layer dependencies.}
Instead of aligning the atom embeddings produced by the merged \gnn force field end-to-end, we align each interaction-layer embeddings \textit{independently} across layers (See Fig.~\ref{fig:independent}). This relaxation decouples the merging objective across layers and removes the need for backpropagation through the full network during merging.

Concretely, rather than feeding the merged embeddings $\ch_{v,M}^{\ell-1}$ and neighborhood representations $\CS_{v,M}^{\ell-1}$ into the $\ell$-th interaction layer of $\Theta_M$, we directly use the corresponding embeddings $\ch_{v,i}^{\ell-1}$ and $\CS_{v,i}^{\ell-1}$ produced by the \textit{relevant source model} $\Theta_i$ when evaluated on the same atomic configuration. 
\vspace{-0.05in}
\begin{equation}
\label{eq:optrelax_ff}
\begin{split}
\sum_{i=1}^n \mathbb{I}(\CG \in \mathcal{D}_i)
\sum_{\ell=1}^L
\sum_{v \in \CV}
\bigg\|
&\Theta_M^{\ell}\!\left(\ch_{v,i}^{\ell-1}, \CS_{v,i}^{\ell-1}, \mathbf{R}\right) \\
&- \Theta_i^{\ell}\!\left(\ch_{v,i}^{\ell-1}, \CS_{v,i}^{\ell-1}, \mathbf{R}\right)
\bigg\|_2^2 .
\end{split}
\end{equation}
\vspace{-0.05in}

This relaxation is motivated by the property of \textit{transitive consistency} across layers. Specifically, at the input layer, all models share the same atomic features and geometry (i.e., $\ch_v^0 = \cx_v$ and $\mathbf{R}$ is fixed). If the merged interaction layer $\Theta_M^{\ell}$ accurately approximates the transformations of the relevant source model $\Theta_i$ for inputs from its domain $\mathcal{D}_i$, then the resulting atom embeddings satisfy $\ch_{v,M}^{\ell} \approx \ch_{v,i}^{\ell}$. Consequently, the induced neighborhood representations $\CS_{v,M}^{\ell}$ remain close to $\CS_{v,i}^{\ell}$.

\textbf{Closed-form solution:} 
In each interaction layer of a GNN force field, the computation can be decomposed into two components: 
(1) non-learnable aggregation operations that depend on molecular geometry, such as neighbor summation or radial basis pooling, and 
(2) learnable linear or affine transformations applied to atom embeddings or aggregated messages. During model merging, the atomic configuration and associated geometric features are treated as fixed inputs. Under this condition, the learnable parameters in an interaction layer appear primarily in linear transformations. Let the learnable weight matrices in layer $\ell$ of model $\Theta_i$ be denoted by $\cW_{1,i}^{\ell}, \ldots, \cW_{K,i}^{\ell}$, where $K$ is the number of linear transformations in that layer. The corresponding input vectors are denoted by $\cz_{v,1,i}^{\ell-1}, \ldots, \cz_{v,K,i}^{\ell-1}$, and the resulting outputs by $\cg_{v,1,i}^{\ell}, \ldots, \cg_{v,K,i}^{\ell}$. While these weight matrices are shared across all atoms, the input embeddings are atom specific. Restricting attention to these linear parameter blocks, the relaxed objective in Eq.~\ref{eq:optrelax_ff} can be rewritten as:
\vspace{-0.05in}
\begin{equation}
\label{eq:cf1}
\sum_{i=1}^n \mathbb{I}(\CG \in \mathcal{D}_i)
\sum_{\ell=1}^L
\sum_{k=1}^K
\sum_{v \in \CV}
\left\|
\cz_{v,k,i}^{\ell-1}\cW_{k,M}^{\ell}
-
\cg_{v,k,i}^{\ell}
\right\|_2^2 .
\end{equation}

Since each linear transformation in each layer is independent, minimizing Eq.~\ref{eq:cf1} is equivalent to optimizing each $\cW_{k,M}^{\ell}$ separately:
\vspace{-0.1in}
\begin{equation}
\label{eq:cf2}
\min_{\cW_{k,M}^{\ell}}
\sum_{i=1}^n \mathbb{I}(\CG \in \mathcal{D}_i)
\sum_{v \in \CV}
\left\|
\cz_{v,k,i}^{\ell-1}\cW_{k,M}^{\ell}
-
\cg_{v,k,i}^{\ell}
\right\|_2^2 .
\end{equation}

Let $\CZ_{k,i}^{\ell-1}$ denote the matrix formed by stacking $\cz_{v,k,i}^{\ell-1}$ over all $v \in \CV$, and let $\cG_{k,i}^{\ell}$ denote the corresponding matrix of outputs $\cg_{v,k,i}^{\ell}$. Eq.~\ref{eq:cf2} can then be rewritten as:
\vspace{-0.1in}
\begin{equation}
\label{eq:cf3}
\min_{\cW_{k,M}^{\ell}}
\sum_{i=1}^n \mathbb{I}(\CG \in \mathcal{D}_i)
\left\|
\CZ_{k,i}^{\ell-1}\cW_{k,M}^{\ell}
-
\cG_{k,i}^{\ell}
\right\|_F^2 ,
\end{equation}
where $\|\cdot\|_F$ denotes the Frobenius norm.

Since Eq.~\ref{eq:cf3} is convex (see Appendix~\ref{app:convexity}), the minimum is achieved when the gradient with respect to $\cW_{k,M}^{\ell}$ is zero, yielding:
\vspace{-0.1in}
\begin{alignat}{2}
\nonumber
(\cW_{k,M}^{\ell})^{\mathtt{T}}
=&
\left(
\sum_{i=1}^n \mathbb{I}(\CG \in \mathcal{D}_i)
(\cG_{k,i}^{\ell})^{\mathtt{T}}\CZ_{k,i}^{\ell-1}
\right)\times\\
\label{eq:cfw}
&\left(
\sum_{i=1}^n \mathbb{I}(\CG \in \mathcal{D}_i)
(\CZ_{k,i}^{\ell-1})^{\mathtt{T}}\CZ_{k,i}^{\ell-1}
\right)^{-1}.
\end{alignat}

\begin{table*}[t!]
    \centering
    \caption{Predictive fidelity across domains on M3GNet. We report Force MAE and Energy MAE means across three seeds. Mean $\pm$ std are provided in Appendix Tables \ref{tab:full_results_mean_std_wt}--\ref{tab:full_results_mean_std_emr}. Best merging results are \textbf{bolded}.
     \textbf{Task Definitions:} 
    \textit{DHA}: Docosahexaenoic Acid.
    $^\ddagger$\textit{MD17 5-Task Mix}: Ethanol + Naphthalene + Salicylic Acid + Uracil + Aspirin. 
    $^\star$\textit{MD22 3-Task Mix}: Ac-Ala3-NHMe + DHA + Stachyose.}
    \label{tab:M3GNet_results}

    \scalebox{0.8}{
    \begin{tabular}{l|cc|cc|cc|cc||cc|cc}
    \toprule
    \multirow{2}{*}{\textbf{Dataset \& Scenario}} & \multicolumn{2}{c|}{\textbf{Wt Averaging}} & \multicolumn{2}{c|}{\textbf{Fisher}} & \multicolumn{2}{c|}{\textbf{TIES}} & \multicolumn{2}{c|}{\textbf{EMR}} & \multicolumn{2}{c||}{\textbf{\nameff}} & \multicolumn{2}{c}{\textbf{Joint FT}} \\
     & F & E & F & E & F & E & F & E & F & E & F & E \\
    \midrule


    \multicolumn{13}{c}{\cellcolor{green!10}\textbf{MD17}\textit{(kcal/mol/\AA{}, kcal/mol)}} \\
    Aspirin + Uracil & 1.79 & 0.82 & 1.61 & 1.05 & 3.00 & 4.08 & 1.87 & 1.17 & \textbf{1.01} & \textbf{0.03} & 0.90 & 0.03 \\
    Ethanol + Malonaldehyde & 1.90 & 1.60 & 1.97 & 2.23 & 3.77 & 2.76 & 1.33 & 0.52 & \textbf{0.73} & \textbf{0.03} & 0.65 & 0.03 \\
    Naphthalene + Salicylic Acid & 1.67 & 0.59 & 1.67 & 0.74 & 2.65 & 3.89 & 1.67 & 0.88 & \textbf{0.82} & \textbf{0.02} & 0.74 & 0.02 \\
    Ethanol + Malonaldehyde + Aspirin & 2.53 & 2.42 & 2.27 & 2.87 & 3.65 & 2.52 & 2.27 & 0.75 & \textbf{0.89} & \textbf{0.04} & 0.78 & 0.03 \\
    Naphthalene + Salicylic Acid + Uracil & 2.05 & 1.30 & 1.72 & 1.54 & 2.77 & 3.88 & 1.81 & 1.65 & \textbf{0.87} & \textbf{0.03} & 0.72 & 0.03 \\
    \textit{5-Task Mix (MD17)$^\ddagger$} & 2.40 & 2.36 & 2.22 & 2.53 & 2.81 & 2.86 & 2.16 & 2.76 & \textbf{0.89} & \textbf{0.03} & 0.72 & 0.03 \\
    \midrule

    \multicolumn{13}{c}{\cellcolor{blue!10}\textbf{MD22}\textit{(kcal/mol/\AA{}, kcal/mol)}} \\
    Ac-Ala3-NHMe + AT-AT & 2.87 & 1.96 & 3.24 & 2.47 & 5.38 & 4.98 & 3.49 & 1.81 & \textbf{1.65} & \textbf{0.05} & 1.24 & 0.04 \\
    DHA + stachyose & 3.34 & 1.88 & 3.69 & 2.59 & 4.61 & 3.72 & 3.53 & 2.13 & \textbf{1.70} & \textbf{0.04} & 1.48 & 0.05 \\
    \textit{3-Task Mix (MD22)$^\star$} & 3.43 & 2.08 & 3.45 & 2.27 & 5.12 & 0.98 & 4.75 & 7.36 & \textbf{1.83} & \textbf{0.06} & 1.38 & 0.03 \\
    \midrule

    \multicolumn{13}{c}{\cellcolor{orange!10}\textbf{LiPS20}\textit{(eV/\AA{}, eV)}} \\
    $\beta$-Li$_3$PS$_4$ + $\gamma$-Li$_3$PS$_4$ & 51.79 & 85.56 & 18.85 & 66.33 & 29.20 & 160.82 & 117.31 & 186.59 & \textbf{0.72} & \textbf{0.16} & 0.44 & 0.18 \\
    Li$_3$P + Li$_2$S & 7.46 & 48.46 & 22.54 & 34.61 & 4.04 & 190.47 & 28.41 & 66.11 & \textbf{0.42} & \textbf{0.09} & 0.32 & 0.07 \\
    Li$_2$S + Li$_3$P + P$_2$S$_5$ & 7.07 & 137.61 & 19.51 & 103.48 & 16.33 & 186.29 & 72.60 & 99.09 & \textbf{1.21} & \textbf{0.86} & 0.34 & 0.11 \\
    \bottomrule
    \end{tabular}
    }
\end{table*}

\vspace{-0.1in}
\textbf{Efficiency implications.}
Because interaction layers and linear transformations are merged independently, the merging procedure avoids end-to-end backpropagation and admits embarrassingly parallel implementation across layers and parameter blocks. This yields orders-of-magnitude reductions in computational cost relative to retraining.

\comment{
Crucially, many interaction blocks in modern \gnn force fields, such as linear projections applied to aggregated messages or atom embeddings, are linear in their parameters. Under the relaxed objective in Eq.~\ref{eq:optrelax_ff}, these linear blocks admit closed form solutions that \textit{optimally} match the activations of the source force fields. In the following, we derive these analytical merging rules for representative \gnn force field architectures, leading to substantial computational savings over gradient based merging or retraining.

To ensure the merged parameters respect the underlying data manifolds of each source model, we formulate a \textbf{feature-reconstruction objective}.

For a given linear layer with weights $W$, we seek a merged weight $W^*$ that minimizes the discrepancy between its output and the outputs of the source models on a representative subset of data. Formally:
\begin{equation}
\label{eq:objective}
    W^* = \arg \min_{W} \sum_{k=1}^K \sum_{G \in \mathcal{S}_k} \| W x_k(G) - W_k x_k(G) \|_2^2
\end{equation}
where $\mathcal{S}_k \subset \mathcal{D}_k$ denotes a calibration subset sampled from the $k$-th dataset, and $x_k(G)$ represents the \textbf{input feature vector} to the current layer when graph $G$ is processed by source model $\mathcal{M}_k$. This objective forces the merged layer to mimic the behavior of model $\mathcal{M}_k$ on its specific domain.

Since Eq. \ref{eq:objective} is a linear least-squares problem, it admits a closed-form analytical solution. Taking the derivative w.r.t. $W$ and setting it to zero yields:
\begin{equation}
\begin{split}
    W^* \left( \sum_{k=1}^K \sum_{G \in \mathcal{S}_k} x_k(G) x_k(G)^\top \right) \\
    = \sum_{k=1}^K \sum_{G \in \mathcal{S}_k} W_k x_k(G) x_k(G)^\top
\end{split}
\end{equation}

Let $C_k = \sum_{G \in \mathcal{S}_k} x_k(G) x_k(G)^\top$ be the uncentered covariance matrix of the activations estimated over the subset $\mathcal{S}_k$. The optimal merged weight is given by:
\begin{equation}
\label{eq:solution}
    W^* = \left( \sum_{k=1}^K W_k C_k \right) \left( \sum_{k=1}^K C_k \right)^{-1}
\end{equation}}

\comment{
\subsection{Switch Embeddings Strategy}
While Eq. \ref{eq:solution} efficiently aligns the interaction layers, the initial atomic embeddings (Layer 0) present a unique challenge. The semantic meaning of an embedding vector (e.g., for Carbon) diverges significantly during \textit{independent fine-tuning}. Consequently, \textbf{forcing these distinct representations into a single unified matrix} leads to destructive interference and loss of domain-specific chemical information.

To address this, we introduce the \textbf{Switch Embeddings} strategy. We maintain the original embedding banks $E_k$ from each source model. During inference, for an input graph $G$, we dynamically route the embedding lookup:
\begin{equation}
    h_v^{(0)} = E_k(\text{atom\_type}(v)) \quad \text{if } G \in \mathcal{D}_k
\end{equation}
This ensures that the analytically merged layers receive input features that match the manifold they were optimized for.
}

\subsection{Targeted Fine-Tuning}
While the closed-form merger yields a strong and well-aligned initialization for the merged model, it does not fully capture the representational capacity of the source models, or in the presence of non-linear transformations. To recover the remaining predictive fidelity, we introduce a brief \textit{targeted fine-tuning} stage.

Specifically, we perform \textit{partial fine-tuning} of the later, task-adaptive layers (such as energy or force heads, or any non-linear layers) while keeping the analytically merged early layers frozen. This design preserves the alignment achieved by the closed-form solution, while allowing the model to absorb residual discrepancies arising from non-linear interactions and task-specific inductive biases.

\comment{
Crucially, this fine-tuning is performed \textbf{in conjunction with the Switch Embeddings strategy}. During each training step, the model dynamically swaps the frozen embedding bank ($E_k$) to match the source domain of the current batch. This forces the trainable shared weights to learn a representation compatible with the distinct chemical inputs of all source models simultaneously.

We freeze the initial node embeddings and early layers, updating only:
\begin{itemize}
    \item \textbf{M3GNet:} The final Readout MLP and the last 3 Interaction Blocks.
    \item \textbf{Orb:} The Force Head and the last 3 Interaction Blocks.
\end{itemize}

Furthermore, since the lower layers are already aligned via the closed-form solution, this stage requires  fewer epochs and a subset of the original training data compared to full joint training.}

\begin{table*}[t!]
    \centering
    \caption{Predictive Fidelity across domains on Orb. We report only force MAE since in Orb the embeddings directly go to a force head, whereas in M3GNet, the force is computed as the negative gradient of the energy. Mean $\pm$ std are provided in Appendix Tables \ref{tab:Orb_results_app_wt_fisher}--\ref{tab:Orb_results_app_ties_emr}. Best merging results are \textbf{bolded}. \textbf{Task Definitions:} See Table~\ref{tab:M3GNet_results} caption for the exact composition of the $^\ddagger$\textit{MD17 5-Task Mix} and $^\star$\textit{MD22 3-Task Mix}. }
    \label{tab:Orb_results}
    
    \scalebox{0.8}{
    \begin{tabular}{l|ccccc|c}
    \toprule
    \textbf{Dataset \& Scenario} & \textbf{Wt Avg} & \textbf{Fisher} & \textbf{TIES} & \textbf{EMR} & \textbf{\nameff} & \textbf{Joint FT} \\
    \midrule
    \multicolumn{7}{c}{\cellcolor{green!10}\textbf{MD17} \textit{(Force MAE in kcal/mol/\AA{})}} \\
    Aspirin + Uracil & 4.06 & 4.87 & 3.80 & 4.66 & \textbf{0.85} & 0.78 \\
    Ethanol + Malonaldehyde & 2.96 & 3.91 & 3.18 & 3.24 & \textbf{0.95} & 0.87 \\
    Naphthalene + Salicylic Acid & 1.89 & 2.37 & 1.99 & 2.24 & \textbf{0.62} & 0.54 \\
    Ethanol + Malonaldehyde + Aspirin & 3.25 & 4.35 & 3.49 & 5.13 & \textbf{1.01} & 0.85 \\
    Naphthalene + Salicylic Acid + Uracil & 3.41 & 4.22 & 3.27 & 5.17 & \textbf{0.67} & 0.52 \\
    \textit{5-Task Mix (MD17)$^\ddagger$} & 4.06 & 5.05 & 3.98 & 7.59 & \textbf{0.94} & 0.63 \\
    \midrule
    \multicolumn{7}{c}{\cellcolor{blue!10}\textbf{MD22}  \textit{(Force MAE in kcal/mol/\AA{})}} \\
    Ac-Ala3-NHMe + AT-AT & 3.84 & 4.79 & 3.73 & 4.65 & \textbf{1.10} & 0.96 \\
    DHA + stachyose & 3.48 & 4.47 & 3.38 & 4.02 & \textbf{1.04} & 0.91 \\
    \textit{3-Task Mix (MD22)$^\star$} & 3.83 & 5.08 & 3.56 & 7.09 & \textbf{1.15} & 0.89 \\
    \midrule
    \multicolumn{7}{c}{\cellcolor{orange!10}\textbf{LiPS20} \textit{(Force MAE in eV/\AA{})}}\\
    $\beta$-Li$_3$PS$_4$ + $\gamma$-Li$_3$PS$_4$ & 0.31 & 0.31 & 0.33 & 0.31 & \textbf{0.10} & 0.06 \\
    Li$_3$P + Li$_2$S & 0.30 & 0.31 & 0.32 & 0.30 & \textbf{0.09} & 0.06 \\
    Li$_2$S + Li$_3$P + P$_2$S$_5$ & 0.41 & 0.48 & 0.41 & 0.45 & \textbf{0.13} & 0.07 \\
    \bottomrule
    \end{tabular}
    }
\end{table*}

\setcounter{table}{3}
\begin{table*}[b]
    \centering
    \caption{Rollout cumulative EV/FV comparison across domains on M3GNet. \textbf{Task Definitions:} $^\ddagger$\textit{5-Task Mix}: Ethanol + Naph. + Sali. Acid + Uracil + Aspirin. $^\star$\textit{3-Task Mix}: Ac-Ala3-NHMe + DHA + Stachyose.}
    \label{tab:M3GNet_ev_fv_main}
    
    \scalebox{0.8}{
    \begin{tabular}{l|cc|cc|cc|cc||cc|cc}
    \toprule
    \multirow{2}{*}{\textbf{Scenario}} & \multicolumn{2}{c|}{\textbf{Wt Avg}} & \multicolumn{2}{c|}{\textbf{Fisher}} & \multicolumn{2}{c|}{\textbf{TIES}} & \multicolumn{2}{c||}{\textbf{EMR}} & \multicolumn{2}{c|}{\textbf{\nameff}} & \multicolumn{2}{c}{\textbf{Joint FT}} \\
     & EV & FV & EV & FV & EV & FV & EV & FV & EV & FV & EV & FV \\
    \midrule

    \multicolumn{13}{c}{\cellcolor{green!10}\textbf{MD17}} \\
    Aspirin + Uracil & 1.62e-09 & 0.543 & 8.75e-10 & 0.561 & 2.04e-08 & 0.707 & 3.45e-09 & 0.653 & 8.67e-11 & 0.509 & 8.76e-11 & 0.496 \\
    Eth. + Malon. + Asp. & 1.18e-08 & 0.680 & 1.64e-08 & 0.673 & 1.24e-08 & 0.671 & 2.34e-09 & 0.729 & 2.85e-10 & 0.594 & 2.71e-10 & 0.562 \\
    \textit{5-Task Mix (MD17)$^\ddagger$} & 1.33e-08 & 0.652 & 1.41e-08 & 0.653 & 1.59e-08 & 0.663 & 2.91e-08 & 0.759 & 2.12e-10 & 0.562 & 2.12e-10 & 0.527 \\
    \midrule

    \multicolumn{13}{c}{\cellcolor{blue!10}\textbf{MD22}} \\
    Ac-Ala3 + AT-AT & 7.35e-09 & 0.619 & 1.16e-08 & 0.684 & 4.49e-08 & 0.795 & 4.55e-09 & 0.721 & 3.55e-11 & 0.586 & 3.88e-11 & 0.480 \\
    \textit{3-Task Mix (MD22)$^\star$} & 1.26e-08 & 0.715 & 1.37e-08 & 0.688 & 3.39e-09 & 0.819 & 1.55e-07 & 0.831 & 4.01e-11 & 0.641 & 4.58e-11 & 0.548 \\
    \bottomrule
    \end{tabular}
    }
\end{table*}

\setcounter{table}{2}
\begin{table}[t]
    \centering
    \caption{Comparison of total training time (seconds) between Joint Fine-Tuning and \nameff. Full timing results for all 24 experimental scenarios are provided in Table~\ref{tab:efficiency_full}.
    \textbf{Task Definitions:} 
    \textit{DHA}: Docosahexaenoic Acid. 
    \textit{Stach.}: Stachyose.}
    \label{tab:efficiency_summary}
    
    \resizebox{\columnwidth}{!}{ 
    \begin{tabular}{l|c|c|c}
    \toprule
    \textbf{Merging Scenario} & \textbf{Joint FT (s)} & \textbf{\nameff\ (s)} & \textbf{Speedup} \\
    \midrule
    
    \multicolumn{4}{c}{\cellcolor{blue!10}\textbf{Architecture: M3GNet}} \\
    MD17: Aspirin + Uracil & $533.56 \pm 2.13$ & $\mathbf{49.82 \pm 0.35}$ & $\mathbf{10.71\times}$ \\
    MD17: Eth. + Malon. + Asp. & $615.58 \pm 0.88$ & $\mathbf{90.37 \pm 0.91}$ & $\mathbf{6.81\times}$ \\
    MD22: DHA + Stachyose & $5714.04 \pm 48.55$ & $\mathbf{655.46 \pm 6.29}$ & $\mathbf{8.72\times}$ \\
    MD22: Ac-Ala3 + DHA + Stach. & $7358.42 \pm 2.13$ & $\mathbf{264.62 \pm 3.30}$ & $\mathbf{27.81\times}$ \\
    LiPS: Li$_3$P + Li$_2$S & $3269.65 \pm 25.02$ & $\mathbf{593.12 \pm 0.07}$ & $\mathbf{5.51\times}$ \\
    LiPS: Li$_2$S + Li$_3$P + P$_2$S$_5$ & $3717.50 \pm 44.83$ & $\mathbf{503.99 \pm 4.73}$ & $\mathbf{7.38\times}$ \\
    \midrule
    
    \multicolumn{4}{c}{\cellcolor{orange!10}\textbf{Architecture: Orb}} \\
    MD17: Aspirin + Uracil & $1028.77 \pm 26.41$ & $\mathbf{143.03 \pm 4.56}$ & $\mathbf{7.19\times}$ \\
    MD17: Eth. + Malon. + Asp. & $1121.72 \pm 79.47$ & $\mathbf{183.03 \pm 17.62}$ & $\mathbf{6.13\times}$ \\
    MD22: DHA + Stachyose & $2937.06 \pm 11.20$ & $\mathbf{487.08 \pm 10.63}$ & $\mathbf{6.03\times}$ \\
    MD22: Ac-Ala3 + DHA + Stach. & $3919.08 \pm 3.96$ & $\mathbf{548.52 \pm 2.40}$ & $\mathbf{7.14\times}$ \\
    LiPS: Li$_3$P + Li$_2$S & $2476.80 \pm 4.38$ & $\mathbf{483.42 \pm 3.55}$ & $\mathbf{5.12\times}$ \\
    LiPS: Li$_2$S + Li$_3$P + P$_2$S$_5$ & $2954.31 \pm 54.21$ & $\mathbf{501.50 \pm 17.77}$ & $\mathbf{5.89\times}$ \\
    \bottomrule
    \end{tabular}
    }
\end{table}

\section{Experiments}
\label{sec:experiments}
In this section, we present the first systematic benchmarking study of model merging for \gnn. Complete implementation is available at \url{https://github.com/idea-iitd/GFFMerge}. Our results establish:

\begin{itemize}
    \item \textbf{Efficacy:} Existing model merging algorithms, developed primarily for discrete prediction tasks on vision and text modalities, suffer from severe performance degradation when applied to force regression on atomistic systems. \nameff{} directly addresses this gap with dramatically superior performance.
    \item \textbf{Computational efficiency:} \nameff{} achieves up to \textbf{27$\times$} speedup, compared to standard Joint Fine-Tuning of pre-trained force fields, while incurring only minor losses in predictive accuracy. This efficiency stems from an analytical merging procedure that exploits structural properties of message-passing \gnn force fields, enabling direct computation of merged parameters without costly gradient-based optimization.
    \item \textbf{Generality beyond force fields:} Through extensive benchmarking, we demonstrate that the proposed closed-form model merging framework seamlessly extends to standard \gnn architectures such as \gcn,\sage, and graph transformers with efficiency gains of up to $3$ orders of magnitude.
\end{itemize}

\subsection{Experimental Setup}
The details of our hardware and software environment are listed in App.~\ref{app:setup}. To ensure statistical meaningful results, all experiments have been repeated thrice and the mean and standard-deviation of the metric are reported.

\textbf{Datasets \& Evaluation Scenarios.} We consider three complementary datasets that differ in molecular size, structural complexity, and interaction characteristics:

\begin{itemize}
    \item \textbf{MD17 (Small Molecules)~\citep{md17}:} Organic molecules with well-defined local chemical environments, enabling controlled evaluation of merging force fields. We evaluate model merging at multiple scales by combining two, three, and five models, each trained on distinct chemical systems.
    \item \textbf{MD22 (Supramolecular Systems)~\citep{md22}:} Large peptides and molecular complexes with long-range interactions and high conformational diversity, testing the scalability of model merging to complex, flexible systems.
    \item \textbf{LiPS20 (Solid-State Electrolytes)~\citep{Chen2025}:} Periodic crystalline systems exhibiting at high temperatures and increased mobility, assessing the robustness of merging in bulk systems.
\end{itemize}

\textbf{Baselines and Gold Standard.} We compare \nameff against several representative model-merging baselines, including simple weight averaging, Fisher merging~\citep{matena2022mergingfishermerging}, Ties~\citep{yadav2023tiesmerging}, and EMR~\citep{emr-merging-DBLP:journals/corr/abs-2405-17461}. Notably, none of these methods are designed for \gnns, and their evaluations have been largely restricted to discrete prediction settings such as classification or next-token prediction. We use the codebases released by the respective authors.

As an upper bound, we report results for the \textbf{Gold Standard}, which corresponds to full fine-tuning (FT) of all parameters of the foundation model on the union of the training datasets from the source models, referred to as Joint FT.

\textbf{Foundation Models.} We evaluate \nameff on two state-of-the-art GNN force field foundation models, \textsc{Orb}~\citep{neumann2024Orbfastscalableneural} and \textsc{M3GNet}~\citep{Chen2022}. To assess generality beyond force fields, we additionally benchmark the method on standard \gnn architectures, including \gcn, \gin, \gat, and \sage, across node classification and link prediction (discussed in App.~\ref{app:gnnmerge}). For these generic \gnns, the fine-tuning stage is omitted; this variant is referred to simply as \name. 

\subsection{Efficacy}
\label{sec:efficacy}
Tables~\ref{tab:M3GNet_results} and \ref{tab:Orb_results} demonstrate that existing model-merging methods fail to transfer to \gnn force fields in M3GNet and Orb respectively. Naïve averaging and prior techniques such as Fisher, TIES, and EMR incur substantial degradation across all benchmarks, with force MAEs often 2–5× higher than Joint FT on MD17 and MD22, and catastrophic failures on LiPS20. These methods do not preserve the fine-grained activation structure required for stable energy and force regression, leading to amplified errors.

In contrast, \nameff consistently achieves superior performance that is close to the Gold Standard (Joint FT) across molecular, supramolecular, and bulk regimes. On MD17 with M3GNet, \nameff remains within 5–15\% of Joint FT even for 5-task mixtures, while on MD22 it reduces force error by more than 2$\times$ relative to the strongest baseline. Table 2 shows the same trends on Orb, confirming that the proposed framework generalizes across architectures. Importantly, baseline performance deteriorates sharply as more domains are merged, whereas \nameff exhibits stable scaling under multi-task composition.

\textbf{Why does \nameff outperform existing baselines?} The performance gains of \nameff follow directly from structural properties of \gnn force fields rather than heuristic design choices. First, the proposed embedding-alignment objective is convex when restricted to \gnn message-passing layers (App.~\ref{app:convexity}), ensuring a unique and stable merged solution. Second, modern \gnn force fields are dominated by linear transformations interleaved with fixed aggregation operators, which allows large portions of the model to be merged in closed form, eliminating approximation error inherent to generic parameter averaging schemes or optimization based approaches on non-convex loss surfaces.

\subsection{Efficiency Analysis}
\label{sec:efficiency}
Table~\ref{tab:efficiency_summary} summarizes the training duration across various scenarios. \nameff achieves speedups ranging from $5\times$ to $27\times$. These gains stem from analytically merging the majority of message-passing layers and restricting optimization to a small subset of parameters during targeted fine-tuning. 

\setcounter{table}{4}

\begin{figure*}[t]

    \centering
    \caption{\textbf{Ablation Analysis on Solid-State Electrolytes (Li$_3$P + Li$_2$S).} 
    \textbf{Top Row:} Test MAE vs. Fine-tuning Epochs. Both M3GNet (a,b) and Orb (c) converge to the Joint Training floor (Gold Dashed) within 15--20 epochs, demonstrating rapid adaptation to bulk systems exhibiting chaotic dynamics.
    \textbf{Bottom Row:} Test MAE vs. Data limit. \nameff maintains high fidelity even when fine-tuned on $<500$ samples (few-shot regime). \textit{Full FT} denotes the fine-tuning of pre-trained force field on combined data, \textbf{Appendix \ref{app:ablation_extended}} (Figures \ref{fig:ablation_md17} and \ref{fig:ablation_md22}) reports the results on other datasets.}
    \setlength{\tabcolsep}{1pt} 
    
    \begin{tabular}{ccc}
        \includegraphics[width=0.25\textwidth]{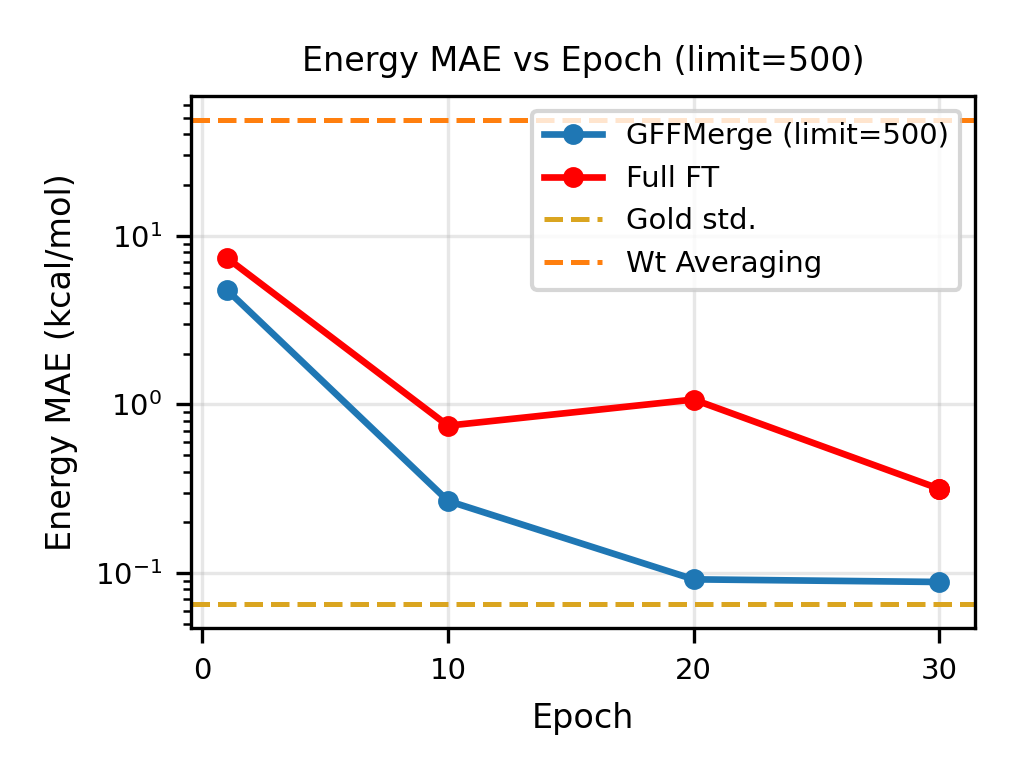} & 
        \includegraphics[width=0.25\textwidth]{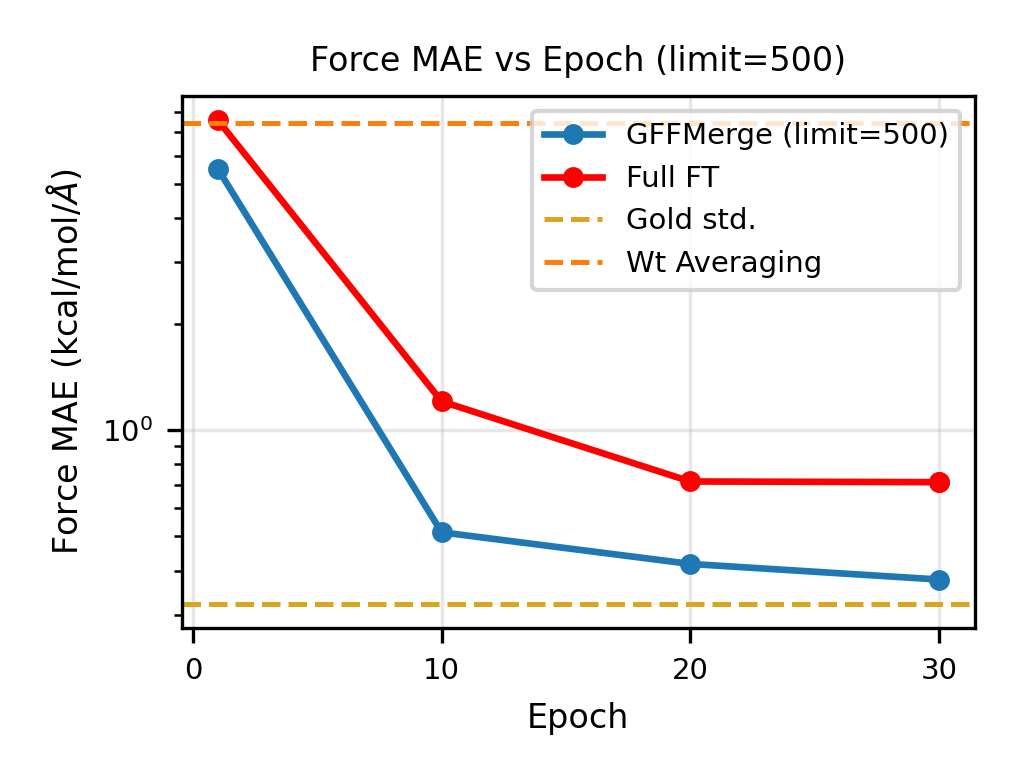} & 
        \includegraphics[width=0.25\textwidth]{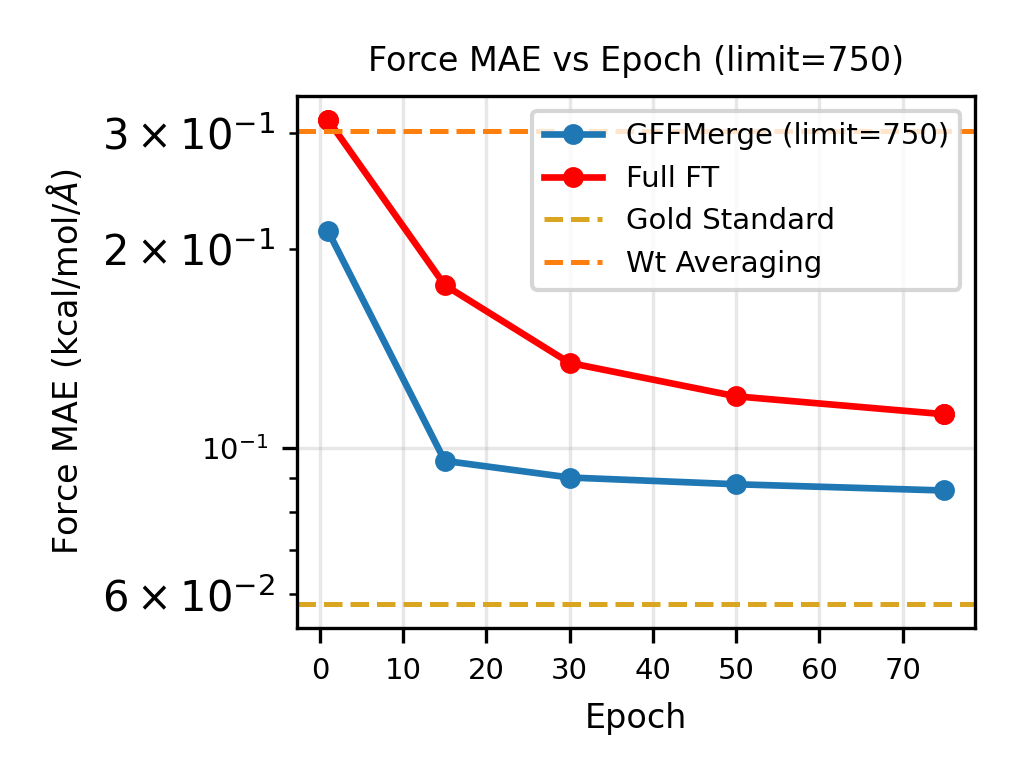} \\
        \footnotesize (a) M3GNet Energy (Convergence) & 
        \footnotesize (b) M3GNet Force (Convergence) & 
        \footnotesize (c) Orb Force (Convergence) \\
    \end{tabular}

        
    \begin{tabular}{ccc}
        \includegraphics[width=0.25\textwidth]{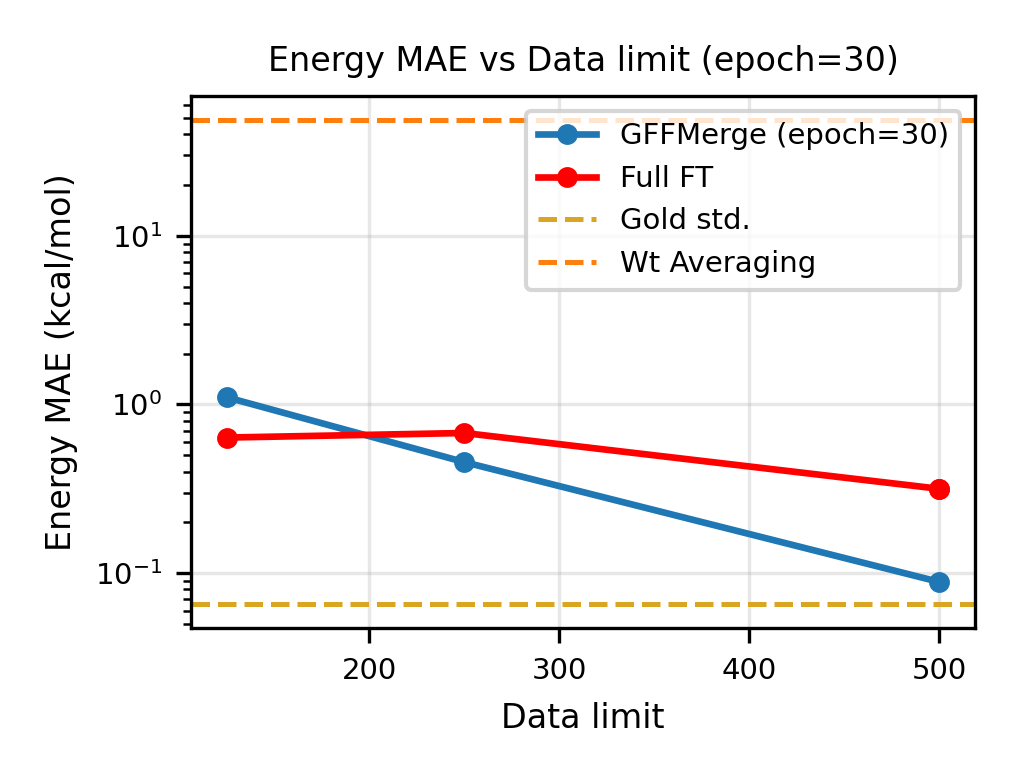} & 
        \includegraphics[width=0.25\textwidth]{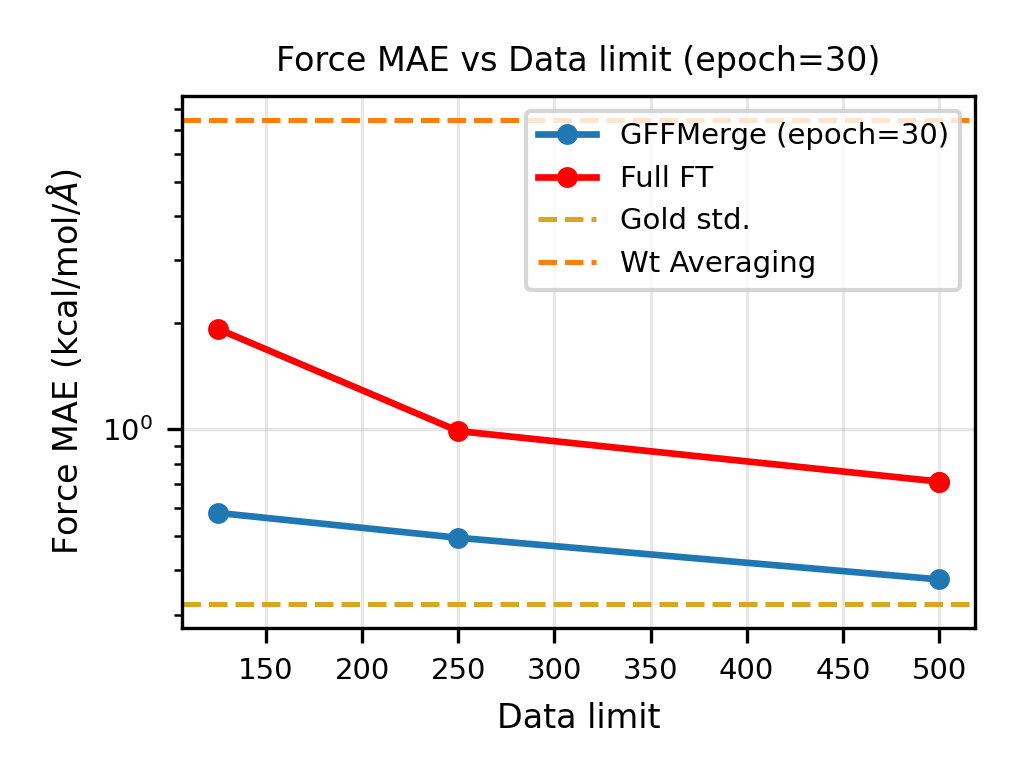} & 
        \includegraphics[width=0.25\textwidth]{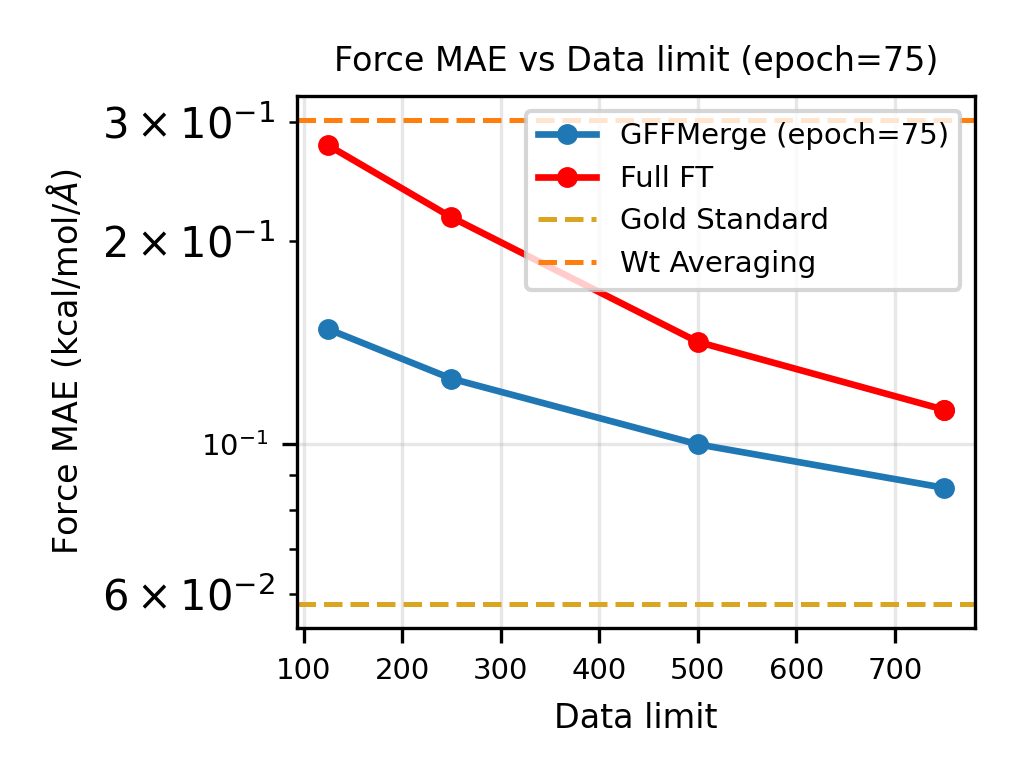} \\
        
        \footnotesize (d) M3GNet Energy (Efficiency) & 
        \footnotesize (e) M3GNet Force (Efficiency) & 
        \footnotesize (f) Orb Force (Efficiency) \\
    \end{tabular}
    
    \label{fig:main_ablation}

\end{figure*}

\begin{review}
\subsection{Stability Diagnostics}

To evaluate the physical viability of the merged models, we assess their structural fidelity and dynamic stability through 1000-step forward Molecular Dynamics (MD) simulations. Detailed formulations for all evaluation metrics are provided in Appendix~\ref{sec:appendix_metrics}.

We rigorously assess dynamic stability by initialising MD rollouts using configurations and velocities sampled from held-out reference trajectories. At each timestep, we compare the predicted energies and forces against the reference values to compute dimensionless, normalized \textbf{Energy Violations (EV)} and \textbf{Force Violations (FV)}. These violation errors are bounded between 0 and 1 where 0 denotes perfect agreement with the ground truth and 1 denotes total divergence; which are aggregated over the trajectory using the geometric mean. To ensure statistical robustness, this process is repeated across five different initial conditions. The EV and FV results for M3GNet are presented in Table~\ref{tab:M3GNet_ev_fv_main}.\looseness=-1

\nameff closely tracks the performance of the gold standard, achieving near-identical dynamic stability without the prohibitive computational costs of full retraining. We also notice that M3GNet is inherently more resilient and accurate than Orb in terms of maintaining rollout stability.
\end{review}

\subsection{Ablation Studies: Robustness \& Efficiency}
\label{sec:ablation}
To validate the robustness of \nameff, we analyze its sensitivity to two critical resource constraints: the volume of training data used for fine-tuning and the number of fine-tuning steps (epochs). We perform this analysis on the held-out \textbf{test set}.

\textbf{Few-Shot Generalization.} A key hypothesis of model merging is that the closed-form solution already positions weights in a favorable region of the loss landscape, requiring fewer samples and epochs to reach the gold standard. We test this by fine-tuning the merged model on restricted subsets of the training data (ranging from \textbf{125 to 750 samples}) and evaluating predictive fidelity on the test set. Epoch 0 shows the solution obtained purely by the closed-form procedure, which \textbf{already outperforms all baselines} before any fine-tuning. As shown in Figure \ref{fig:main_ablation} (Bottom Row), \nameff exhibits remarkable data efficiency. Starting from the closed-form initialization, our method recovers near-optimal fidelity (approaching the gold standard) using as few as \textbf{250 samples}, converging substantially faster than full fine-tuning from the foundation model.

\textbf{Convergence Dynamics.} Figure \ref{fig:main_ablation} (Top Row) reports the Force MAE achieved when the fine-tuning process is restricted to varying computational budgets (defined by the maximum number of epochs). Similar to the earlier case, we note that the closed-form initialization provides a significant ``jump start" over training from the foundation model confirming that the weights are indeed in a favorable configuration in the loss landscape. \nameff reaches the gold standard plateau with only \textbf{10--20 epochs}, whereas full fine-tuning requires significantly longer to learn the geometric representations. This accelerated convergence from a better initialization explains the speedups reported in Section \ref{sec:efficiency}.
\begin{table*}[t]
\centering
\caption{\textbf{In-domain Dataset Results.} \name compared with baselines on merging models trained on disjoint label splits of the same dataset. Metric: Accuracy $(\%)$.}
\renewcommand{\arraystretch}{1}
\scalebox{0.8}{
\begin{tabular}{l|ccccccccc}
\toprule
 \textbf{Dataset} & \textbf{M} & \textbf{Base} & \textbf{\wa} & \textbf{GIT} & \textbf{\permute} & \textbf{\zipit} & \textbf{\surgery} & \textbf{\name} \\ 
 \midrule

\multirow{2}{*}{\textbf{Arxiv}} 
& GNN A & 80.85 & 68.56 & 65.83 & 70.56 & 69.53 & 68.41$\pm$\scriptsize{1.82} & \textbf{77.47} \\ 
& GNN B & 82.51 & 42.09 & 10.48 & 58.34 & 57.21 & 66.41$\pm$\scriptsize{1.45} & \textbf{73.02} \\ 
\midrule

\multirow{2}{*}{\textbf{AmzComp}} 
& GNN A & 95.81 & 64.34 & 65.96 & 71.03 & 73.07 & 85.62$\pm$\scriptsize{1.16} & \textbf{93.94} \\ 
& GNN B & 93.09 & 79.42 & 49.08 & 85.42 & 81.35 & 78.45$\pm$\scriptsize{2.2} & \textbf{92.18} \\ 
\midrule

\multirow{2}{*}{\textbf{AmzPhoto}} 
& GNN A & 94.72 & 63.10 & 70.09 & 70.22 & 88.21 & 85.37$\pm$\scriptsize{1.22} & \textbf{94.72} \\ 
& GNN B & 94.81 & 68.61 & 39.16 & 81.12 & 73.53 & 90.89$\pm$\scriptsize{0.71} & \textbf{94.81} \\ 
\midrule

\multirow{2}{*}{\textbf{Cora}} 
& GNN A & 86.54 & 67.10 & 84.50 & 75.75 & 67.02 & 87.13$\pm$\scriptsize{0.12} & \textbf{85.38} \\ 
& GNN B & 93.03 & 92.72 & 48.10 & 89.71 & 88.33 & 89.43$\pm$\scriptsize{0.14} & \textbf{93.35} \\ 
\midrule

\multirow{2}{*}{\textbf{Reddit}} 
& GNN A & 97.59 & 88.38 & 92.10 & 92.33 & 90.01 & 83.59$\pm$\scriptsize{1.90} & \textbf{96.80} \\ 
& GNN B & 94.35 & 48.23 & 05.82 & 72.21 & 74.03 & 74.58$\pm$\scriptsize{2.41} & \textbf{94.30} \\ 
\midrule

\multirow{2}{*}{\textbf{WikiCS}} 
& GNN A & 86.63 & 59.31 & 81.09 & 62.41 & 69.92 & 68.07$\pm$\scriptsize{1.55} & \textbf{86.79} \\ 
& GNN B & 85.95 & 42.64 & 28.54 & 62.21 & 69.71 & 58.25$\pm$\scriptsize{2.65} & \textbf{84.54} \\ 
\midrule

\multirow{2}{*}{\textbf{Mag240M}} 
& GNN A & 72.73 & 47.82 & 57.29 & 52.62 & 53.88 & - & \textbf{69.31} \\ 
& GNN B & 74.18 & 43.72 & 03.51 & 51.47 & 54.16 & - & \textbf{69.82} \\ 
\midrule

\multicolumn{2}{c}{\textbf{Average}} 
& 90.49 & 65.37 & 53.39 & 74.27 & 75.16 & 78.01 & \textbf{88.94} \\ 
\bottomrule
\end{tabular}}
\label{tab:results1}
\vspace{-0.15in}
\end{table*}

\section{Generalization Beyond \gnn Force Fields}
\label{sec:gnnmerge}

The core principle of \nameff, that is, embedding alignment through closed-form linear merging, is architectural and not application-specific. In this section, we demonstrate generalization beyond \gnn force fields. We refer to this variant as \name. In this setting, only the closed-form merging step is applied. Their closed-form derivations from several mainstream \gnns and graph transformer is discussed in App.~\ref{app:derivation}. Since this procedure does not depend on labeled data, \name remains applicable even when the original training data are unavailable. Since the notion of foundation models is uncommon for generic \gnns such as \gcn or \sage, we evaluate \name in a more challenging regime where the source models are trained independently from scratch. 
We benchmark \name on three types of model merging scenarios:

\begin{enumerate}
    \item \textbf{Node Classification on In-domain Datasets:}\label{sec:in-domain}
    We create two disjoint label splits from the same dataset and train a model on each split independently. These models are subsequently merged simulating a scenario where new labels are introduced after the initial training.
    \item \textbf{Same Task on Different Datasets:}\label{sec:diff_datasets}  
    Given $N$ models, each trained for a task on a distinct dataset, we merge these models into a single unified model. The performance of the merged model is subsequently evaluated on the test sets of the respective datasets.
    \item \textbf{Node Classification and Link Prediction on Different Datasets:}\label{sec:diff_tasks}
    We merge models trained on different tasks on different datasets.
\end{enumerate}
\vspace{-0.05in}
\textbf{Datasets.} Table~\ref{tab:datasets1} lists the datasets used for experiments.\\
\textbf{Baselines.}  Since we do not have a common foundation model from which the source models have been fine-tuned, we include those model merging baselines that can merge independently trained models. These include Weight Averaging (\wa), \git~\citep{ainsworth2023gitrebasin}, \permute~\citep{entezari2022the}, and \zipit~\cite{stoica2024zipit} and  \surgery~\citep{yang2024representationsurgery}.\\
\textbf{Architectures.} While our main results are presented on \gcn, we also evaluate generalizations to \sage and \node~\cite{wu2022nodeformer} in Tables~\ref{tab:results8} and~\ref{tab:results7}.

\subsection{Results}
\label{sec:main_res}
\vspace{-0.05in}
\textbf{In-domain Datasets.} In Table~\ref{tab:results1}, we present the results of merging \gcn{}s trained on disjoint label splits (of equal sizes) for node classification tasks on the same dataset. \name{} achieves a significant improvement over existing methods, outperforming \surgery by $\mathbf{10.93\%}$, \zipit by $\mathbf{13.78\%}$, \permute by $\mathbf{14.67\%}$, \git by $\mathbf{35.53\%}$, and weight averaging by $\mathbf{23.57\%}$.

\textbf{Different Datasets.} Table~\ref{tab:results2half} (Appendix) presents the performance of merging \gcn{}s trained for node classification across two distinct datasets. \name outperforms the closest baseline by \textbf{7.02\%}. 

\textbf{Different Tasks.} Table~\ref{tab:results6} presents results  on merging \gcn{}s trained on two distinct tasks, node classification and link prediction, across two different datasets. In most cases, baseline methods perform no better than a randomly initialized \gcn on the link prediction task. Preserving node classification accuracy leads to a severe degradation in link prediction AUC for the baselines. In contrast, \name  achieves accuracies comparable to the individual models on their respective tasks. 
\vspace{-0.025in}
\textbf{More results:} In App.~\ref{app:gnnmerge}, we present results on merging \gnns of different parameter sizes and heterogeneous architectures. We also show performance on NodeFormer~\cite{wu2022nodeformer}, \sage, and \gat.

\begin{table}[htbp] 
\centering
\caption{Datasets used for benchmarking}
\label{tab:datasets1} 

\resizebox{\columnwidth}{!}{
\begin{tabular}{lcccc}
\toprule
\textbf{Dataset} & \textbf{\#Nodes} & \textbf{\#Edges} & \textbf{\#Class} & \textbf{\#Feat} \\
\midrule
Cora~\cite{pmlr-v48-yanga16} & 2,708 & 5,429 & 7 & 1,433 \\
Citeseer~\cite{pmlr-v48-yanga16} & 3,312 & 4,732 & 6 & 3,703\\
Pubmed~\cite{pmlr-v48-yanga16} &19,717 & 44,338 & 3 & 500\\
Arxiv~\cite{NEURIPS2020_fb60d411-ogbn} & 169,443 & 2,315,598 & 40 & 128 \\
Mag240M~\cite{hu2021ogblsc} & 244,160,499 & 1,728,364,232 & 153 & 768 \\
Papers100M~\cite{NEURIPS2020_fb60d411-ogbn} &  111,059,956 & 1,615,685,872 & 172 & 128 \\
WikiCS~\cite{mernyei2020wiki-wikics} & 11,701 & 431,726 & 10 & 300 \\
AmzPhoto~\cite{shchur2018pitfalls} & 7,650 & 238,162 & 8 & 745 \\
AmzComp~\cite{shchur2018pitfalls} & 13,752 & 491,722 & 10 & 767 \\
Reddit~\cite{NIPS2017_5dd9db5e-inductive-rep} & 232,965 & 114,615,892 & 41 & 602 \\
\bottomrule
\end{tabular}
}
\end{table}
\vspace{-0.15in}
\subsection{Scalability} 
MAG240M and Papers100M are two of OGBN's largest datasets. The leaderboards models (\sage) are available at OGBN~\cite{NEURIPS2020_fb60d411-ogbn}. To demonstrate a concrete application of model merging in scalability benefits, we aim to build a unified model capable of handling both datasets simultaneously. A joint training on the union of MAG240M and Papers100M require  \textbf{650GB} of RAM. Without sufficient memory, loading the datasets directly from disk results in about \textbf{2.5 hours} per training epoch, leading to total training times exceeding $5$ days. In contrast, \name{} requires \textbf{$\approx 2$ GB} of RAM and the entire merging process completes in under \textbf{8 minutes}, offering \textbf{$\approx 1000x$} speed-up in a memory-efficient manner.

\vspace{-0.05in}
\section{Conclusions}
We introduced \nameff (and \name), a principled framework for closed-form model merging that enables efficient and reliable reuse of pretrained graph neural networks without joint retraining. By exploiting the linear structure inherent in GNN message-passing and enforcing convex embedding alignment, \name provides an analytical alternative to existing heuristic merging methods, which are largely designed for discrete prediction settings and do not extend naturally to graph-structured or continuous domains.

Across a diverse set of benchmarks, including molecular force fields and graph learning tasks, we demonstrated that \nameff consistently achieves predictive fidelity comparable to joint fine-tuning, while reducing computational cost by fraction of magnitude.
Beyond performance gains, \name reframes model merging as a first-class primitive for scalable and modular learning in graph-based systems. This opens new opportunities for rapid adaptation, continual learning, and deployment in resource-constrained settings, where retraining large foundation models is infeasible. Future work includes studying robustness under distribution shift, and exploring dynamic or incremental merging in lifelong learning scenarios, and extending the merging to equivariant and other classes of \gnn{s}.
\pagebreak
\begin{review}
\section*{Acknowledgements}
 This work was supported by Yardi School of Artificial Intelligence, Department of Computer Science and Department of Civil Engineering at IIT Delhi. This work was partially supported by the CSE Research \mbox{Acceleration} Fund of IIT Delhi. The authors acknowledge the computational and storage resources provided by the IIT Delhi HPC facility.  Finally, the authors thank the reviewers for their insightful comments.
\end{review}

\section*{Impact Statement}
This work advances machine learning efficiency for atomistic simulations and graph learning through model merging in \gnns. The primary positive impact is democratizing access to computational chemistry and materials tools through 5-27$\times$ speedups, reducing both cost and carbon footprint compared to retraining. Potential risks include: (1) possible misuse in designing harmful compounds, though this is a general risk in computational chemistry not specific to our method; (2) unreliable predictions if users merge incompatible models without proper validation. We strongly recommend domain-specific validation before deployment in safety-critical applications. Finally, our method's efficiency gains apply to inference, re-training or fine-tuning, and deployment, not to the substantial environmental cost of training foundation models. We encourage responsible use through careful validation, assessment of domain compatibility, and awareness that computational efficiency should complement, not replace, rigorous scientific validation.

\bibliography{example_paper}
\bibliographystyle{icml2026}
\comment{
\section{Related Work}
\label{sec:related}

\textbf{Neural Force Fields.} Machine learning potentials, particularly \gnn-based ones like NequIP \cite{batzner20223}, M3GNet \cite{chen2022universal}, and Orb \cite{Orb2023}, capture complex atomic interactions. However, they typically suffer from catastrophic forgetting when sequentially fine-tuned, necessitating efficient merging strategies.

\textbf{Model Merging.} Techniques such as Weight Averaging (WA) \cite{izmailov2018averaging} and Fisher Merging \cite{matena2022merging} have shown success in linearizing loss landscapes for convex problems. RegMean \cite{jin2022dataless} attempts to merge via regression on feature activations. \nameff builds upon these by addressing the specific non-linearities and permutation invariances inherent to \gnns.}


\clearpage
\appendix
\onecolumn
\section{Convexity of the Minimization Objective}
\label{app:convexity}
We consider the following minimization objective for \nameff{}++:
\begin{equation}
    \min\limits_{\cW_{k,M}^{\ell}} \sum_{i=1}^n \mathbb{I}(\CG \in \mathcal{D}_i) \|\CZ_{k,i}^{\ell-1}\cW_{k,M}^{\ell} - \cG_{k,i}^{\ell}\|_F^2.
\end{equation}
For notational simplicity, we omit the indices $\ell, k,$ and $M$, yielding
\begin{equation}
    \min\limits_{\cW} \sum_{i=1}^n \mathbb{I}(\CG \in \mathcal{D}_i) \|\CZ_i \cW - \cG_i\|_F^2.
\end{equation}

Define each summand as
\[
f_i(\cW) = \|\CZ_i \cW - \cG_i\|_F^2.
\]
We show that each $f_i(\cW)$ is convex in $\cW$, which immediately implies that their sum is also convex.

Expanding $f_i(\cW)$, we obtain
\[
f_i(\cW)
= \operatorname{Tr}\!\left((\CZ_i \cW - \cG_i)^\top (\CZ_i \cW - \cG_i)\right)
\]
\[
= \operatorname{Tr}(\cW^\top \CZ_i^\top \CZ_i \cW)
- 2 \operatorname{Tr}(\cW^\top \CZ_i^\top \cG_i)
+ \operatorname{Tr}(\cG_i^\top \cG_i).
\]

The first term is a quadratic form in $\cW$ with matrix
\[
A = \CZ_i^\top \CZ_i,
\]
\section{Analytical Derivation For Other GNNs}

Graph Neural Networks (GNNs) have become the dominant paradigm for representation learning on graph-structured data~\citep{velickovic2018adriana, kipf2017semi-gcn, graphsage, xu2018how, graphreach, grafenne}. Their effectiveness has been established across a wide range of domains, including drug and material discovery~\cite{drugs, graphgen, stridernet}, traffic forecasting~\cite{frigate, neuromlr}, recommendation systems~\cite{triper, graphgini}, and the modeling of physically interacting systems~\cite{brognet, gnode, rigidbody}. In this section, we generalize model merging for generic \gnns. 
\label{app:derivation}
\subsection{GraphSAGE}
The node embedding update equation for \textbf{GraphSAGE} is as follows:
\begin{equation}
\mathbf{h}_v^{(\ell)} = \sigma \left(\mathbf{h}_v^{(\ell-1)} \mathbf{W_1}^{(\ell)} \left| \right| \sum_{u \in \mathcal{N}_v} \frac{1}{\lvert \mathcal{N}_v \rvert} \mathbf{h}_u^{(\ell-1)} \mathbf{W_2}^{(\ell)}\right)
\end{equation}
where:
\begin{itemize}
    \item \(\mathcal{N}_v\): Set of neighbors of node \(i\).
    \item \(\mathbf{h}_u^{(\ell)}\): Feature vector of node \(u\) at layer $\ell$.
    \item $\sigma$: Activation function like ReLU. 
    \item \(\mathbf{W_1}^{(\ell)},\mathbf{W_2}^{(\ell)}\): Trainable weight matrices at layer $\ell$.
\end{itemize}

The trainable weight matrix \(\mathbf{W_2}^{(\ell)}\) can be factored out to obtain:
\begin{equation}    
\mathbf{h}_v^{(\ell)} = \sigma \left(\mathbf{h}_v^{(\ell-1)} \mathbf{W_1}^{(\ell)} \left| \right| \left(\sum_{u \in \mathcal{N}_v} \frac{1}{\lvert \mathcal{N}_v \rvert} \mathbf{h}_u^{(\ell-1)}\right) \mathbf{W_2}^{(\ell)}\right)
\end{equation}

For a given target node, the term $\left(\sum_{u \in \mathcal{N}_v} \frac{1}{\lvert \mathcal{N}_v \rvert} \mathbf{h}_u^{(\ell-1)}\right)$ can be computed independent of $\mathbf{W_2}^{(\ell)}$ and can be denoted by $\mathbf{k}_v^{(\ell)}$. Hence, the node update equation becomes:
\begin{equation}
\label{eq:sage}
\mathbf{h}_v^{(\ell)} = \sigma \left(\mathbf{h}_v^{(\ell-1)} \mathbf{W_1}^{(\ell)}\left| \right| \mathbf{k}_v^{(\ell-1)}\mathbf{W_2}^{(\ell)}\right)
\end{equation}
Hence, when applied to the generic framework, $K=2$, i.e. there is only two learnable weight matrix per layer. Now, to compute $\cW^{\ell}_{k,M}$ using Eq.~\ref{eq:cfw}, we need to know $\cG^{\ell}_{k,i}=\{\cg_{v,k,i}^{\ell}\mid v\in\CV\}$ and $\CZ^{\ell-1}_{k,i}=\{\cz_{v,k,i}^{\ell-1}\mid v\in\CV\}$. From Eq.~\ref{eq:sage}, it is easy to see that for any GraphSAGE model $\Theta_i$, we have:
\begin{align}    
 \mathbf{g}_{v,1,i}^{\ell} = \underbrace{\left(\mathbf{h_{v,i}^{(\ell-1)}} \right)}_{\cz_{v,1,i}^{\ell-1}}\mathbf{W}^{(\ell)}_{1,i}    \\
 \mathbf{g}_{v,2,i}^{\ell} = \underbrace{\left(\mathbf{k_{v,i}^{(\ell-1)}} \right)}_{\cz_{v,2,i}^{\ell-1}}\mathbf{W}^{(\ell)}_{2,i}    
\end{align}

\subsection{Graph Isomorphism Network(GIN)}
The node embedding update equation for \textbf{GIN} is as follows:
\[
\mathbf{h}_v^{(\ell)} = \boldsymbol{\phi}^{(\ell)}\left(\mathbf{h}_v^{(\ell-1)}+\sum_{u \in \mathcal{N}_v} \mathbf{h}_u^{(\ell-1)} \right)
\]
where:
\begin{itemize}
    \item \(\mathbf{h}_v^{(\ell)}\): Updated feature vector of node \(i\) at layer \(\ell\).
    \item \(\mathcal{N}_v\): Set of neighbors of node \(v\).
    \item \(\mathbf{h}_u^{(\ell-1)}\): Feature vector of node \(u\) at layer $\ell$.
    \item \(\boldsymbol{\phi}^{(\ell)}\): Trainable \textbf{MLP} at layer $\ell$.
\end{itemize}
Here, every node collections messages from its neighbours, as well as itself, and takes the sum of the messages. The term $\left(\mathbf{h}_v^{(\ell-1)}+\sum_{u \in \mathcal{N}_v} \mathbf{h}_u^{(\ell-1)} \right)$ can be computed independent of $\boldsymbol{\phi}^{(\ell)}$ and can be denoted by $\mathbf{k}_v^{(\ell-1)}$. Hence, the node update equation becomes:
\[
\mathbf{h}_v^{\ell} = \boldsymbol{\phi}^{(\ell)}(\mathbf{k}_v^{(\ell-1)})
\]
The node update equation is just an MLP applied on $\mathbf{k}_v^{(\ell-1)}$. A typical $N$ layer MLP is as follows:
\begin{equation}
\label{eq:gin_mlp}
\mathbf{y} = \text{MLP}^{(N)}(\mathbf{x}) = f_N(\cW_N \cdot f_{N-1}(\cW_{N-1} \cdot \dots f_1(\cW_1 \cdot \mathbf{x})))
\end{equation}
\text{Where:} \\
$\mathbf{x}$ is the input vector, \\
$\mathbf{W}_i$ are the weight matrices for each layer, \\
$\mathbf{f}_i$ are the activation functions for each layer, and \\
$\mathbf{y}$ is the output. \\
The operation at layer $\mathbf{n}$ is just a linear transform $\mathbf{W}_n$ followed by an activation function $\mathbf{f}_n$. Hence, the MLP can be broken down into a series of linear transforms. \\
Hence, when applied to the generic framework, $K = N$, i.e, there are $N$ learnable weight matrices per layer. Now, to compute $\cW^{\ell}_{n,M}$ using Eq.~\ref{eq:cfw}, we need to know $\cG^{\ell}_{n,i}=\{\cg_{v,n,i}^{\ell}\mid v\in\CV\}$ and $\CZ^{\ell-1}_{n,i}=\{\cz_{v,n,i}^{\ell-1}\mid v\in\CV\}$. \\
$\cG^{\ell}_{1,i}$ and $\CZ^{\ell-1}_{n,i}$ can simply be obtained by :
\begin{equation}
    \mathbf{g}_{v,1,i}^{\ell} = \underbrace{\left(\mathbf{k_{v,i}^{(\ell-1)}} \right)}_{\cz_{v,1,i}^{\ell-1}}\mathbf{W}^{(\ell)}_{1,i} 
\end{equation}
From eq~\ref{eq:gin_mlp}, we can write $\mathbf{g}_{v,n,i}^{\ell}$ inductively as
\begin{equation}
    \mathbf{g}_{v,n,i}^{\ell} = \underbrace{\left(f_{n-1}\left(\mathbf{g}_{v,n-1,i}^{\ell} \right)\right)}_{\cz_{v,n,i}^{\ell-1}}\mathbf{W}^{(\ell)}_{n,i} 
\end{equation}
to obtain $\cG^{\ell}_{n,i}=\{\cg_{v,n,i}^{\ell}\mid v\in\CV\}$ and $\CZ^{\ell-1}_{n,i}=\{\cz_{v,n,i}^{\ell-1}\mid v\in\CV\}$.
\subsection{Graph Attention Network(GAT)}
The node embedding update equation for \textbf{GAT} before activation is as follows:
\[
\mathbf{h}_v^{(\ell)} = \sigma\left( \sum_{u \in \mathcal{N}_v} \alpha_{uv} \mathbf{h}_u^{(\ell-1)}\mathbf{W}^{(\ell)}  \right)
\]
where:
\begin{itemize}
    \item \(\mathcal{N}_v\): Set of neighbors of node \(v\).
    \item \(\mathbf{h}_u^{(\ell)}\): Feature vector of node \(u\) at layer $\ell$.
    \item \(\mathbf{W}^{(\ell)}\): Trainable weight matrix at layer $\ell$.
    \item \(\alpha_{vu}\) : Attention coefficients between node $v$ and its neighbour node $u$.
\end{itemize}
The attention coefficients $\alpha_{vu}$ are computed using the attention mechanism. typically involving a self-attention mechanism such as:
\begin{equation}
\alpha_{uv} = \frac{\exp \left( \text{LeakyReLU} \left( \mathbf{a^{(\ell)}}^{\mathtt{T}} [\mathbf{W}^{(\ell)} \mathbf{h}_v^{(\ell-1)} \parallel \mathbf{W}^{(\ell)} \mathbf{h}_u^{(\ell-1)}] \right) \right)}{\sum_{k \in \mathcal{N}_v} \exp \left( \text{LeakyReLU} \left( \mathbf{a^{(\ell)}}^{\mathtt{T}}[\mathbf{W}^{(\ell)} \mathbf{h}_v^{(\ell-1)} \parallel \mathbf{W}^{(\ell)} \mathbf{h}_k^{(\ell-1)}] \right) \right)}
\end{equation}
which involves a learnable vector $\mathbf{a^{(\ell)}}$.\\
Hence, when applied to the generic framework, $K=2$, i.e, there are $2$ learnable weight matrices per layer. \\
For $\mathbf{W}^{(\ell)}_M$, we simply have:
\begin{equation}
    \mathbf{g}_{v,i}^{\ell} = \underbrace{\left( \sum_{u \in \mathcal{N}_v} \alpha_{uv,i}^{\ell} \mathbf{h}_{u,i}^{(\ell-1)} \right)}_{\cz_{v,i}^{\ell-1}}\mathbf{W}^{(\ell)}_{i}  
\end{equation}
where, $\alpha_{uv,i}^{\ell}$ is computed as:
\[
    \alpha_{uv,i}^{\ell} = \frac{\exp \left( \text{LeakyReLU} \left( \mathbf{a_i^{(\ell)}}^{\mathtt{T}} [\mathbf{W}_{i}^{(\ell)} \mathbf{h}_{v,i}^{(\ell-1)} \parallel \mathbf{W}_{i}^{(\ell)} \mathbf{h}_{u,i}^{(\ell-1)}] \right) \right)}{\sum_{k \in \mathcal{N}_v} \exp \left( \text{LeakyReLU} \left( \mathbf{a_{i}^{(\ell)}}^{\mathtt{T}}[\mathbf{W}_{i}^{(\ell)} \mathbf{h}_{v,i}^{(\ell-1)} \parallel \mathbf{W}_{i}^{(\ell)} \mathbf{h}_{k,i}^{(\ell-1)}] \right) \right)}
\]
For $\mathbf{a_M^{(\ell)}}$, we have:
\[
    \mathbf{g}_{uv,i}^{\ell} = \underbrace{\left( [\mathbf{W}_{i}^{(\ell)} \mathbf{h}_{v,i}^{(\ell-1)} \parallel \mathbf{W}_{i}^{(\ell)} \mathbf{h}_{u,i}^{(\ell-1)}] \right)}_{\cz_{uv,i}^{\ell-1}}\mathbf{a}^{(\ell)}_{i}
\]
\subsection{NodeFormer}
NodeFormer follows the general idea of Queries, Keys, and Values present in Transformers. In each transformer layer, we have the $\mathbf{W_{Q}^{\ell}}$, $\mathbf{W_{K}^{\ell}}$ and $\mathbf{W_{V}^{\ell}}$ matrices that are used to compute queries, keys and values for each node as:
\begin{align*}
\mathbf{q_v^{\ell}} &= \mathbf{W_{Q}^{\ell}}\mathbf{z}_v^{\ell-1} \\ 
\mathbf{k_v^{\ell}} &= \mathbf{W_{K}^{\ell}}\mathbf{z}_v^{\ell-1} \\ 
\mathbf{v_v^{\ell}} &= \mathbf{W_{V}^{\ell}}\mathbf{z}_v^{\ell-1} 
\end{align*}
where $\mathbf{z}_v^{\ell-1}$ is the node embedding produced by the previous layer. Additionally, it also has a $\mathbf{W}_{O}^{\ell}$ which is used to aggregate the results of multiple heads to obtain the final node embedding for the layer $\ell$ as follows:
\begin{align*}
\mathbf{z_v^{\ell}} &= \mathbf{W_{O}^{\ell}}\mathbf{z'}_v^{\ell} 
\end{align*}
where $\mathbf{z'}_v^{\ell-1}$ is obtained by applying attention pooling using $\mathbf{q_v^{\ell}}$, $\mathbf{k_v^{\ell}}$ and $\mathbf{v_v^{\ell}}$, according to NodeFormer equation:
\[
    \mathbf{z'}_v^{\ell} = \sum_{u=1}^{|V|} \left(\frac{\kappa\left(\mathbf{q_v^{\ell}},\mathbf{k_u^{\ell}}\right)}{\sum_{w=1}^{|V|} \kappa\left(\mathbf{q_v^{\ell}},\mathbf{k_w^{\ell}}\right)}\right)\mathbf{v_u^{\ell}}
\]
where, $\kappa$ is a kernel measuring pairwise similarity.
All of $\mathbf{W}_{Q,M}^{\ell}$, $\mathbf{W}_{K,M}^{\ell}$, $\mathbf{W}_{V,M}^{\ell}$ and $\mathbf{W}_{O,M}^{\ell}$ can be computed analytically using similar formulation as discussed above for MPNNs.
\section{Merging Heterogeneous Architectures}
\subsection{Different MPNNs}
Consider merging a \gat and \sage into a \gcn.\\
~\label{app:gatsagetogcn}
The node embedding update equation for \textbf{GAT} before activation is as follows:
\[
\mathbf{h}_v^{(\ell)} = \sigma\left( \sum_{u \in \mathcal{N}_v} \alpha_{uv} \mathbf{h}_u^{(\ell-1)}\mathbf{W}^{(\ell)}  \right)
\]
where:
\begin{itemize}
    \item \(\mathcal{N}_v\): Set of neighbors of node \(v\).
    \item \(\mathbf{h}_u^{(\ell)}\): Feature vector of node \(u\) at layer $\ell$.
    \item \(\mathbf{W}^{(\ell)}\): Trainable weight matrix at layer $\ell$.
    \item \(\alpha_{vu}\) : Attention coefficients between node $v$ and its neighbour node $u$.
\end{itemize}
The node embedding update equation for \textbf{GraphSAGE} is as follows:
\begin{equation}
\mathbf{h}_v^{(\ell)} = \sigma \left(\mathbf{h}_v^{(\ell-1)} \mathbf{W_1}^{(\ell)} \left| \right| \sum_{u \in \mathcal{N}_v} \frac{1}{\lvert \mathcal{N}_v \rvert} \mathbf{h}_u^{(\ell-1)} \mathbf{W_2}^{(\ell)}\right)
\end{equation}
where:
\begin{itemize}
    \item \(\mathcal{N}_v\): Set of neighbors of node \(i\).
    \item \(\mathbf{h}_u^{(\ell)}\): Feature vector of node \(u\) at layer $\ell$.
    \item $\sigma$: Activation function like ReLU. 
    \item \(\mathbf{W_1}^{(\ell)},\mathbf{W_2}^{(\ell)}\): Trainable weight matrices at layer $\ell$.
\end{itemize}

Ultimately, both of these update the node embeddings by means of message passing and aggregation. If the first model to be merged is a \gat, and the second model is a \sage; before activation, we have:
\[
\mathbf{g}_{v,1}^{(\ell)} =  \sum_{u \in \mathcal{N}_v} \alpha_{uv} \mathbf{h}_{u,1}^{(\ell-1)}\mathbf{W}^{(\ell)}  
\]
\begin{equation}
\mathbf{g}_{v,2}^{(\ell)} = \mathbf{h}_{v,2}^{(\ell-1)} \mathbf{W_1}^{(\ell)} \left| \right| \sum_{u \in \mathcal{N}_v} \frac{1}{\lvert \mathcal{N}_v \rvert} \mathbf{h}_{u,2}^{(\ell-1)} \mathbf{W_2}^{(\ell)}
\end{equation}
These are the target embeddings that the merged \gcn must aim to align to, given the set of input embeddings $h_{v,1}^{\ell-1}$ for \gat and $h_{v,2}^{\ell-1}$ for \sage. Following Eq.~\ref{eq:gcneq}, \gcn computes target embeddings as:
\begin{equation}
\label{eq:gcneq}
 \mathbf{g'}_{v,i}^{\ell} = \underbrace{\left(\sum_{j \in \mathcal{N}_v \cup \{i\}} \frac{1}{\sqrt{d_v d_u}} \mathbf{h}_{u,i}^{(\ell-1)} \right)}_{\cz_{v,i}^{\ell-1}}\mathbf{W}^{(\ell)}    
\end{equation}
which must be aligned with $g_{v,1}^{\ell}$ and $g_{v,2}^{\ell}$ for inputs $h_{v,1}^{\ell-1}$ and $h_{v,2}^{\ell-1}$. Let $\mathbf{Z_{i}^{\ell-1}}$ be the matrix containing $\cz_{v,i}^{\ell-1}$, and $\mathbf{G_{i}^{\ell}}$ be the matrix containing $\cg_{v,i}^{\ell}$, where $i=1$ corresponds to the \gat, and $i=2$ corresponds to the \sage. $\mathbf{W^{\ell}}$ can then be optimised using Eq.~\ref{eq:cfw}.
\subsection{Different Hidden Dimensions}
~\label{app:hidden}
Consider the scenario of merging layers with differing input and output dimensions. We reduce this to a combination of two simpler cases:
    \begin{enumerate}
        \item \textbf{Different Input Dimensions: }Consider merging two heterogeneous model layers parameterized by $\cW_1\in\mathbb{R} ^{d\times d_1}$ and $\cW_2\in\mathbb{R}^ {d\times d_2}$, with $d_2 > d_1$. They can be merged into a $\cW\in\mathbb{R}^{d \times d_2}$ by zero-padding the $d_1$ dimensional inputs $\CZ_1$ to $\cW_1$ to make $d_2$ dimensional inputs $\CZ_1'$. The merged matrix $\cW$ is then optimized by minimizing the following objective:
        \begin{equation}
            \|\CZ_1'\cW -  \CZ_1\cW_1\|_F^2 + \|\CZ_2\cW -  \CZ_2\cW_2\|_F^2
        \end{equation}
        \item \textbf{Different Output Dimensions: }Consider merging $\cW_1\in\mathbb{R}^{d_1\times d}$ and $\cW_2\in\mathbb{R}^{d_2\times d}$, with $d_2 > d_1$. They can be merged into a $\cW\in\mathbb{R}^{d_2 \times d}$ by aligning only the first $d_1$ dimensions of output of $\cW$ with output of $\cW_1$, as follows:
        \begin{equation}
            \|(\CZ_1\cW) [:d_1] -  \CZ_1\cW_1\|_F^2 + \|\CZ_2\cW -  \CZ_2\cW_2\|_F^2
        \end{equation}
        Each term can be normalized by its dimensionality, to ensure scale invariance across different embedding sizes.
    \end{enumerate}

\begin{review}
\section{Evaluation Metrics for Forward Simulations}
\label{sec:appendix_metrics}
Once trained, the forward simulations generated by the merged force fields should closely match the ground truth in terms of both atomic structure and dynamics. To rigorously quantify this, we evaluate the models using four metrics based on forward Molecular Dynamics (MD) simulations \cite{mannan2025evaluatinguniversalmachinelearning}. These simulations are initialised from an arbitrary structure and run for $n$ steps utilising the learned force fields—a task for which the models are not explicitly trained.

\subsection{Structure Metrics}
We use two metrics to evaluate the proximity of the simulated structures to the ground truth \cite{egraff}. Both metrics rely on the Radial Distribution Function (RDF), $g(r)$, which concisely captures the structure simulated by a force field in one dimension by representing the local time-averaged density of atoms at a distance $r$ from a central atom.

\textbf{Wright's Factor (WF), $R_{\chi}$:} This metric represents the relative difference between the RDF of the ground truth atomic structure ($g_{\text{ref}}(r)$) and the structure obtained from the atomistic simulations employing the learned force fields ($g(r)$):
$$R_{\chi} = \sqrt{\frac{\sum_{i=1}^n (g(r_i) - g_{\text{ref}}(r_i))^2}{\sum_{i=1}^n (g_{\text{ref}}(r_i))^2}}$$
A force field is generally considered acceptable if it yields a WF of less than 9\% for bulk systems \cite{10.1063/1.4886421}.

\textbf{Jensen-Shannon Divergence (JSD):} JSD is a robust tool for quantifying the similarity between two probability distributions, overcoming the asymmetry limitations of the Kullback-Leibler (KL) Divergence. Since the RDF is essentially a distribution of atomic density, the JSD between the predicted and ground-truth RDFs is computed as:
$$JSD(g(r) \parallel g_{\text{ref}}(r)) = \frac{1}{2} \left( KL(g(r) \parallel \hat{g}(r)) + KL(g_{\text{ref}}(r) \parallel \hat{g}(r)) \right)$$
where $\hat{g}(r) = \frac{1}{2}(g(r) + g_{\text{ref}}(r))$ is the mean of the predicted and ground-truth RDF distributions.

\subsection{Dynamics Metrics}
To evaluate how closely the predicted dynamics reflect the ground truth, we monitor the energy and force errors over the forward simulation trajectory. We define the Energy Violation error ($EV(t)$) and Force Violation error ($FV(t)$) as:
$$EV(t) = \frac{(\hat{E}(t) - E(t))^2}{\hat{E}(t)^2 + E(t)^2 + \epsilon}, \quad FV(t) = \frac{\|\hat{F}(t) - F(t)\|_2^2}{\|\hat{F}(t)\|_2^2 + \|F(t)\|_2^2 + \epsilon}$$
where $\hat{E}(t)$ and $E(t)$ are the predicted and ground truth energies, respectively, and $\hat{F}(t)$ and $F(t)$ are the predicted and ground truth forces. A small numerical constant $\epsilon$ is added to avoid division-by-zero instability. This normalization ensures that the violation errors are dimensionless and bounded between 0 and 1, where 0 represents exact agreement with the ground truth and 1 represents total divergence. Finally, we compute the geometric mean of $EV(t)$ and $FV(t)$ over the full trajectory to represent the cumulative energy and force violations for the rollout.

\subsection{Rollout Results}
Tables \ref{tab:rollout_ev_fv_m3gnet} through \ref{tab:rollout_jsd_wf_orb} detail these rollout metrics.  LiPS20 is
omitted from these rollout metrics results as no OOD data was available for this dataset. We notice Orb models have an unusually high Force Violation error (FV). As detailed in Table \ref{tab:rollout_fv_orb}, this elevated error is a systemic trend across all tested datasets (MD17, rmd17, and MD22) and scenarios, regardless of the specific merging technique or fine-tuning baseline applied. This leads us to believe that the high force violation rate is likely an inherent limitation of the Orb architecture's rollout stability in these specific environments, rather than a degradation introduced by the model merging strategies themselves

\begin{table}[htbp]
    \centering
    \caption{Rollout cumulative EV/FV comparison across domains on M3GNet. Values are averages over five forward simulations for 1000 time steps with different initial conditions; parentheses show the standard deviation.}
    \label{tab:rollout_ev_fv_m3gnet}
    \resizebox{\textwidth}{!}{
    \begin{tabular}{l|cc|cc|cc|cc|cc|cc|cc}
    \toprule
    \multirow{2}{*}{\textbf{Scenario}} & \multicolumn{2}{c|}{\textbf{Wt Avg}} & \multicolumn{2}{c|}{\textbf{Fisher}} & \multicolumn{2}{c|}{\textbf{TIES}} & \multicolumn{2}{c|}{\textbf{EMR}} & \multicolumn{2}{c|}{\textbf{GFFMerge}} & \multicolumn{2}{c|}{\textbf{Joint FT SB}} & \multicolumn{2}{c}{\textbf{Joint FT}} \\
     & EV & FV & EV & FV & EV & FV & EV & FV & EV & FV & EV & FV & EV & FV \\
    \midrule
    \multicolumn{15}{c}{\cellcolor{green!10}\textbf{MD17}\textit{(kcal/mol/\AA{}, kcal/mol)}} \\
    Aspirin+Uracil & 1.6e-9 & 0.54 & 8.7e-10 & 0.56 & 2.0e-8 & 0.70 & 3.4e-9 & 0.65 & \textbf{8.6e-11} & \textbf{0.50} & 8.7e-11 & 0.57 & 8.7e-11 & 0.49 \\
    Eth+Mal+Asp & 1.1e-8 & 0.68 & 1.6e-8 & 0.67 & 1.2e-8 & 0.67 & 2.3e-9 & 0.72 & \textbf{2.8e-10} & \textbf{0.59} & 2.6e-10 & 0.61 & 2.7e-10 & 0.56 \\
    5-Task Mix & 1.3e-8 & 0.65 & 1.4e-8 & 0.65 & 1.5e-8 & 0.66 & 2.9e-8 & 0.75 & \textbf{2.1e-10} & \textbf{0.56} & 2.0e-10 & 0.55 & 2.1e-10 & 0.52 \\
    \midrule
    \multicolumn{15}{c}{\cellcolor{blue!10}\textbf{MD22}\textit{(kcal/mol/\AA{}, kcal/mol)}} \\
    Ac-Ala3+AT-AT & 7.3e-9 & 0.61 & 1.1e-8 & 0.68 & 4.4e-8 & 0.79 & 4.5e-9 & 0.72 & \textbf{3.5e-11} & \textbf{0.58} & 3.6e-11 & 0.61 & 3.8e-11 & 0.48 \\
    Ac-Ala3+DHA+Stach & 1.2e-8 & 0.71 & 1.3e-8 & 0.68 & 3.3e-9 & 0.81 & 1.5e-7 & 0.83 & \textbf{4.0e-11} & \textbf{0.64} & 4.6e-11 & 0.63 & 4.5e-11 & 0.54 \\
    \bottomrule
    \end{tabular}
    }
\end{table}

\begin{table}[htbp]
    \centering
    \caption{Rollout cumulative Force Violation (FV) Error on Orb.}
    \label{tab:rollout_fv_orb}
    \resizebox{\textwidth}{!}{
    \begin{tabular}{l|c|c|c|c|c|c|c}
    \toprule
    \textbf{Scenario} & \textbf{Wt Avg} & \textbf{Fisher} & \textbf{TIES} & \textbf{EMR} & \textbf{GFFMerge} & \textbf{Joint FT SB} & \textbf{Joint FT} \\
    \midrule
    \multicolumn{8}{c}{\cellcolor{green!10}\textbf{MD17}\textit{(kcal/mol/\AA{}, kcal/mol)}} \\
    Aspirin + Uracil & 0.984 & 0.984 & 0.965 & 0.978 & 0.997 & 0.971 & 0.970 \\
    Ethanol + Malonaldehyde + Aspirin & 0.939 & 0.942 & 0.931 & 0.965 & 0.997 & 0.973 & 0.974 \\
    5-Task Mix & 0.951 & 0.950 & 0.952 & 0.965 & 0.997 & 0.964 & 0.968 \\
    \midrule
    \multicolumn{8}{c}{\cellcolor{blue!10}\textbf{MD22}\textit{(kcal/mol/\AA{}, kcal/mol)}} \\
    Ac-Ala3-NHMe + AT-AT & 0.979 & 0.979 & 0.964 & 0.984 & 0.999 & 0.978 & 0.984 \\
    Ac-Ala3-NHMe + DHA + Stachyose & 0.982 & 0.979 & 0.975 & 0.983 & 0.999 & 0.983 & 0.985 \\
    \bottomrule
    \end{tabular}
    }
\end{table}

\begin{table}[htbp]
    \centering
    \caption{Rollout RDF Jensen-Shannon Divergence (JSD) and Wright's Factor (WF) comparison across domains on M3GNet. Lower JSD and higher WF represent distributions closer to the ground truth.}
    \label{tab:rollout_jsd_wf_m3gnet}
    \resizebox{\textwidth}{!}{
    \begin{tabular}{l|cc|cc|cc|cc|cc|cc|cc}
    \toprule
    \multirow{2}{*}{\textbf{Scenario}} & \multicolumn{2}{c|}{\textbf{Wt Avg}} & \multicolumn{2}{c|}{\textbf{Fisher}} & \multicolumn{2}{c|}{\textbf{TIES}} & \multicolumn{2}{c|}{\textbf{EMR}} & \multicolumn{2}{c|}{\textbf{GFFMerge}} & \multicolumn{2}{c|}{\textbf{Joint FT SB}} & \multicolumn{2}{c}{\textbf{Joint FT}} \\
     & JSD & WF & JSD & WF & JSD & WF & JSD & WF & JSD & WF & JSD & WF & JSD & WF \\
    \midrule
    \multicolumn{15}{c}{\cellcolor{green!10}\textbf{MD17}\textit{(kcal/mol/\AA{}, kcal/mol)}} \\
    Asp+Ura & 0.0041 & 10.67 & 0.0046 & 10.96 & 0.0059 & \textbf{13.46} & \textbf{0.0040} & 10.88 & 0.0042 & 10.71 & 0.0042 & 10.79 & 0.0040 & 9.85 \\
    Eth+Mal+Asp & 0.0094 & 18.20 & 0.0105 & 18.18 & 0.0111 & 19.30 & 0.0102 & \textbf{19.92} & \textbf{0.0090} & 15.91 & 0.0093 & 17.22 & 0.0101 & 18.62 \\
    5-Task Mix & 0.0069 & 13.74 & \textbf{0.0060} & 12.59 & 0.0086 & \textbf{16.72} & 0.0081 & 15.27 & 0.0066 & 13.29 & 0.0064 & 12.95 & 0.0072 & 13.38 \\
    \midrule
    \multicolumn{15}{c}{\cellcolor{blue!10}\textbf{MD22}\textit{(kcal/mol/\AA{}, kcal/mol)}} \\
    Ac-Ala3+AT & 0.0033 & 6.73 & 0.0023 & 7.11 & 0.0037 & \textbf{12.12} & 0.0043 & 7.07 & \textbf{0.0023} & 5.48 & 0.0020 & 5.58 & 0.0013 & 5.41 \\
    Ac-Ala3+DHA & 0.0010 & 5.47 & 0.0011 & 5.70 & 0.0014 & 6.67 & 0.0016 & \textbf{7.77} & \textbf{0.0007} & 3.67 & 0.0012 & 4.67 & 0.0006 & 3.75 \\
    \bottomrule
    \end{tabular}
    }
\end{table}

\begin{table}[htbp]
    \centering
    \caption{Rollout RDF Jensen-Shannon Divergence (JSD) and Wright's Factor (WF) comparison across domains on Orb.}
    \label{tab:rollout_jsd_wf_orb}
    \resizebox{\textwidth}{!}{
    \begin{tabular}{l|cc|cc|cc|cc|cc|cc|cc}
    \toprule
    \multirow{2}{*}{\textbf{Scenario}} & \multicolumn{2}{c|}{\textbf{Wt Avg}} & \multicolumn{2}{c|}{\textbf{Fisher}} & \multicolumn{2}{c|}{\textbf{TIES}} & \multicolumn{2}{c|}{\textbf{EMR}} & \multicolumn{2}{c|}{\textbf{GFFMerge}} & \multicolumn{2}{c|}{\textbf{Joint FT SB}} & \multicolumn{2}{c}{\textbf{Joint FT}} \\
     & JSD & WF & JSD & WF & JSD & WF & JSD & WF & JSD & WF & JSD & WF & JSD & WF \\
    \midrule
    \multicolumn{15}{c}{\cellcolor{green!10}\textbf{MD17}\textit{(kcal/mol/\AA{}, kcal/mol)}} \\
    Asp+Ura & 0.0657 & 50.17 & 0.0615 & 49.19 & 0.0427 & \textbf{54.69} & 0.0797 & 48.95 & \textbf{0.0334} & 48.91 & 0.0507 & 44.50 & 0.0660 & 48.38 \\
    Eth+Mal+Asp & 0.0706 & \textbf{56.89} & 0.0462 & 50.42 & 0.0569 & 54.23 & 0.0803 & 47.75 & \textbf{0.0496} & 69.74 & 0.0676 & 52.84 & 0.0598 & 50.16 \\
    5-Task Mix & 0.0507 & 50.72 & 0.0741 & \textbf{53.96} & 0.0563 & 47.65 & 0.0591 & 43.19 & \textbf{0.0266} & 50.97 & 0.0653 & 47.36 & 0.0651 & 44.34 \\
    \midrule
    \multicolumn{15}{c}{\cellcolor{blue!10}\textbf{MD22}\textit{(kcal/mol/\AA{}, kcal/mol)}} \\
    Ac-Ala3+AT & 0.0208 & 41.67 & 0.0244 & 38.12 & 0.0260 & \textbf{50.19} & 0.0213 & 29.64 & \textbf{0.0217} & 38.93 & 0.0235 & 38.84 & 0.0211 & 37.12 \\
    Ac-Ala3+DHA & 0.0283 & 41.68 & 0.0306 & 40.40 & 0.0297 & \textbf{45.43} & 0.0246 & 33.67 & \textbf{0.0146} & 35.62 & 0.0281 & 38.12 & 0.0192 & 36.67 \\
    \bottomrule
    \end{tabular}
    }
\end{table}
\end{review}

\section{Experimental Setup}
\label{app:setup}
\subsection{Hardware Configuration}
All experiments were conducted on a high-performance computing system with the following specifications:
\begin{itemize}
    \item \textbf{CPU:} 96 logical cores
    \item \textbf{RAM:} 512 GB
    \item \textbf{GPU:} NVIDIA A100-PCIE-40GB
\end{itemize}

\subsection{Software Configuration}
The software environment for our experiments was configured as follows:
\begin{itemize}
    \item \textbf{Operating System:} Linux (Ubuntu 20.04.4 LTS (GNU/Linux 5.4.0-124-generic x86\_64))
    \item \textbf{PyTorch Version:} 1.13.1+cu117
    \item \textbf{CUDA Version:} 11.7
    \item \textbf{PyTorch Geometric Version:} 2.3.1
\end{itemize}

\subsection{Parameters used for \nameff}

\subparagraph{Orb}
\begin{itemize}
    \item Variant: Orb-d3-xs-v2
    \item All experiments are conducted using force only Orb models.
    \item We fine tune last 3 interaction blocks and force head during Targeted Fine tuning phase.
    \item Train/Val/Test split is maintained 8:1:1
    \item All experiments are run 3 times with varying training seeds (42,43,45)
    \item Optimizer: Adam
    \item Number of samples used (per molecule): 1250
\end{itemize}

\subparagraph{M3GNet}
\begin{itemize}
    \item Variant used for md17 and md22 datasets merging: M3GNet-ANI-1x-Subset-PES 
    \item Variant used for Lips20 datasets merging: M3GNet-MatPES-PBE-v2025.1-PES 
    \item  We fine tune last 3 interaction blocks and energy readout during Targeted Fine tuning phase.
    \item For md17, md22 experiments, force weight is fixed to 0.1; For Lips20 experiments, force weight is fixed to 10
    \item Train/Val/Test split is maintained 8:1:1
    \item All experiments are run 3 times with varying training seeds (42,43,45)
    \item Optimizer: Adam
    \item Number of samples used (per molecule): 1250
\end{itemize}

\subsubsection{Additional Experimental Details}
\begin{itemize}
    \item For Fisher and EMR merging baseline, the default hyperparameters provided in the source code were used.
    \item EMR-Merging departs from traditional approaches by creating one base model through selecting maximum absolute parameter values from sign-aligned task vectors. It then generates task-specific masks and rescalers that modulate this base, effectively producing multiple task-specific models from a single unified model. The tables show performance using the unified model.
    \item For TIES merging baseline, grid search was used to find the best hyperparameters.
\end{itemize}

\subsection{Parameters used for \name}

\begin{itemize}
    \item Default \gnn : \gcn
    \item Default number of layers in \gnn: 2, with ReLU in between.
    \item Hidden Dimension: 128
    \item Learning rate: 0.05
    \item Optimizer: Adam
    \item For MAG240M and Papers100M, we used state-of-the-art OGBN leaderboard model : \sage with hidden-layers = 2 and hidden-dimensions = 1024. 
\end{itemize}

\subsubsection{Additional Experimental Details}
\begin{itemize}
    \item For the node classification tasks, we used the default train-val-test splits available with the respective datasets.
    \item For link prediction tasks, we generated a 70-10-20 train-val-test split using the RandomLinkSplit function in Pytorch. The ratio of positive to negative links was set to 1.0.
    \item The disjoint label splits were created by taking the nodes that belonged to the first $\frac{N}{2}$ classes in the first dataset, and the nodes belonging to the next $\frac{N}{2}$ classes in the second dataset. $N=$ total different classes in the dataset.
    \item For all the the baselines, the default hyperparameters provided in the source code were used.
\end{itemize}


\begin{table*}[t!]
    \centering
    \caption{\textbf{Predictive Fidelity across Domains (Mean $\pm$ Std. Dev.).} Detailed comparison of Weight Averaging, \nameff, and Joint Fine-Tuning with full precision.
    \textbf{Task Definitions:} 
    \textit{DHA}: Docosahexaenoic Acid.
    $^\ddagger$\textit{MD17 5-Task Mix}: Ethanol + Naphthalene + Salicylic Acid + Uracil + Aspirin. 
    $^\star$\textit{MD22 3-Task Mix}: Ac-Ala3-NHMe + DHA + Stachyose.}
    \label{tab:full_results_mean_std_wt}

    \tiny
    \setlength{\tabcolsep}{2.5pt}
    \renewcommand{\arraystretch}{1.1}

    \resizebox{\textwidth}{!}{
    \begin{tabular}{l|cc||cc|cc}
    \toprule
    \multirow{2}{*}{\textbf{Dataset \& Scenario}} & \multicolumn{2}{c||}{\textbf{Wt Averaging}} & \multicolumn{2}{c|}{\nameff} & \multicolumn{2}{c}{\textbf{Joint FT}} \\
     & F & E & F & E & F & E \\
    \midrule

    \multicolumn{7}{c}{\cellcolor{green!10}\textbf{MD17}\textit{(kcal/mol/\AA{}, kcal/mol)}} \\
    Aspirin + Uracil & $1.79434 \pm 0.00738$ & $0.81819 \pm 0.00300$ & $\mathbf{1.01342 \pm 0.00088}$ & $\mathbf{0.03342 \pm 0.00004}$ & $0.90018 \pm 0.00125$ & $0.03454 \pm 0.00458$ \\
    Ethanol + Malonaldehyde & $1.90226 \pm 0.00277$ & $1.59925 \pm 0.00161$ & $\mathbf{0.73304 \pm 0.00216}$ & $\mathbf{0.03321 \pm 0.00047}$ & $0.64621 \pm 0.00434$ & $0.02887 \pm 0.00169$ \\
    Naphthalene + Salicylic Acid & $1.67235 \pm 0.00138$ & $0.58528 \pm 0.00069$ & $\mathbf{0.81953 \pm 0.00028}$ & $\mathbf{0.02447 \pm 0.00009}$ & $0.73867 \pm 0.00261$ & $0.02417 \pm 0.00084$ \\
    Ethanol + Malon. + Aspirin & $2.52720 \pm 0.00346$ & $2.42043 \pm 0.00231$ & $\mathbf{0.89161 \pm 0.00515}$ & $\mathbf{0.03566 \pm 0.00153}$ & $0.77565 \pm 0.00307$ & $0.03040 \pm 0.00191$ \\
    Naph. + Salicylic + Uracil & $2.05469 \pm 0.00715$ & $1.30453 \pm 0.00415$ & $\mathbf{0.86592 \pm 0.00253}$ & $\mathbf{0.02544 \pm 0.00058}$ & $0.72498 \pm 0.00327$ & $0.02733 \pm 0.00385$ \\
    \textit{5-Task Mix (MD17)$^\ddagger$} & $2.39829 \pm 0.00184$ & $2.36140 \pm 0.00208$ & $\mathbf{0.88912 \pm 0.00078}$ & $\mathbf{0.03052 \pm 0.00016}$ & $0.72327 \pm 0.00445$ & $0.02911 \pm 0.00208$ \\
    \midrule

    \multicolumn{7}{c}{\cellcolor{blue!10}\textbf{MD22}\textit{(kcal/mol/\AA{}, kcal/mol)}} \\
    Ac-Ala3-NHMe + AT-AT & $2.86734 \pm 0.00553$ & $1.95553 \pm 0.00807$ & $\mathbf{1.65223 \pm 0.00136}$ & $\mathbf{0.05278 \pm 0.00085}$ & $1.24499 \pm 0.00655$ & $0.04033 \pm 0.00622$ \\
    DHA + Stachyose & $3.34239 \pm 0.00484$ & $1.88243 \pm 0.02075$ & $\mathbf{1.69773 \pm 0.00705}$ & $\mathbf{0.03913 \pm 0.00319}$ & $1.48247 \pm 0.01104$ & $0.04575 \pm 0.01417$ \\
    \textit{3-Task Mix (MD22)$^\star$} & $3.42933 \pm 0.01822$ & $2.07706 \pm 0.00692$ & $\mathbf{1.82968 \pm 0.00166}$ & $\mathbf{0.05670 \pm 0.00469}$ & $1.37621 \pm 0.01193$ & $0.03498 \pm 0.00119$ \\
    \midrule

    \multicolumn{7}{c}{\cellcolor{orange!10}\textbf{LiPS20}\textit{(eV/\AA{}, eV)}} \\
    $\beta$-Li$_3$PS$_4$ + $\gamma$-Li$_3$PS$_4$ & $51.78511 \pm 77.44589$ & $85.56390 \pm 23.83803$ & $\mathbf{0.72233 \pm 0.32170}$ & $\mathbf{0.15718 \pm 0.14543}$ & $0.43817 \pm 0.00597$ & $0.17530 \pm 0.12641$ \\
    Li$_3$P + Li$_2$S & $7.45958 \pm 1.24786$ & $48.45973 \pm 9.95105$ & $\mathbf{0.41767 \pm 0.05757}$ & $\mathbf{0.09184 \pm 0.04706}$ & $0.32156 \pm 0.00277$ & $0.06562 \pm 0.01921$ \\
    Li$_2$S + Li$_3$P + P$_2$S$_5$ & $7.07134 \pm 1.04311$ & $137.60602 \pm 11.09772$ & $\mathbf{1.21035 \pm 0.78647}$ & $\mathbf{0.86130 \pm 0.63587}$ & $0.34083 \pm 0.00482$ & $0.11110 \pm 0.05115$ \\
    \bottomrule
    \end{tabular}
    }
\end{table*}

\begin{table*}[t!]
    \centering
    \caption{\textbf{Predictive Fidelity across Domains (Mean $\pm$ Std. Dev.).} Detailed comparison of Fisher Merging, \nameff, and Joint Fine-Tuning with full precision.
    \textbf{Task Definitions:} 
    \textit{DHA}: Docosahexaenoic Acid.
    $^\ddagger$\textit{MD17 5-Task Mix}: Ethanol + Naphthalene + Salicylic Acid + Uracil + Aspirin. 
    $^\star$\textit{MD22 3-Task Mix}: Ac-Ala3-NHMe + DHA + Stachyose.}
    \label{tab:full_results_mean_std_fisher}

    \tiny
    \setlength{\tabcolsep}{2.5pt}
    \renewcommand{\arraystretch}{1.1}

    \resizebox{\textwidth}{!}{
    \begin{tabular}{l|cc||cc|cc}
    \toprule
    \multirow{2}{*}{\textbf{Dataset \& Scenario}} & \multicolumn{2}{c||}{\textbf{Fisher}} & \multicolumn{2}{c|}{\nameff} & \multicolumn{2}{c}{\textbf{Joint FT}} \\
     & F & E & F & E & F & E \\
    \midrule

    \multicolumn{7}{c}{\cellcolor{green!10}\textbf{MD17}\textit{(kcal/mol/\AA{}, kcal/mol)}} \\
    Aspirin + Uracil & $1.61400 \pm 0.00761$ & $1.04879 \pm 0.00876$ & $\mathbf{1.01342 \pm 0.00088}$ & $\mathbf{0.03342 \pm 0.00004}$ & $0.90018 \pm 0.00125$ & $0.03454 \pm 0.00458$ \\
    Ethanol + Malonaldehyde & $1.97029 \pm 0.00254$ & $2.23410 \pm 0.00484$ & $\mathbf{0.73304 \pm 0.00216}$ & $\mathbf{0.03321 \pm 0.00047}$ & $0.64621 \pm 0.00434$ & $0.02887 \pm 0.00169$ \\
    Naphthalene + Salicylic Acid & $1.66658 \pm 0.00161$ & $0.74163 \pm 0.00530$ & $\mathbf{0.81953 \pm 0.00028}$ & $\mathbf{0.02447 \pm 0.00009}$ & $0.73867 \pm 0.00261$ & $0.02417 \pm 0.00084$ \\
    Ethanol + Malon. + Aspirin & $2.26685 \pm 0.00969$ & $2.87265 \pm 0.00208$ & $\mathbf{0.89161 \pm 0.00515}$ & $\mathbf{0.03566 \pm 0.00153}$ & $0.77565 \pm 0.00307$ & $0.03040 \pm 0.00191$ \\
    Naph. + Salicylic + Uracil & $1.71847 \pm 0.00692$ & $1.54390 \pm 0.02906$ & $\mathbf{0.86592 \pm 0.00253}$ & $\mathbf{0.02544 \pm 0.00058}$ & $0.72498 \pm 0.00327$ & $0.02733 \pm 0.00385$ \\
    \textit{5-Task Mix (MD17)$^\ddagger$} & $2.22326 \pm 0.00461$ & $2.53043 \pm 0.00507$ & $\mathbf{0.88912 \pm 0.00078}$ & $\mathbf{0.03052 \pm 0.00016}$ & $0.72327 \pm 0.00445$ & $0.02911 \pm 0.00208$ \\
    \midrule

    \multicolumn{7}{c}{\cellcolor{blue!10}\textbf{MD22}\textit{(kcal/mol/\AA{}, kcal/mol)}} \\
    Ac-Ala3-NHMe + AT-AT & $3.24461 \pm 0.00254$ & $2.47093 \pm 0.00438$ & $\mathbf{1.65223 \pm 0.00136}$ & $\mathbf{0.05278 \pm 0.00085}$ & $1.24499 \pm 0.00655$ & $0.04033 \pm 0.00622$ \\
    DHA + Stachyose & $3.69475 \pm 0.07218$ & $2.58554 \pm 0.04958$ & $\mathbf{1.69773 \pm 0.00705}$ & $\mathbf{0.03913 \pm 0.00319}$ & $1.48247 \pm 0.01104$ & $0.04575 \pm 0.01417$ \\
    \textit{3-Task Mix (MD22)$^\star$} & $3.44893 \pm 0.02537$ & $2.26869 \pm 0.01845$ & $\mathbf{1.82968 \pm 0.00166}$ & $\mathbf{0.05670 \pm 0.00469}$ & $1.37621 \pm 0.01193$ & $0.03498 \pm 0.00119$ \\
    \midrule

    \multicolumn{7}{c}{\cellcolor{orange!10}\textbf{LiPS20}\textit{(eV/\AA{}, eV)}} \\
    $\beta$-Li$_3$PS$_4$ + $\gamma$-Li$_3$PS$_4$ & $18.85063 \pm 14.81913$ & $66.33108 \pm 77.40540$ & $\mathbf{0.72233 \pm 0.32170}$ & $\mathbf{0.15718 \pm 0.14543}$ & $0.43817 \pm 0.00597$ & $0.17530 \pm 0.12641$ \\
    Li$_3$P + Li$_2$S & $22.53662 \pm 21.18284$ & $34.60769 \pm 15.28153$ & $\mathbf{0.41767 \pm 0.05757}$ & $\mathbf{0.09184 \pm 0.04706}$ & $0.32156 \pm 0.00277$ & $0.06562 \pm 0.01921$ \\
    Li$_2$S + Li$_3$P + P$_2$S$_5$ & $19.51286 \pm 2.35257$ & $103.48033 \pm 3.34037$ & $\mathbf{1.21035 \pm 0.78647}$ & $\mathbf{0.86130 \pm 0.63587}$ & $0.34083 \pm 0.00482$ & $0.11110 \pm 0.05115$ \\
    \bottomrule
    \end{tabular}
    }
\end{table*}

\begin{table*}[t!]
    \centering
    \caption{\textbf{Predictive Fidelity across Domains (Mean $\pm$ Std. Dev.).} Detailed comparison of TIES Merging, \nameff, and Joint Fine-Tuning with full precision.
    \textbf{Task Definitions:} 
    \textit{DHA}: Docosahexaenoic Acid.
    $^\ddagger$\textit{MD17 5-Task Mix}: Ethanol + Naphthalene + Salicylic Acid + Uracil + Aspirin. 
    $^\star$\textit{MD22 3-Task Mix}: Ac-Ala3-NHMe + DHA + Stachyose.}
    \label{tab:full_results_mean_std_ties}

    \tiny
    \setlength{\tabcolsep}{2.5pt}
    \renewcommand{\arraystretch}{1.1}

    \resizebox{\textwidth}{!}{
    \begin{tabular}{l|cc||cc|cc}
    \toprule
    \multirow{2}{*}{\textbf{Dataset \& Scenario}} & \multicolumn{2}{c||}{\textbf{TIES}} & \multicolumn{2}{c|}{\nameff} & \multicolumn{2}{c}{\textbf{Joint FT}} \\
     & F & E & F & E & F & E \\
    \midrule

    \multicolumn{7}{c}{\cellcolor{green!10}\textbf{MD17}\textit{(kcal/mol/\AA{}, kcal/mol)}} \\
    Aspirin + Uracil & $2.99994 \pm 0.01199$ & $4.07917 \pm 0.00530$ & $\mathbf{1.01342 \pm 0.00088}$ & $\mathbf{0.03342 \pm 0.00004}$ & $0.90018 \pm 0.00125$ & $0.03454 \pm 0.00458$ \\
    Ethanol + Malonaldehyde & $3.76832 \pm 0.01384$ & $2.76196 \pm 0.01038$ & $\mathbf{0.73304 \pm 0.00216}$ & $\mathbf{0.03321 \pm 0.00047}$ & $0.64621 \pm 0.00434$ & $0.02887 \pm 0.00169$ \\
    Naphthalene + Salicylic Acid & $2.65426 \pm 0.01176$ & $3.88800 \pm 0.00992$ & $\mathbf{0.81953 \pm 0.00028}$ & $\mathbf{0.02447 \pm 0.00009}$ & $0.73867 \pm 0.00261$ & $0.02417 \pm 0.00084$ \\
    Ethanol + Malon. + Aspirin & $3.65348 \pm 0.00715$ & $2.52374 \pm 0.00092$ & $\mathbf{0.89161 \pm 0.00515}$ & $\mathbf{0.03566 \pm 0.00153}$ & $0.77565 \pm 0.00307$ & $0.03040 \pm 0.00191$ \\
    Naph. + Salicylic + Uracil & $2.76887 \pm 0.00254$ & $3.87532 \pm 0.00530$ & $\mathbf{0.86592 \pm 0.00253}$ & $\mathbf{0.02544 \pm 0.00058}$ & $0.72498 \pm 0.00327$ & $0.02733 \pm 0.00385$ \\
    \textit{5-Task Mix (MD17)$^\ddagger$} & $2.80969 \pm 0.00600$ & $2.85697 \pm 0.02237$ & $\mathbf{0.88912 \pm 0.00078}$ & $\mathbf{0.03052 \pm 0.00016}$ & $0.72327 \pm 0.00445$ & $0.02911 \pm 0.00208$ \\
    \midrule

    \multicolumn{7}{c}{\cellcolor{blue!10}\textbf{MD22}\textit{(kcal/mol/\AA{}, kcal/mol)}} \\
    Ac-Ala3-NHMe + AT-AT & $5.38071 \pm 0.01499$ & $4.97807 \pm 0.00461$ & $\mathbf{1.65223 \pm 0.00136}$ & $\mathbf{0.05278 \pm 0.00085}$ & $1.24499 \pm 0.00655$ & $0.04033 \pm 0.00622$ \\
    DHA + Stachyose & $4.61233 \pm 0.07126$ & $3.72220 \pm 0.04935$ & $\mathbf{1.69773 \pm 0.00705}$ & $\mathbf{0.03913 \pm 0.00319}$ & $1.48247 \pm 0.01104$ & $0.04575 \pm 0.01417$ \\
    \textit{3-Task Mix (MD22)$^\star$} & $5.12012 \pm 0.08486$ & $0.97892 \pm 0.01084$ & $\mathbf{1.82968 \pm 0.00166}$ & $\mathbf{0.05670 \pm 0.00469}$ & $1.37621 \pm 0.01193$ & $0.03498 \pm 0.00119$ \\
    \midrule

    \multicolumn{7}{c}{\cellcolor{orange!10}\textbf{LiPS20}\textit{(eV/\AA{}, eV)}} \\
    $\beta$-Li$_3$PS$_4$ + $\gamma$-Li$_3$PS$_4$ & $29.19796 \pm 31.64254$ & $160.82457 \pm 49.53846$ & $\mathbf{0.72233 \pm 0.32170}$ & $\mathbf{0.15718 \pm 0.14543}$ & $0.43817 \pm 0.00597$ & $0.17530 \pm 0.12641$ \\
    Li$_3$P + Li$_2$S & $4.03530 \pm 0.39264$ & $190.46968 \pm 8.61258$ & $\mathbf{0.41767 \pm 0.05757}$ & $\mathbf{0.09184 \pm 0.04706}$ & $0.32156 \pm 0.00277$ & $0.06562 \pm 0.01921$ \\
    Li$_2$S + Li$_3$P + P$_2$S$_5$ & $16.32979 \pm 8.91525$ & $186.28537 \pm 2.34361$ & $\mathbf{1.21035 \pm 0.78647}$ & $\mathbf{0.86130 \pm 0.63587}$ & $0.34083 \pm 0.00482$ & $0.11110 \pm 0.05115$ \\
    \bottomrule
    \end{tabular}
    }
\end{table*}

\begin{table*}[t!]
    \centering
    \caption{\textbf{Predictive Fidelity across Domains (Mean $\pm$ Std. Dev.).} Detailed comparison of EMR Merging, \nameff, and Joint Fine-Tuning with full precision.
    \textbf{Task Definitions:} 
    \textit{DHA}: Docosahexaenoic Acid.
    $^\ddagger$\textit{MD17 5-Task Mix}: Ethanol + Naphthalene + Salicylic Acid + Uracil + Aspirin. 
    $^\star$\textit{MD22 3-Task Mix}: Ac-Ala3-NHMe + DHA + Stachyose.}
    \label{tab:full_results_mean_std_emr}

    \tiny
    \setlength{\tabcolsep}{2.5pt}
    \renewcommand{\arraystretch}{1.1}

    \resizebox{\textwidth}{!}{
    \begin{tabular}{l|cc||cc|cc}
    \toprule
    \multirow{2}{*}{\textbf{Dataset \& Scenario}} & \multicolumn{2}{c||}{\textbf{EMR}} & \multicolumn{2}{c|}{\nameff} & \multicolumn{2}{c}{\textbf{Joint FT}} \\
     & F & E & F & E & F & E \\
    \midrule

    \multicolumn{7}{c}{\cellcolor{green!10}\textbf{MD17}\textit{(kcal/mol/\AA{}, kcal/mol)}} \\
    Aspirin + Uracil & $1.86813 \pm 0.00415$ & $1.17078 \pm 0.00392$ & $\mathbf{1.01342 \pm 0.00088}$ & $\mathbf{0.03342 \pm 0.00004}$ & $0.90018 \pm 0.00125$ & $0.03454 \pm 0.00458$ \\
    Ethanol + Malonaldehyde & $1.32967 \pm 0.00507$ & $0.52209 \pm 0.00346$ & $\mathbf{0.73304 \pm 0.00216}$ & $\mathbf{0.03321 \pm 0.00047}$ & $0.64621 \pm 0.00434$ & $0.02887 \pm 0.00169$ \\
    Naphthalene + Salicylic Acid & $1.66774 \pm 0.00069$ & $0.87976 \pm 0.00323$ & $\mathbf{0.81953 \pm 0.00028}$ & $\mathbf{0.02447 \pm 0.00009}$ & $0.73867 \pm 0.00261$ & $0.02417 \pm 0.00084$ \\
    Ethanol + Malon. + Aspirin & $2.26546 \pm 0.01960$ & $0.74854 \pm 0.00254$ & $\mathbf{0.89161 \pm 0.00515}$ & $\mathbf{0.03566 \pm 0.00153}$ & $0.77565 \pm 0.00307$ & $0.03040 \pm 0.00191$ \\
    Naph. + Salicylic + Uracil & $1.80817 \pm 0.00577$ & $1.65021 \pm 0.00254$ & $\mathbf{0.86592 \pm 0.00253}$ & $\mathbf{0.02544 \pm 0.00058}$ & $0.72498 \pm 0.00327$ & $0.02733 \pm 0.00385$ \\
    \textit{5-Task Mix (MD17)$^\ddagger$} & $2.16446 \pm 0.01199$ & $2.76472 \pm 0.01291$ & $\mathbf{0.88912 \pm 0.00078}$ & $\mathbf{0.03052 \pm 0.00016}$ & $0.72327 \pm 0.00445$ & $0.02911 \pm 0.00208$ \\
    \midrule

    \multicolumn{7}{c}{\cellcolor{blue!10}\textbf{MD22}\textit{(kcal/mol/\AA{}, kcal/mol)}} \\
    Ac-Ala3-NHMe + AT-AT & $3.48698 \pm 0.02491$ & $1.80840 \pm 0.02514$ & $\mathbf{1.65223 \pm 0.00136}$ & $\mathbf{0.05278 \pm 0.00085}$ & $1.24499 \pm 0.00655$ & $0.04033 \pm 0.00622$ \\
    DHA + Stachyose & $3.52803 \pm 0.01799$ & $2.13264 \pm 0.04750$ & $\mathbf{1.69773 \pm 0.00705}$ & $\mathbf{0.03913 \pm 0.00319}$ & $1.48247 \pm 0.01104$ & $0.04575 \pm 0.01417$ \\
    \textit{3-Task Mix (MD22)$^\star$} & $4.74677 \pm 0.06226$ & $7.35999 \pm 0.06457$ & $\mathbf{1.82968 \pm 0.00166}$ & $\mathbf{0.05670 \pm 0.00469}$ & $1.37621 \pm 0.01193$ & $0.03498 \pm 0.00119$ \\
    \midrule

    \multicolumn{7}{c}{\cellcolor{orange!10}\textbf{LiPS20}\textit{(eV/\AA{}, eV)}} \\
    $\beta$-Li$_3$PS$_4$ + $\gamma$-Li$_3$PS$_4$ & $117.31155 \pm 163.07083$ & $186.59096 \pm 273.21195$ & $\mathbf{0.72233 \pm 0.32170}$ & $\mathbf{0.15718 \pm 0.14543}$ & $0.43817 \pm 0.00597$ & $0.17530 \pm 0.12641$ \\
    Li$_3$P + Li$_2$S & $28.40555 \pm 25.35568$ & $66.10543 \pm 29.91355$ & $\mathbf{0.41767 \pm 0.05757}$ & $\mathbf{0.09184 \pm 0.04706}$ & $0.32156 \pm 0.00277$ & $0.06562 \pm 0.01921$ \\
    Li$_2$S + Li$_3$P + P$_2$S$_5$ & $72.59601 \pm 64.22975$ & $99.08896 \pm 38.98840$ & $\mathbf{1.21035 \pm 0.78647}$ & $\mathbf{0.86130 \pm 0.63587}$ & $0.34083 \pm 0.00482$ & $0.11110 \pm 0.05115$ \\
    \bottomrule
    \end{tabular}
    }
\end{table*}

\begin{table*}[t]
    \centering
    \caption{\textbf{Orb: Predictive Fidelity (Forces Only, Mean $\pm$ Std).} Baseline comparison (Wt Avg and Fisher) vs. \nameff{} and Joint FT.
    \textbf{Task Definitions:} 
    \textit{DHA}: Docosahexaenoic Acid.
    $^\ddagger$\textit{MD17 5-Task Mix}: Ethanol + Naphthalene + Salicylic Acid + Uracil + Aspirin. 
    $^\star$\textit{MD22 3-Task Mix}: Ac-Ala3-NHMe + DHA + Stachyose.}
    \label{tab:Orb_results_app_wt_fisher}
    
    \scriptsize
    \setlength{\tabcolsep}{5pt}
    \renewcommand{\arraystretch}{1.2}
    
    \begin{tabular}{l|c|c||c|c}
    \toprule
    \textbf{Dataset \& Scenario} & \textbf{Wt Avg} & \textbf{Fisher} & \textbf{\nameff} & \textbf{Joint FT} \\
    \midrule
    
    \multicolumn{5}{c}{\cellcolor{green!10}\textbf{MD17} \textit{(Force MAE in kcal/mol/\AA{})}} \\
    Aspirin + Uracil & 4.0646 $\pm$ 0.0913 & 4.8746 $\pm$ 0.1175 & \textbf{0.8543 $\pm$ 0.0015} & 0.7810 $\pm$ 0.0135 \\
    Ethanol + Malonaldehyde & 2.9590 $\pm$ 0.0289 & 3.9141 $\pm$ 0.1535 & \textbf{0.9521 $\pm$ 0.0010} & 0.8724 $\pm$ 0.0045 \\
    Naphthalene + Salicylic Acid & 1.8898 $\pm$ 0.0331 & 2.3657 $\pm$ 0.0431 & \textbf{0.6217 $\pm$ 0.0141} & 0.5440 $\pm$ 0.0067 \\
    Ethanol + Malon. + Aspirin & 3.2522 $\pm$ 0.0219 & 4.3496 $\pm$ 0.0780 & \textbf{1.0123 $\pm$ 0.0050} & 0.8479 $\pm$ 0.0034 \\
    Naph. + Salicylic + Uracil & 3.4143 $\pm$ 0.0229 & 4.2215 $\pm$ 0.1296 & \textbf{0.6686 $\pm$ 0.0041} & 0.5246 $\pm$ 0.0048 \\
    \textit{5-Task Mix (MD17)$^\ddagger$} & 4.0580 $\pm$ 0.0262 & 5.0479 $\pm$ 0.0511 & \textbf{0.9372 $\pm$ 0.0097} & 0.6254 $\pm$ 0.0099 \\
    \midrule
    
    \multicolumn{5}{c}{\cellcolor{blue!10}\textbf{MD22} \textit{(Force MAE in kcal/mol/\AA{})}} \\
    Ac-Ala3-NHMe + AT-AT & 3.8369 $\pm$ 0.0158 & 4.7890 $\pm$ 0.2071 & \textbf{1.0978 $\pm$ 0.0066} & 0.9624 $\pm$ 0.0065 \\
    DHA + Stachyose & 3.4758 $\pm$ 0.0202 & 4.4674 $\pm$ 0.1626 & \textbf{1.0412 $\pm$ 0.0020} & 0.9121 $\pm$ 0.0053 \\
    \textit{3-Task Mix (MD22)$^\star$} & 3.8259 $\pm$ 0.0129 & 5.0768 $\pm$ 0.2276 & \textbf{1.1518 $\pm$ 0.0070} & 0.8888 $\pm$ 0.0092 \\
    \midrule
    
    \multicolumn{5}{c}{\cellcolor{orange!10}\textbf{LiPS20} \textit{(Force MAE in eV/\AA{})}} \\
    $\beta$-Li$_3$PS$_4$ + $\gamma$-Li$_3$PS$_4$ & 0.3144 $\pm$ 0.0014 & 0.3096 $\pm$ 0.0035 & \textbf{0.0982 $\pm$ 0.0052} & 0.0620 $\pm$ 0.0001 \\
    Li$_3$P + Li$_2$S & 0.3020 $\pm$ 0.0021 & 0.3119 $\pm$ 0.0016 & \textbf{0.0862 $\pm$ 0.0031} & 0.0580 $\pm$ 0.0002 \\
    Li$_2$S + Li$_3$P + P$_2$S$_5$ & 0.4136 $\pm$ 0.0028 & 0.4804 $\pm$ 0.0083 & \textbf{0.1310 $\pm$ 0.0048} & 0.0741 $\pm$ 0.0003 \\
    \bottomrule
    \end{tabular}
\end{table*}

\begin{table*}[t]
    \centering
    \caption{\textbf{Orb: Predictive Fidelity (Forces Only, Mean $\pm$ Std).} Baseline comparison (TIES and EMR) vs. \nameff{} and Joint FT.
    \textbf{Task Definitions:} 
    \textit{DHA}: Docosahexaenoic Acid.
    $^\ddagger$\textit{MD17 5-Task Mix}: Ethanol + Naphthalene + Salicylic Acid + Uracil + Aspirin. 
    $^\star$\textit{MD22 3-Task Mix}: Ac-Ala3-NHMe + DHA + Stachyose.}
    \label{tab:Orb_results_app_ties_emr}
    
    \scriptsize
    \setlength{\tabcolsep}{5pt}
    \renewcommand{\arraystretch}{1.2}
    
    \begin{tabular}{l|c|c||c|c}
    \toprule
    \textbf{Dataset \& Scenario} & \textbf{TIES} & \textbf{EMR} & \textbf{\nameff} & \textbf{Joint FT} \\
    \midrule
    
    \multicolumn{5}{c}{\cellcolor{green!10}\textbf{MD17} \textit{(Force MAE in kcal/mol/\AA{})}} \\
    Aspirin + Uracil & 3.8015 $\pm$ 0.0494 & 4.6623 $\pm$ 0.0857 & \textbf{0.8543 $\pm$ 0.0015} & 0.7810 $\pm$ 0.0135 \\
    Ethanol + Malonaldehyde & 3.1828 $\pm$ 0.0375 & 3.2366 $\pm$ 0.0512 & \textbf{0.9521 $\pm$ 0.0010} & 0.8724 $\pm$ 0.0045 \\
    Naphthalene + Salicylic Acid & 1.9862 $\pm$ 0.0130 & 2.2434 $\pm$ 0.0452 & \textbf{0.6217 $\pm$ 0.0141} & 0.5440 $\pm$ 0.0067 \\
    Ethanol + Malon. + Aspirin & 3.4903 $\pm$ 0.0173 & 5.1306 $\pm$ 0.0366 & \textbf{1.0123 $\pm$ 0.0050} & 0.8479 $\pm$ 0.0034 \\
    Naph. + Salicylic + Uracil & 3.2744 $\pm$ 0.0472 & 5.1665 $\pm$ 0.0507 & \textbf{0.6686 $\pm$ 0.0041} & 0.5246 $\pm$ 0.0048 \\
    \textit{5-Task Mix (MD17)$^\ddagger$} & 3.9755 $\pm$ 0.0162 & 7.5926 $\pm$ 0.0877 & \textbf{0.9372 $\pm$ 0.0097} & 0.6254 $\pm$ 0.0099 \\
    \midrule
    
    \multicolumn{5}{c}{\cellcolor{blue!10}\textbf{MD22} \textit{(Force MAE in kcal/mol/\AA{})}} \\
    Ac-Ala3-NHMe + AT-AT & 3.7324 $\pm$ 0.0109 & 4.6524 $\pm$ 0.0328 & \textbf{1.0978 $\pm$ 0.0066} & 0.9624 $\pm$ 0.0065 \\
    DHA + Stachyose & 3.3783 $\pm$ 0.0107 & 4.0178 $\pm$ 0.0316 & \textbf{1.0412 $\pm$ 0.0020} & 0.9121 $\pm$ 0.0053 \\
    \textit{3-Task Mix (MD22)$^\star$} & 3.5591 $\pm$ 0.0047 & 7.0931 $\pm$ 0.0365 & \textbf{1.1518 $\pm$ 0.0070} & 0.8888 $\pm$ 0.0092 \\
    \midrule
    
    \multicolumn{5}{c}{\cellcolor{orange!10}\textbf{LiPS20} \textit{(Force MAE in eV/\AA{})}} \\
    $\beta$-Li$_3$PS$_4$ + $\gamma$-Li$_3$PS$_4$ & 0.3319 $\pm$ 0.0028 & 0.3134 $\pm$ 0.0016 & \textbf{0.0982 $\pm$ 0.0052} & 0.0620 $\pm$ 0.0001 \\
    Li$_3$P + Li$_2$S & 0.3233 $\pm$ 0.0006 & 0.3035 $\pm$ 0.0023 & \textbf{0.0862 $\pm$ 0.0031} & 0.0580 $\pm$ 0.0002 \\
    Li$_2$S + Li$_3$P + P$_2$S$_5$ & 0.4054 $\pm$ 0.0039 & 0.4538 $\pm$ 0.0035 & \textbf{0.1310 $\pm$ 0.0048} & 0.0741 $\pm$ 0.0003 \\
    \bottomrule
    \end{tabular}
\end{table*}

\begin{table*}[t]
    \centering
    \caption{\textbf{Full Computational Efficiency Analysis.} Comparison of total training time (seconds) for all datasets and molecular combinations. Mean $\pm$ Std reported over 3 seeds.
    \textbf{Task Definitions:} 
    \textit{DHA}: Docosahexaenoic Acid.
    $^\ddagger$\textit{MD17 5-Task Mix}: Ethanol + Naphthalene + Salicylic Acid + Uracil + Aspirin. 
    $^\star$\textit{MD22 3-Task Mix}: Ac-Ala3-NHMe + DHA + Stachyose.}
    \label{tab:efficiency_full}
    
    \small 
    \setlength{\tabcolsep}{4pt}
    \renewcommand{\arraystretch}{1.15}
    
    \begin{tabular}{l|c|c|c|c|c}
    \toprule
    \textbf{Merging Scenario} & \textbf{Closed Form (s)} & \textbf{Targeted Fine Tune (s)} & \textbf{GFFMerge (s)} & \textbf{Joint FT (s)} & \textbf{Speedup} \\
    \midrule
    \multicolumn{6}{c}{\cellcolor{blue!10}\textbf{Architecture: M3GNet}} \\ 
    MD17: Aspirin + Uracil & $4.34 \pm 0.09$ & $45.48 \pm 0.36$ & $\mathbf{49.82 \pm 0.35}$ & $533.56 \pm 2.13$ & $\mathbf{10.71\times}$ \\
    MD17: Ethanol + Malonaldehyde & $3.31 \pm 0.32$ & $48.29 \pm 0.52$ & $\mathbf{51.60 \pm 0.41}$ & $289.90 \pm 1.56$ & $\mathbf{5.62\times}$ \\
    MD17: Naphthalene + Salicylic & $4.64 \pm 0.20$ & $51.32 \pm 0.54$ & $\mathbf{55.96 \pm 0.50}$ & $578.17 \pm 1.65$ & $\mathbf{10.33\times}$ \\
    MD17: Ethanol + Malon. + Asp. & $5.28 \pm 0.12$ & $85.09 \pm 0.92$ & $\mathbf{90.37 \pm 0.91}$ & $615.58 \pm 0.88$ & $\mathbf{6.81\times}$ \\
    MD17: Naph. + Salicylic + Ura. & $6.14 \pm 0.27$ & $107.91 \pm 2.09$ & $\mathbf{114.05 \pm 2.07}$ & $752.61 \pm 4.54$ & $\mathbf{6.60\times}$ \\
    MD17: \textit{5-Task Mix}$^\ddagger$ & $8.88 \pm 0.10$ & $233.00 \pm 5.20$ & $\mathbf{241.88 \pm 5.20}$ & $1238.18 \pm 15.44$ & $\mathbf{5.12\times}$ \\
    MD22: Ac-Ala3 + AT-AT & $12.32 \pm 0.21$ & $296.96 \pm 1.16$ & $\mathbf{309.28 \pm 1.14}$ & $2499.84 \pm 16.19$ & $\mathbf{8.08\times}$ \\
    MD22: DHA + Stachyose & $21.91 \pm 0.21$ & $633.55 \pm 6.29$ & $\mathbf{655.46 \pm 6.29}$ & $5714.04 \pm 48.55$ & $\mathbf{8.72\times}$ \\
    MD22: \textit{3-Task Mix}$^\star$ & $28.60 \pm 0.25$ & $236.02 \pm 3.31$ & $\mathbf{264.62 \pm 3.30}$ & $7358.42 \pm 2.13$ & $\mathbf{27.81\times}$ \\
    LiPS: $\beta$-Li$_3$PS$_4$ + $\gamma$-Li$_3$PS$_4$ & $14.34 \pm 0.14$ & $453.09 \pm 3.69$ & $\mathbf{467.43 \pm 3.69}$ & $2211.59 \pm 7.06$ & $\mathbf{4.73\times}$ \\
    LiPS: Li$_3$P + Li$_2$S & $13.42 \pm 0.06$ & $579.70 \pm 0.09$ & $\mathbf{593.12 \pm 0.07}$ & $3269.65 \pm 25.02$ & $\mathbf{5.51\times}$ \\
    LiPS: Li$_2$S + Li$_3$P + P$_2$S$_5$ & $15.15 \pm 0.16$ & $488.84 \pm 4.73$ & $\mathbf{503.99 \pm 4.73}$ & $3717.50 \pm 44.83$ & $\mathbf{7.38\times}$ \\
    \midrule
    \multicolumn{6}{c}{\cellcolor{orange!10}\textbf{Architecture: Orb}} \\ 
    MD17: Aspirin + Uracil & $9.59 \pm 0.09$ & $133.44 \pm 4.56$ & $\mathbf{143.03 \pm 4.56}$ & $1028.77 \pm 26.41$ & $\mathbf{7.19\times}$ \\
    MD17: Ethanol + Malonaldehyde & $8.25 \pm 0.03$ & $75.00 \pm 2.06$ & $\mathbf{83.25 \pm 2.06}$ & $591.51 \pm 14.05$ & $\mathbf{7.11\times}$ \\
    MD17: Naphthalene + Salicylic & $9.84 \pm 0.09$ & $124.69 \pm 5.74$ & $\mathbf{134.53 \pm 5.74}$ & $739.21 \pm 50.09$ & $\mathbf{5.49\times}$ \\
    MD17: Ethanol + Malon. + Asp. & $10.29 \pm 0.20$ & $172.74 \pm 17.62$ & $\mathbf{183.03 \pm 17.62}$ & $1121.72 \pm 79.47$ & $\mathbf{6.13\times}$ \\
    MD17: Naph. + Salicylic + Ura. & $10.86 \pm 0.27$ & $213.05 \pm 10.61$ & $\mathbf{223.91 \pm 10.61}$ & $1209.82 \pm 91.08$ & $\mathbf{5.40\times}$ \\
    MD17: \textit{5-Task Mix}$^\ddagger$ & $13.46 \pm 0.52$ & $445.11 \pm 55.40$ & $\mathbf{458.57 \pm 55.40}$ & $2492.53 \pm 174.99$ & $\mathbf{5.44\times}$ \\
    MD22: Ac-Ala3 + AT-AT & $15.34 \pm 0.15$ & $378.65 \pm 17.08$ & $\mathbf{393.99 \pm 17.08}$ & $2246.82 \pm 64.46$ & $\mathbf{5.70\times}$ \\
    MD22: DHA + Stachyose & $18.93 \pm 0.06$ & $468.15 \pm 10.63$ & $\mathbf{487.08 \pm 10.63}$ & $2937.06 \pm 11.20$ & $\mathbf{6.03\times}$ \\
    MD22: \textit{3-Task Mix}$^\star$ & $22.64 \pm 0.04$ & $525.88 \pm 2.40$ & $\mathbf{548.52 \pm 2.40}$ & $3919.08 \pm 3.96$ & $\mathbf{7.14\times}$ \\
    LiPS: $\beta$-Li$_3$PS$_4$ + $\gamma$-Li$_3$PS$_4$ & $21.42 \pm 0.11$ & $363.24 \pm 19.11$ & $\mathbf{384.66 \pm 19.11}$ & $2743.24 \pm 16.84$ & $\mathbf{7.13\times}$ \\
    LiPS: Li$_3$P + Li$_2$S & $19.38 \pm 0.12$ & $464.04 \pm 3.55$ & $\mathbf{483.42 \pm 3.55}$ & $2476.80 \pm 4.38$ & $\mathbf{5.12\times}$ \\
    LiPS: Li$_2$S + Li$_3$P + P$_2$S$_5$ & $21.16 \pm 0.10$ & $480.34 \pm 17.77$ & $\mathbf{501.50 \pm 17.77}$ & $2954.31 \pm 54.21$ & $\mathbf{5.89\times}$ \\
    \bottomrule
    \end{tabular}
\end{table*}
\clearpage

\section{Extended Ablation Studies}
\label{app:ablation_extended}

\subsection{Ablations on other datasets}
We extend our ablation analysis to small molecules (MD17) and supramolecular systems (MD22) to confirm that the trends observed in the solid-state LiPS system generalize across chemical domains.

\begin{figure*}[!h]
    \centering
    \caption{\textbf{MD17 Ablation Results.}
    \textbf{Top Row:} Test MAE vs. Fine-tuning Epochs for the 5-Task mix (Ethanol, Naphthalene, Salicylic Acid, Uracil, Aspirin).
    \textbf{Bottom Row:} Test MAE vs. Data limit. Trends are consistent with the LiPS results.}
    \label{fig:ablation_md17}
    \setlength{\tabcolsep}{1pt}
    
    \begin{tabular}{ccc}
        \includegraphics[trim=0 737 950 0,clip,width=0.25\textwidth]{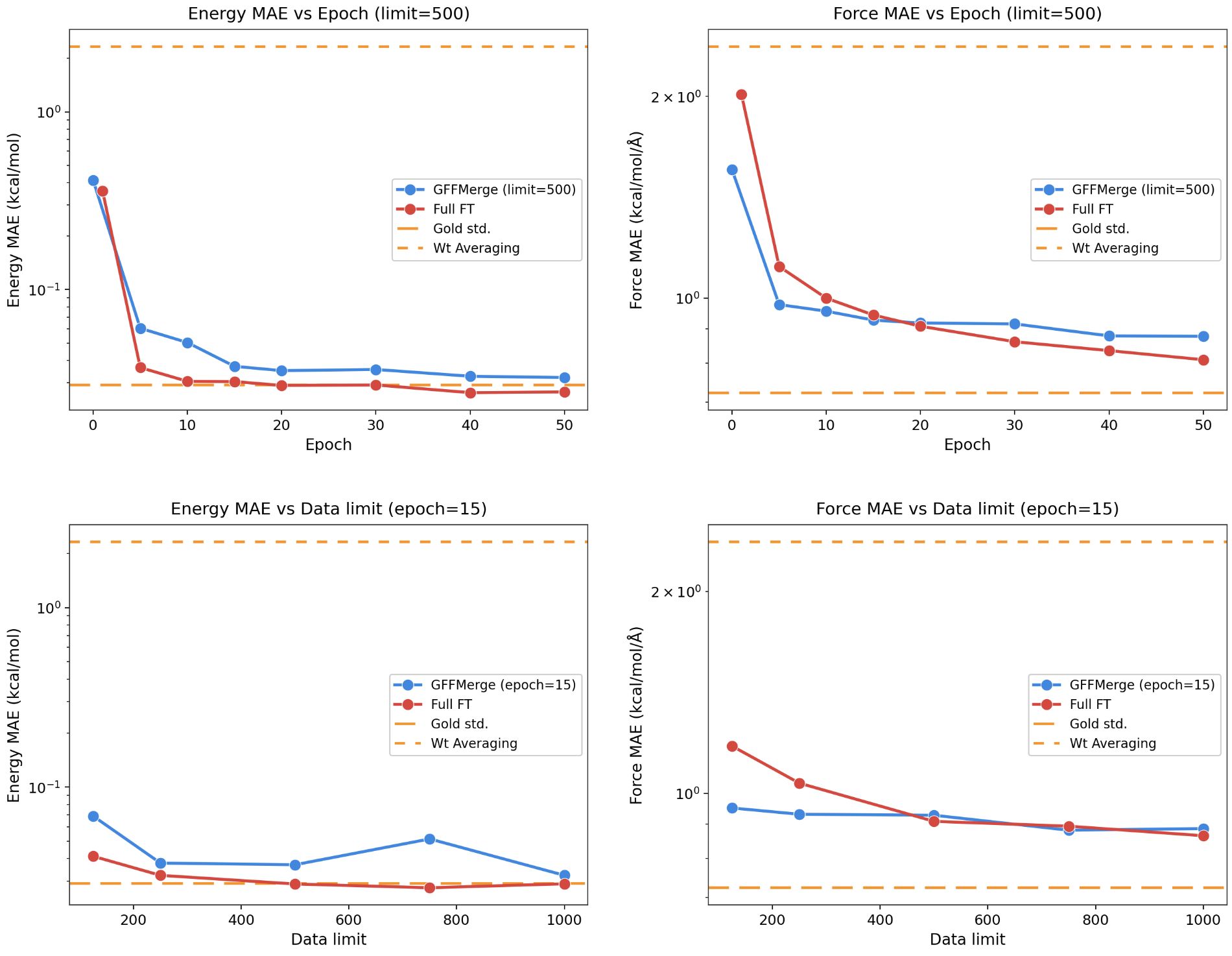} &
        \includegraphics[trim=950 737 0 0,clip,width=0.25\textwidth]{images/Plots/updated_ablation/md17_m3gnet.jpg} &
        \includegraphics[trim=0 875 0 0,clip,width=0.25\textwidth]{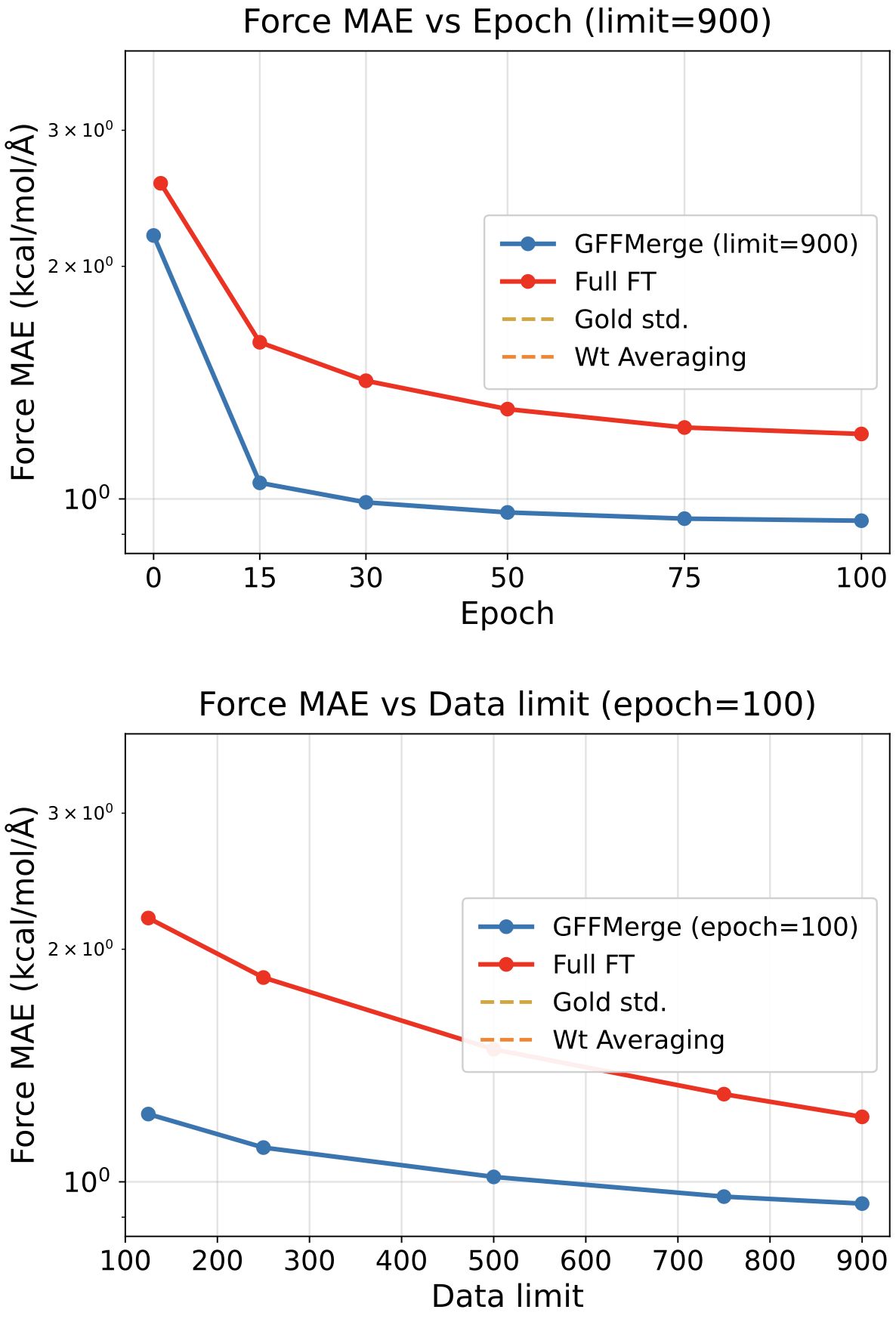} \\
        \footnotesize (a) M3GNet Energy (Convergence) &
        \footnotesize (b) M3GNet Force (Convergence) &
        \footnotesize (c) Orb Force (Convergence) \\
    \end{tabular}
    
    \begin{tabular}{ccc}
        \includegraphics[trim=0 0 950 737,clip,width=0.25\textwidth]{images/Plots/updated_ablation/md17_m3gnet.jpg} &
        \includegraphics[trim=950 0 0 737,clip,width=0.25\textwidth]{images/Plots/updated_ablation/md17_m3gnet.jpg} &
        \includegraphics[trim=0 0 0 875,clip,width=0.25\textwidth]{images/Plots/updated_ablation/md17_orb.jpg} \\
        \footnotesize (d) M3GNet Energy (Efficiency) &
        \footnotesize (e) M3GNet Force (Efficiency) &
        \footnotesize (f) Orb Force (Efficiency) \\
    \end{tabular}
\end{figure*}

\begin{figure*}[!hbtp]
    \centering
    \caption{\textbf{MD22 Ablation Results.}
    \textbf{Top Row:} Test MAE vs. Fine-tuning Epochs for the supramolecular DHA + Stachyose task.
    \textbf{Bottom Row:} Test MAE vs. Data limit.}
    \label{fig:ablation_md22}
    \setlength{\tabcolsep}{1pt}
    
    \begin{tabular}{ccc}
        \includegraphics[trim=0 692 930 0,clip,width=0.25\textwidth]{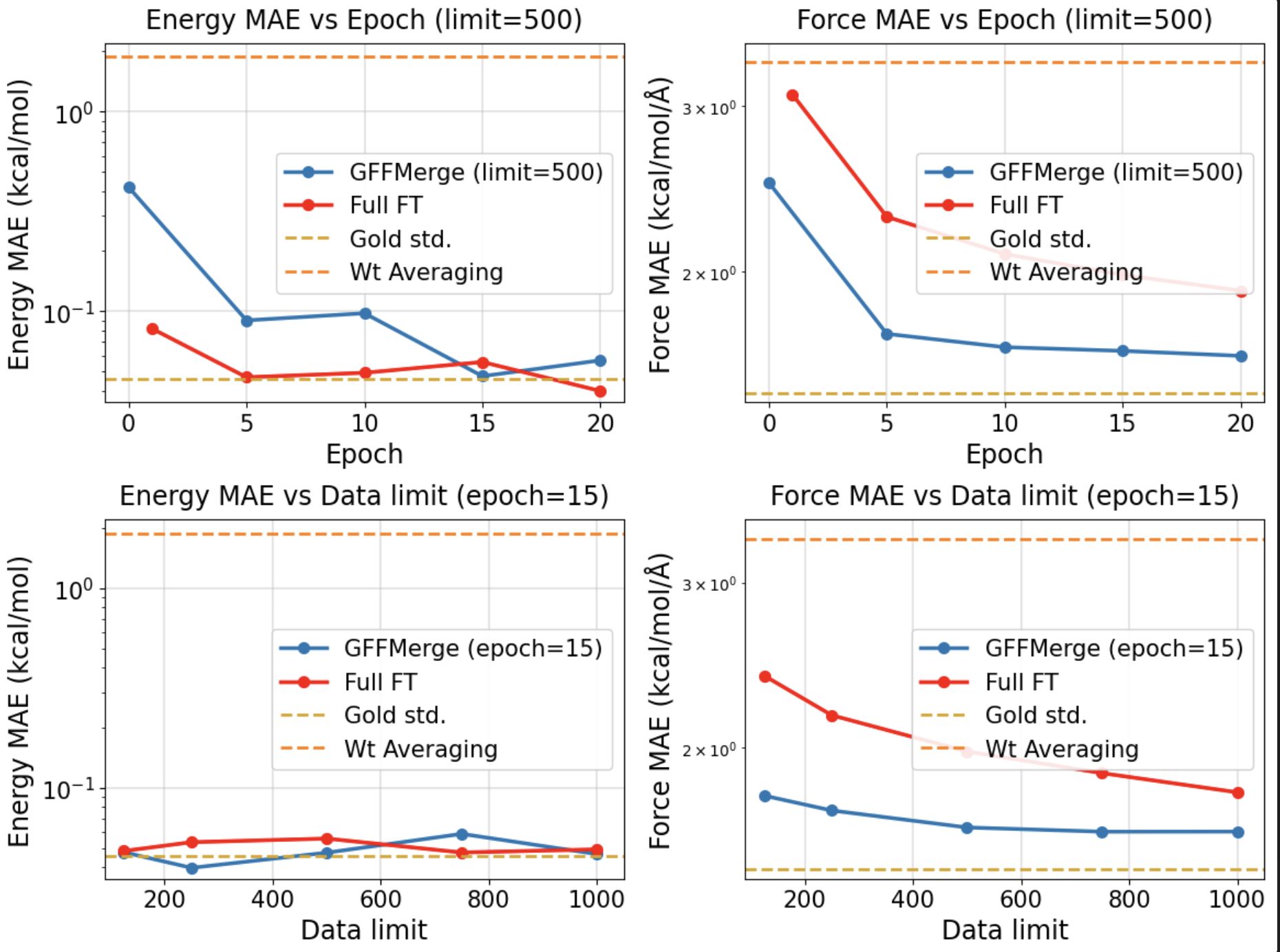} &
        \includegraphics[trim=930 692 0 0,clip,width=0.25\textwidth]{images/Plots/updated_ablation/md22_m3gnet.jpg} &
        \includegraphics[trim=0 694 0 0,clip,width=0.25\textwidth]{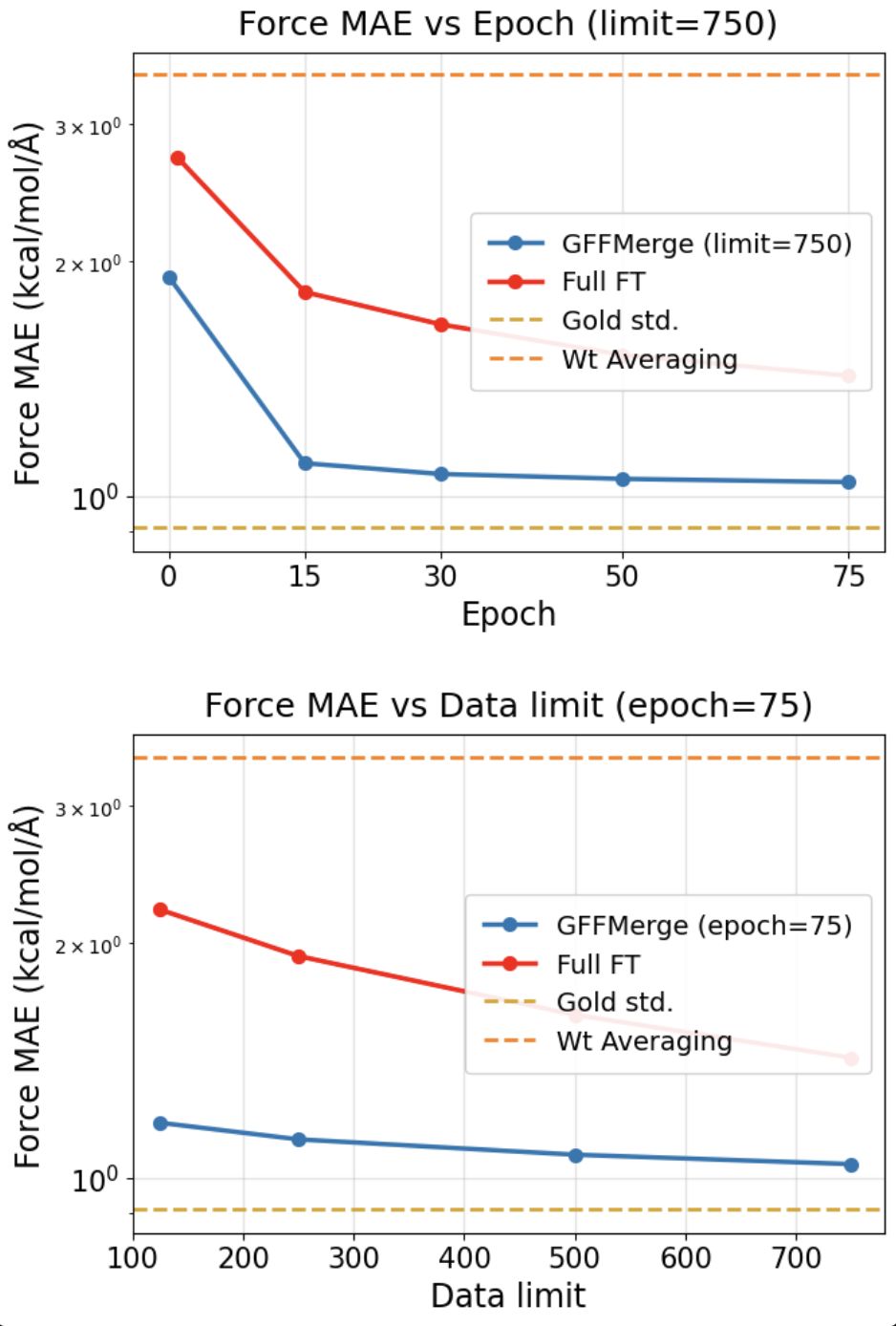} \\
        \footnotesize (a) M3GNet Energy (Convergence) &
        \footnotesize (b) M3GNet Force (Convergence) &
        \footnotesize (c) Orb Force (Convergence) \\
    \end{tabular}
    
    \begin{tabular}{ccc}
        \includegraphics[trim=0 0 930 692,clip,width=0.25\textwidth]{images/Plots/updated_ablation/md22_m3gnet.jpg} &
        \includegraphics[trim=930 0 0 692,clip,width=0.25\textwidth]{images/Plots/updated_ablation/md22_m3gnet.jpg} &
        \includegraphics[trim=0 0 0 694,clip,width=0.25\textwidth]{images/Plots/updated_ablation/md22_orb.jpg} \\
        \footnotesize (d) M3GNet Energy (Efficiency) &
        \footnotesize (e) M3GNet Force (Efficiency) &
        \footnotesize (f) Orb Force (Efficiency) \\
    \end{tabular}
\end{figure*}

\begin{review}
\subsection{Ablation Studies on Unfrozen Layers}
We investigate the sensitivity of the fine-tuning and merging process to the number of active parameter blocks. The number of unfrozen blocks dictates both the downstream accuracy and the computational overhead. Figures \ref{ablation_m3gnet_block} and \ref{ablation_orb_block} reveal a natural trade-off: unfreezing more blocks yields improvements in Force MAE but monotonically increases the wall-clock merging time.

\begin{figure}[!hbtp]
    \centering
    \caption{Unfreezing-layer ablation on M3GNet. X-axis indicate the number of last blocks unfrozen during fine-tuning and Y-axis indicates the MAE in log scale. For Ethanol+Malonaldehyde+Aspirin, Force MAE (F) is in kcal/mol/\AA{} and Energy (E) in kcal/mol. For Li$_2$S+Li$_3$P, F is in eV/\AA{} and E in eV.}
    \label{ablation_m3gnet_block}
\includegraphics[width=1\textwidth]{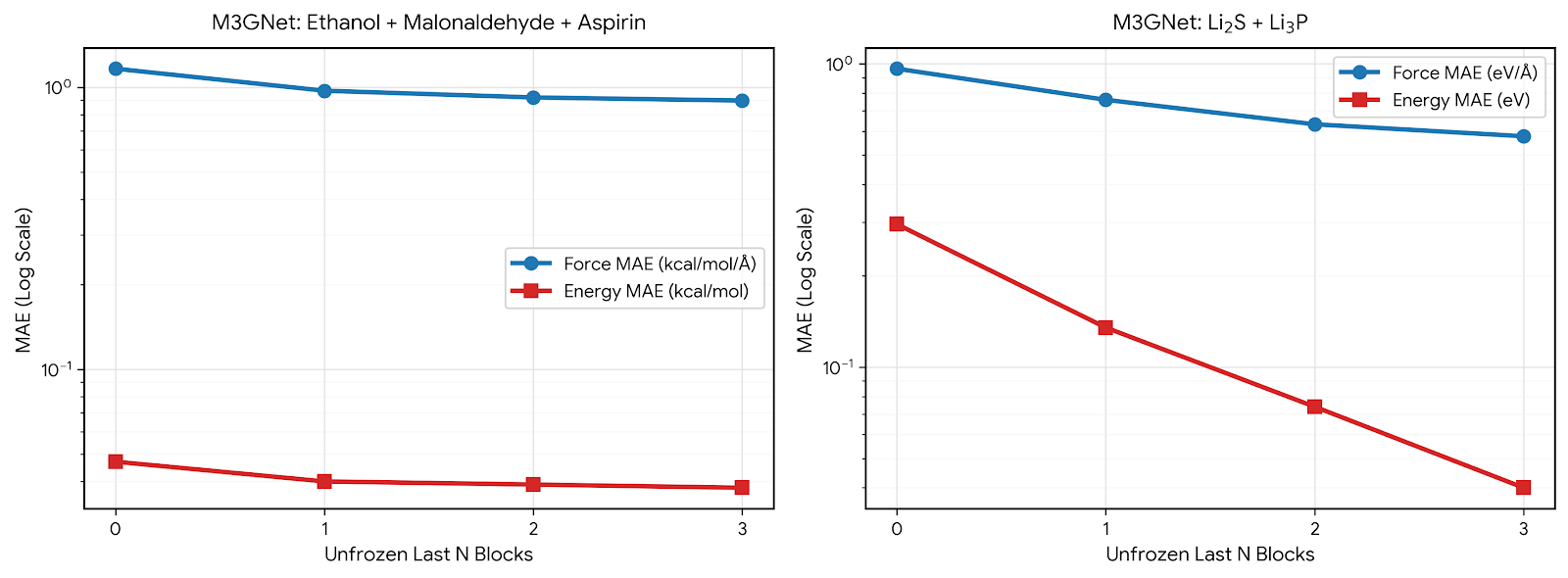}
\end{figure}

\begin{figure}[!hbtp]
    \centering
    \caption{Unfreezing-layer ablation on Orb. X-axis indicate the number of last blocks unfrozen during fine-tuning and Y-axis indicates the MAE in log scale. For Ethanol+Malonaldehyde+Aspirin, Force MAE (F) is in kcal/mol/\AA{} and Energy (E) in kcal/mol. For Li$_2$S+Li$_3$P, F is in eV/\AA{} and E in eV.}
    \label{ablation_orb_block}
\includegraphics[width=1\textwidth]{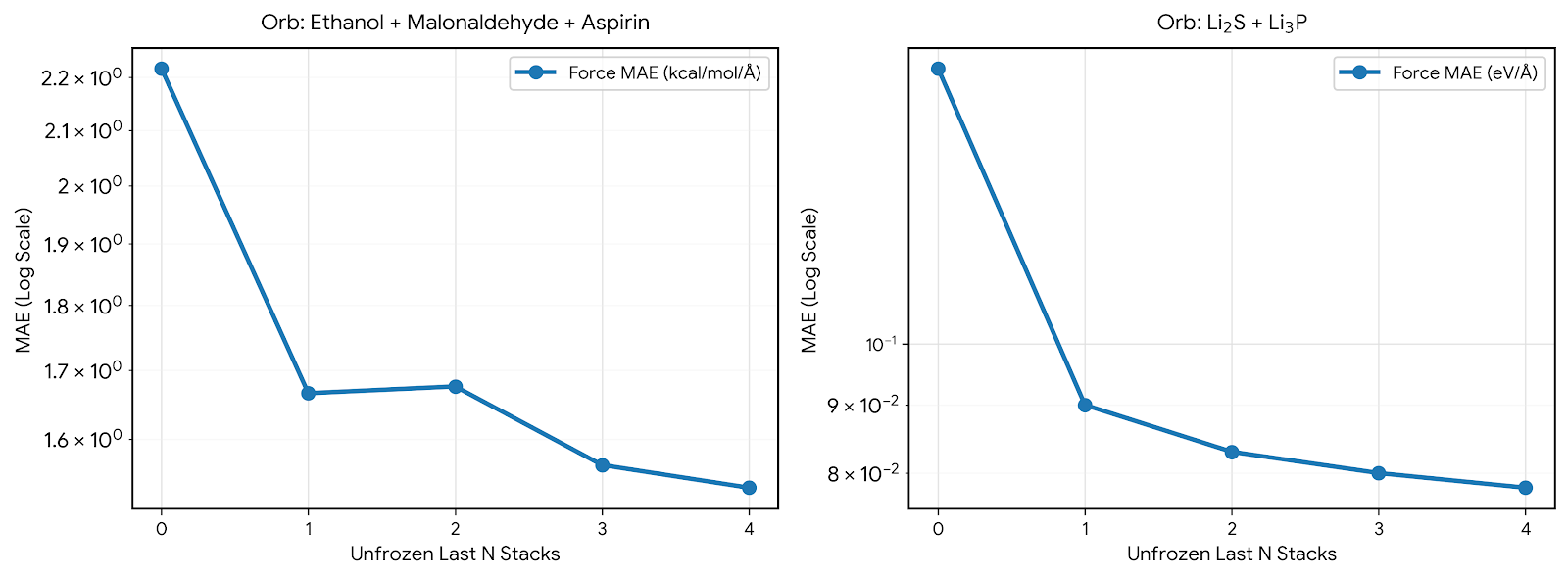}
\end{figure}
\end{review}

\FloatBarrier

\section{Additional experiments on Generic \gnns}
\label{app:gnnmerge}
\begin{figure*}[t!]
    \centering
        \includegraphics[width=0.4\textwidth]{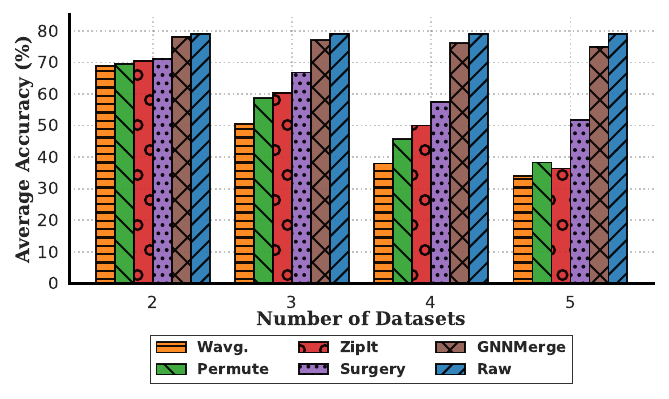}
        \caption{Variation of average accuracy of merging methods as the number of models varies.}
        \label{fig:datasets}
\end{figure*}

To stress-test the methods, we extend the analysis by merging $x$ models corresponding to $x$ different datasets, where $x\geq2$. In Fig.~\ref{fig:datasets}, we present the average accuracy of the merged models as $x$ is varied. 
\begin{table*}[!h]
\centering
\scalebox{0.87}{
\begin{tabular}{l|ccccccc}
\toprule
\textbf{Datasets} & \textbf{Base}& \textbf{\wa} & \textbf{\git} & \textbf{\permute} & \textbf{\zipit} & \textbf{\surgery} & \textbf{\name} \\
\midrule
\textbf{Citeseer} & 81.97 & 78.09 & 80.25 & 79.15 & 78.68 & 79.50$\pm$\scriptsize{1.42} & \textbf{82.44} \\
\textbf{Pubmed}  & 79.02 & 75.94 & 22.23 & 78.47 & 77.25 & 68.69$\pm$\scriptsize{1.76} & \textbf{79.04} \\
\midrule
\textbf{Citeseer} & 81.97 & 67.54 & 71.78 & 73.19 & 74.92 & 79.56$\pm$\scriptsize{1.89} & \textbf{82.60} \\
\textbf{WikiCS}  & 79.32 & 60.27 & 22.90 & 61.99 & 63.28 & 71.19$\pm$\scriptsize{1.21} & \textbf{78.21} \\
\midrule
\textbf{Arxiv}   & 73.10 & 68.43 & 53.12 & 53.56 & 50.11 & 60.46$\pm$\scriptsize{1.73} & \textbf{71.98} \\
\textbf{WikiCS}  & 79.32 & 66.89 & 25.98 & 61.55 & 67.16 & 72.40$\pm$\scriptsize{1.58} & \textbf{78.67} \\
\midrule
\textbf{Arxiv}   & 73.10 & 61.40 & 60.47 & 57.64 & 59.05 & 57.66$\pm$\scriptsize{1.65} & \textbf{72.65} \\
\textbf{Pubmed}  & 79.02 & 74.28 & 20.88 & 78.04 & 78.12 & 75.39$\pm$\scriptsize{1.48} & \textbf{79.13} \\
\midrule
\textbf{Pubmed}  & 79.02 & 76.20 & 67.88 & 75.81 & 75.16 & 74.97$\pm$\scriptsize{1.36} & \textbf{78.96} \\
\textbf{WikiCS}  & 79.32 & 70.68 & 8.02  & 69.95 & 73.16 & 69.36$\pm$\scriptsize{1.52} & \textbf{78.89} \\
\midrule
\textbf{Mag240M} & 66.46 & 44.14 & 51.97 & 54.12 & 56.37 & - & \textbf{66.14} \\
\textbf{Papers100M} & 66.44 & 48.56 & 10.03 & 51.73 & 55.15 & - & \textbf{66.28} \\
\midrule
\textbf{Average} & 79.05 & 68.86 & 46.86 & 69.51 & 70.49 & 71.10 & \textbf{78.07} \\
\bottomrule
\end{tabular}}
\caption{Merging of models trained on different datasets. Metric reported: Accuracy $(\%)$.}
\label{tab:results2half}
\end{table*}

\begin{table*}[!h]
\centering
\scalebox{0.9}{
\begin{tabular}{ l|ccccccc}
\toprule
\textbf{Tasks} & \textbf{Base} & \textbf{Random} & \textbf{\wa} & \textbf{\permute} & \textbf{\zipit} & \textbf{\surgery} & \textbf{\name} \\
\midrule
\textbf{Arxiv-NC}  & 73.10 & 9.31  & 57.99 & 61.03 & 60.72 & 66.54{\scriptsize$\pm$1.42} & \textbf{73.01} \\
\textbf{Pubmed-LP} & 97.05 & 90.22 & 91.32 & 91.78 & 91.50 & 93.67{\scriptsize$\pm$1.63} & \textbf{96.23} \\
\midrule
\textbf{WikiCS-NC} & 79.32 & 13.99 & 75.18 & 74.79 & 75.88 & 77.02{\scriptsize$\pm$1.37} & \textbf{78.88} \\
\textbf{Cora-LP}   & 94.34 & 83.62 & 77.28 & 76.90 & 80.56 & 88.39{\scriptsize$\pm$1.94} & \textbf{94.45} \\
\midrule
\textbf{WikiCS-NC} & 79.32 & 22.87 & 69.46 & 71.30 & 73.35 & 76.93{\scriptsize$\pm$1.55} & \textbf{78.91} \\
\textbf{Pubmed-LP} & 97.05 & 91.50 & 88.70 & 90.07 & 89.45 & 92.71{\scriptsize$\pm$1.81} & \textbf{96.37} \\
\midrule
\end{tabular}}
\caption{\textbf{Different Tasks.} \name compared with baselines when merging models trained for different tasks and datasets. NC: Node Classification (Accuracy \%). LP: Link Prediction (ROC-AUC).}
\label{tab:results6}
\end{table*}

As depicted in Fig.~\ref{fig:datasets}, \name demonstrates superior performance across all values of $x$, with up to \textbf{23.53\%} improvement over the next best baseline. 

\comment{
\subsection{Ablation Study: Impact of layer-wise independent merging}
\label{sec:ablation}
As discussed in \S~\ref{sec:closed_form}, learning parameters through \textit{joint} node embedding alignment introduces computational overhead and slower convergence (Eq.~\ref{eq:opt_ff}). To mitigate this, we propose layer-wise independent node embedding alignment (Eq.~\ref{eq:optrelax_ff}). Fig.~\ref{fig:layers} presents the performance of the two optimization strategies as we vary the number of \gcn layers in the merging models on the arXiv dataset. A clear trend emerges: as the number of \gnn layers increases, the performance of the joint node alignment strategy deteriorates (indicated by a growing gap with the base model accuracies). In contrast, the relaxed optimization strategy, which aligns each layer independently, remains stable and does not suffer from this degradation. This behavior is attributed to the vanishing gradient problem becoming more pronounced in joint node alignment as the number of layers increases. Treating layers independently circumvents this issue, as the optimization problem is decoupled from \gnn depth.}

\subsection{Merging of Heterogenous Architectures}
\label{app:hetero}
We consider two cases of merging heterogeneous architectures and show \name's applicability.
\begin{itemize}
    \item \textbf{Different MPNN Architectures: }Unlike existing methods, \name can be used to merge \gnn models of differing architectures-e.g., merging \sage ~\cite{graphsage}, and  \gat~\cite{velickovic2018graph} into a \gcn ~\cite{kipf2017semi-gcn}. Since \name operates by aligning the embeddings of the new model with those of the base models, it remains agnostic to the base architectures once the intermediate representations are computed. This allows the new \gnn to be optimized solely with respect to its own architecture and the target embeddings. 
    \item \textbf{Different Hidden Dimensions: }Adapting \name to support varying hidden dimensions can be readily accomplished via padding. 
\end{itemize}
 Tables ~\ref{tab:results11} and ~\ref{tab:results13} present results for merging heterogeneous architectures. These results demonstrate that \name is not only applicable but also effective in this setting, highlighting its robustness compared to existing baselines, which are \textbf{inapplicable} in such scenarios.

\begin{table*}[h]
\centering
\begin{subtable}[t]{0.45\textwidth}
    \centering
\scalebox{0.85}{
\begin{tabular}{l|cc}
\toprule
\textbf{Datasets} & \textbf{Base} & \textbf{\name} \\
\midrule
\textbf{pubmed}   & 79.25 & 79.23 \\
\textbf{citeseer} & 82.91 & 82.76 \\
\midrule
\textbf{wikics}   & 78.86 & 78.11 \\
\textbf{citeseer} & 82.91 & 82.13 \\
\midrule
\textbf{cora}     & 80.27 & 80.03 \\
\textbf{pubmed}   & 78.91 & 78.72 \\
\midrule
\textbf{wikics}   & 78.86 & 78.21 \\
\textbf{cora}     & 81.81 & 81.05 \\
\midrule
\textbf{arxiv}    & 73.70 & 71.47 \\
\textbf{pubmed}   & 78.91 & 78.65 \\
\bottomrule
\end{tabular}
}
\caption{First row: GCN with \textbf{hidden dimension 256}; Second row: GCN with \textbf{hidden dimension 128}.}
\label{tab:results11}
\end{subtable}%
\hspace{0.08\textwidth}
\begin{subtable}[h]{0.45\textwidth}
   \centering
\scalebox{0.85}{
\begin{tabular}{l|cc}
\toprule
\textbf{Datasets} & \textbf{Base} & \textbf{\name} \\
\midrule
\textbf{pubmed} & 77.96 & 78.81 \\ 
\textbf{cora}   & 81.43 & 80.17 \\
\midrule
\textbf{wikics} & 78.82 & 78.90 \\ 
\textbf{cora}   & 81.43 & 80.21 \\
\midrule
\textbf{arxiv}  & 74.65 & 73.33 \\ 
\textbf{pubmed} & 78.21 & 78.59 \\
\midrule
\textbf{arxiv}  & 74.60 & 72.72 \\ 
\textbf{wikics} & 79.49 & 78.48 \\
\midrule
\textbf{pubmed} & 77.96 & 78.48 \\ 
\textbf{wikics} & 79.49 & 79.17 \\
\bottomrule
\end{tabular}
}
\caption{First row: \textbf{\sage}; Second row: \textbf{\gat}; Merged into \textbf{\gcn}.}
\label{tab:results13}
\end{subtable}
\caption{Performance of \name while merging heterogeneous \gnn architectures.}
\label{tab:combined}
\end{table*}

\begin{table*}[!h]
\centering
\scalebox{0.85}{
\begin{tabular}{ l|ccccccc}
\midrule
\textbf{Dataset} & \textbf{M} & \textbf{Base} & \textbf{\wa} & \textbf{\permute} & \textbf{\zipit} & \textbf{\surgery} & \textbf{\name} \\ 
\midrule
\multirow{4}{*}{\textbf{AmzComp}} 
 & \sage 1 & 95.46 & 58.55 & 84.47 & 66.92 & 85.88 & \textbf{94.11} \\ 
 & \sage 2 & 92.83 & 75.91 & 69.14 & 74.74 & 69.92 & \textbf{91.51} \\ 
\cmidrule{2-8}
 & \node 1 & 93.33 & 49.15 & - & - & 87.86 & \textbf{89.42} \\ 
 & \node 2 & 91.96 & 66.51 & - & - & 80.85 & \textbf{86.71} \\ 
\midrule
\multirow{4}{*}{\textbf{WikiCS}} 
 & \sage 1 & 86.64 & 80.36 & 83.35 & 76.00 & 84.14 & \textbf{84.43} \\ 
 & \sage 2 & 84.99 & 76.80 & 56.85 & 64.50 & 82.23 & \textbf{83.56} \\ 
\cmidrule{2-8}
 & \node 1 & 79.21 & 56.12 & - & - & 76.11 & \textbf{76.57} \\ 
 & \node 2 & 78.54 & 70.19 & - & - & 71.34 & \textbf{72.43} \\ 
\midrule
\end{tabular}}
\caption{\textbf{In-domain Dataset} experiments on \sage and \node. Metric reported: Accuracy$(\%)$. \permute and \zipit are not applicable for transformer architectures.}
\label{tab:results7}
\end{table*}

\begin{table*}[!h]
\centering
\scalebox{0.75}{
\begin{tabular}{ l|ccccccc}
\midrule
\textbf{Arch.} & \textbf{Datasets} & \textbf{Base} & \textbf{\wa} & \textbf{\permute} & \textbf{\zipit} & \textbf{\surgery} & \textbf{\name} \\
\midrule
\multirow{4}{*}{\sage} 
 & \textbf{Arxiv} & 74.65 & 55.13 & 58.45 & 59.13 & 68.39 {\scriptsize$\pm$1.54} & \textbf{72.78} \\
 & \textbf{WikiCS} & 78.82 & 72.84 & 56.17 & 56.68 & 69.22 {\scriptsize$\pm$1.72} & \textbf{78.63} \\
\cmidrule{2-8}
 & \textbf{Arxiv} & 74.65 & 68.77 & 60.00 & 63.60 & 67.85 {\scriptsize$\pm$1.09} & \textbf{74.29} \\
 & \textbf{Pubmed} & 77.96 & 72.32 & 62.21 & 65.94 & 75.12 {\scriptsize$\pm$1.66} & \textbf{77.97} \\
\midrule
\multirow{4}{*}{\node} 
 & \textbf{Cora} & 81.09 & 50.72 & - & - & \textbf{77.61} {\scriptsize$\pm$1.88} & 75.27 \\
 & \textbf{Citeseer} & 81.35 & 71.71 & - & - & \textbf{79.15} {\scriptsize$\pm$1.11} & 76.08 \\
\cmidrule{2-8}
 & \textbf{Pubmed} & 80.08 & 57.60 & - & - & 75.74 {\scriptsize$\pm$1.37} & \textbf{79.10} \\
 & \textbf{WikiCS} & 74.17 & 62.80 & - & - & 67.48 {\scriptsize$\pm$1.91} & \textbf{69.21} \\
\midrule
\end{tabular}}
\caption{\textbf{Two different datasets} experiments for \sage and \node. Metric reported: Accuracy$(\%)$. \permute and \zipit do not support transformer architectures.}
\label{tab:results8}
\end{table*}

\clearpage
\begin{review}
\section{Additional Experiments on GNN Force Fields}
\label{sec:additional_gnn_experiments}

We do further experiments using \nameff on \gnn Force Fields for new datasets and combinations in this section.

\subsection{Joint Fine-tuning on Same Budget (Joint FT-SB)}
We have expanded the experiment done in Fig 3 for (Li$_3$P + Li$_2$S) across other combinations in this section. Tables \ref{tab:m3gnet_joint_ft_sb} and \ref{tab:orb_joint_ft_sb} demonstrate that when restricted to identical computational constraints, \nameff significantly outperforms Joint FT-SB across both M3GNet and Orb architectures. This confirms that the success of \nameff is not an artifact of extended training, but a fundamental advantage of the merging topology.

\begin{table}[htbp]
    \centering
    \caption{Joint Fine-tuning on Same Budget (Joint FT-SB) MAE for M3GNet. We report Force MAE and Energy MAE.}
    \label{tab:m3gnet_joint_ft_sb}
    \scalebox{0.9}{
    \begin{tabular}{l|cc|cc}
    \toprule
    \multirow{2}{*}{\textbf{Scenario}} & \multicolumn{2}{c|}{\textbf{GFFMerge}} & \multicolumn{2}{c}{\textbf{Joint FT-SB}} \\
     & F & E & F & E \\
    \midrule
    \multicolumn{5}{c}{\cellcolor{green!10}\textbf{MD17}\textit{(kcal/mol/\AA{}, kcal/mol)}} \\
    Aspirin + Uracil & \textbf{0.94} & \textbf{0.03} & 1.54 & 0.06 \\
    Ethanol + Malonaldehyde + Aspirin & \textbf{0.90} & \textbf{0.04} & 1.18 & 0.05 \\
    5-Task Mix (MD17) & \textbf{0.85} & 0.03 & 0.97 & 0.03 \\
    \midrule
    \multicolumn{5}{c}{\cellcolor{yellow!10}\textbf{rMD17}\textit{(kcal/mol/\AA{}, kcal/mol)}} \\
    Aspirin + Uracil & \textbf{1.54} & \textbf{0.04} & 4.24 & 0.43 \\
    Ethanol + Malonaldehyde + Aspirin & \textbf{1.51} & \textbf{0.05} & 3.06 & 0.14 \\
    5-Task Mix (rMD17) & \textbf{1.58} & \textbf{0.06} & 1.97 & 0.07 \\
    \midrule
    \multicolumn{5}{c}{\cellcolor{blue!10}\textbf{MD22}\textit{(kcal/mol/\AA{}, kcal/mol)}} \\
    Ac-Ala3-NHMe + AT-AT & \textbf{1.65} & 0.05 & 2.02 & 0.05 \\
    Ac-Ala3-NHMe + DHA + Stachyose & \textbf{1.82} & \textbf{0.05} & 2.09 & 0.06 \\
    \midrule
    \multicolumn{5}{c}{\cellcolor{orange!10}\textbf{LiPS20}\textit{(eV/\AA{}, eV)}} \\
    Li$_2$S + Li$_3$P & \textbf{0.58} & \textbf{0.04} & 0.74 & 0.30 \\
    \bottomrule
    \end{tabular}
    }
\end{table}

\begin{table}[htbp]
    \centering
    \caption{Joint Fine-tuning on Same Budget (Joint FT-SB) MAE for Orb. Force MAE is reported.}
    \label{tab:orb_joint_ft_sb}
    \scalebox{0.9}{
    \begin{tabular}{l|c|c}
    \toprule
    \textbf{Scenario} & \textbf{GFFMerge F} & \textbf{Joint FT-SB F} \\
    \midrule
    \multicolumn{3}{c}{\cellcolor{green!10}\textbf{MD17}\textit{(kcal/mol/\AA{}, kcal/mol)}} \\
    Aspirin + Uracil & \textbf{1.175} & 1.785 \\
    Ethanol + Malonaldehyde + Aspirin & \textbf{1.564} & 2.160 \\
    5-Task Mix (MD17) & \textbf{1.343} & 1.508 \\
    \midrule
    \multicolumn{3}{c}{\cellcolor{yellow!10}\textbf{rMD17}\textit{(kcal/mol/\AA{}, kcal/mol)}} \\
    Aspirin + Uracil & \textbf{1.355} & 1.907 \\
    Ethanol + Malonaldehyde + Aspirin & \textbf{1.605} & 2.092 \\
    5-Task Mix (rMD17) & \textbf{1.363} & 1.569 \\
    \midrule
    \multicolumn{3}{c}{\cellcolor{blue!10}\textbf{MD22}\textit{(kcal/mol/\AA{}, kcal/mol)}} \\
    Ac-Ala3-NHMe + AT-AT & \textbf{1.357} & 1.892 \\
    Ac-Ala3-NHMe + DHA + Stachyose & \textbf{1.529} & 1.818 \\
    \midrule
    \multicolumn{3}{c}{\cellcolor{orange!10}\textbf{LiPS20}\textit{(eV/\AA{}, eV)}} \\
    Li$_2$S + Li$_3$P & \textbf{0.080} & 0.097 \\
    \bottomrule
    \end{tabular}
    }
\end{table}

\subsection{Root Mean Square Error (RMSE) and Coefficient of Determination ($R^2$)}
While Mean Absolute Error (MAE) serves as the primary evaluation metric, Root Mean Square Error (RMSE) and the Coefficient of Determination ($R^2$) provide complementary insights. RMSE inherently penalizes large prediction errors (outliers) more severely than MAE, which is critical for identifying potential instability points in high-energy states. The $R^2$ value measures the proportion of variance in the physical forces and energies explained by the model. It is a unitless metric; the higher it is, the better.The results in Tables \ref{tab:rmse_m3gnet} through \ref{tab:r2_orb} show that GFFMerge consistently matches or exceeds the statistical reliability of traditional Joint FT.

\begin{table}[htbp]
    \centering
    \caption{RMSE comparison across domains on M3GNet.}
    \label{tab:rmse_m3gnet}
    \resizebox{\textwidth}{!}{
    \begin{tabular}{l|cc|cc|cc|cc|cc|cc|cc}
    \toprule
    \multirow{2}{*}{\textbf{Scenario}} & \multicolumn{2}{c|}{\textbf{Wt Avg}} & \multicolumn{2}{c|}{\textbf{Fisher}} & \multicolumn{2}{c|}{\textbf{TIES}} & \multicolumn{2}{c|}{\textbf{EMR}} & \multicolumn{2}{c|}{\textbf{GFFMerge}} & \multicolumn{2}{c|}{\textbf{Joint FT SB}} & \multicolumn{2}{c}{\textbf{Joint FT}} \\
     & F & E & F & E & F & E & F & E & F & E & F & E & F & E \\
    \midrule
    \multicolumn{15}{c}{\cellcolor{green!10}\textbf{MD17}\textit{(kcal/mol/\AA{}, kcal/mol)}} \\
    Aspirin+Uracil & 2.38 & 1.13 & 2.03 & 0.83 & 3.76 & 4.02 & 2.81 & 1.80 & \textbf{1.32} & \textbf{0.04} & 2.11 & 0.08 & 1.23 & 0.04 \\
    Eth+Mal+Asp & 3.63 & 2.44 & 3.39 & 2.90 & 5.13 & 2.60 & 3.42 & 1.21 & \textbf{1.22} & \textbf{0.05} & 1.67 & 0.07 & 1.07 & 0.04 \\
    5-Task Mix & 3.40 & 2.44 & 3.13 & 2.52 & 3.54 & 2.75 & 3.66 & 3.78 & \textbf{1.18} & \textbf{0.04} & 1.40 & 0.04 & 0.99 & 0.03 \\
    \midrule
    \multicolumn{15}{c}{\cellcolor{yellow!10}\textbf{rMD17}\textit{(kcal/mol/\AA{}, kcal/mol)}} \\
    Aspirin+Uracil & 6.20 & 4.37 & 5.83 & 5.11 & 10.02 & 13.88 & 7.08 & 1.35 & \textbf{2.09} & \textbf{0.06} & 5.75 & 0.46 & 1.69 & 0.05 \\
    Eth+Mal+Asp & 7.60 & 10.61 & 8.00 & 10.55 & 11.61 & 12.15 & 9.41 & 2.37 & \textbf{2.05} & \textbf{0.07} & 4.16 & 0.17 & 1.41 & 0.05 \\
    5-Task Mix & 8.76 & 10.24 & 10.74 & 10.09 & 9.19 & 9.17 & 7.82 & 12.93 & \textbf{2.17} & \textbf{0.07} & 2.80 & 0.09 & 1.18 & 0.03 \\
    \midrule
    \multicolumn{15}{c}{\cellcolor{blue!10}\textbf{MD22}\textit{(kcal/mol/\AA{}, kcal/mol)}} \\
    Ac-Ala3+AT-AT & 4.13 & 1.97 & 4.65 & 2.51 & 7.47 & 5.03 & 4.80 & 1.67 & \textbf{2.32} & \textbf{0.06} & 2.85 & 0.07 & 1.84 & 0.04 \\
    Ac-Ala3+DHA & 4.96 & 2.46 & 4.96 & 2.58 & 6.55 & 1.07 & 6.36 & 7.35 & \textbf{2.54} & \textbf{0.06} & 2.98 & 0.07 & 1.98 & 0.06 \\
    \midrule
    \multicolumn{15}{c}{\cellcolor{orange!10}\textbf{LiPS20}\textit{(eV/\AA{}, eV)}} \\
    Li$_2$S + Li$_3$P & 11.78 & 62.31 & 68.47 & 53.50 & 5.15 & 191.56 & 78.78 & 37.29 & \textbf{0.77} & \textbf{0.05} & 1.05 & 0.39 & 0.45 & 0.09 \\
    \bottomrule
    \end{tabular}
    }
\end{table}

\begin{table}[htbp]
    \centering
    \caption{Force RMSE comparison across domains on Orb.}
    \label{tab:rmse_orb}
    \resizebox{\textwidth}{!}{
    \begin{tabular}{l|c|c|c|c|c|c|c}
    \toprule
    \textbf{Scenario} & \textbf{Wt Avg} & \textbf{Fisher} & \textbf{TIES} & \textbf{EMR} & \textbf{GFFMerge} & \textbf{Joint FT SB} & \textbf{Joint FT} \\
    \midrule
    \multicolumn{8}{c}{\cellcolor{green!10}\textbf{MD17}\textit{(kcal/mol/\AA{}, kcal/mol)}} \\
    Aspirin + Uracil & 7.898 & 10.979 & 8.492 & 11.907 & \textbf{1.635} & 2.426 & 1.339 \\
    Ethanol + Malonaldehyde + Aspirin & 12.436 & 12.121 & 10.164 & 20.231 & \textbf{2.157} & 2.942 & 1.602 \\
    5-Task Mix & 11.125 & 12.494 & 8.351 & 35.140 & \textbf{1.865} & 2.122 & 1.119 \\
    \midrule
    \multicolumn{8}{c}{\cellcolor{yellow!10}\textbf{rMD17}\textit{(kcal/mol/\AA{}, kcal/mol)}} \\
    Aspirin + Uracil & 7.978 & 12.448 & 8.620 & 15.647 & \textbf{1.855} & 2.573 & 1.408 \\
    Ethanol + Malonaldehyde + Aspirin & 13.983 & 12.173 & 11.006 & 18.915 & \textbf{2.193} & 2.884 & 1.717 \\
    5-Task Mix & 11.681 & 12.654 & 8.059 & 35.670 & \textbf{1.892} & 2.217 & 1.191 \\
    \midrule
    \multicolumn{8}{c}{\cellcolor{blue!10}\textbf{MD22}\textit{(kcal/mol/\AA{}, kcal/mol)}} \\
    Ac-Ala3-NHMe + AT-AT & 7.509 & 8.626 & 8.594 & 11.106 & \textbf{1.848} & 2.560 & 1.653 \\
    Ac-Ala3-NHMe + DHA + Stachyose & 8.105 & 10.375 & 8.164 & 25.475 & \textbf{2.120} & 2.506 & 1.503 \\
    \midrule
    \multicolumn{8}{c}{\cellcolor{orange!10}\textbf{LiPS20}\textit{(eV/\AA{}, eV)}} \\
    Li$_2$S + Li$_3$P & 0.425 & 0.444 & 0.438 & 0.431 & \textbf{0.133} & 0.188 & 0.151 \\
    \bottomrule
    \end{tabular}
    }
\end{table}

\begin{table}[htbp]
    \centering
    \caption{$R^2$ comparison across domains on M3GNet.}
    \label{tab:r2_m3gnet}
    \resizebox{\textwidth}{!}{
    \begin{tabular}{l|cc|cc|cc|cc|cc|cc|cc}
    \toprule
    \multirow{2}{*}{\textbf{Scenario}} & \multicolumn{2}{c|}{\textbf{Wt Avg}} & \multicolumn{2}{c|}{\textbf{Fisher}} & \multicolumn{2}{c|}{\textbf{TIES}} & \multicolumn{2}{c|}{\textbf{EMR}} & \multicolumn{2}{c|}{\textbf{GFFMerge}} & \multicolumn{2}{c|}{\textbf{Joint FT SB}} & \multicolumn{2}{c}{\textbf{Joint FT}} \\
     & F & E & F & E & F & E & F & E & F & E & F & E & F & E \\
    \midrule
    \multicolumn{15}{c}{\cellcolor{green!10}\textbf{MD17}\textit{(kcal/mol/\AA{}, kcal/mol)}} \\
    Aspirin+Uracil & 0.993 & 1.000 & 0.995 & 1.000 & 0.982 & 1.000 & 0.990 & 1.000 & \textbf{0.998} & 1.000 & 0.994 & 1.000 & 0.998 & 1.000 \\
    Eth+Mal+Asp & 0.981 & 1.000 & 0.984 & 1.000 & 0.963 & 1.000 & 0.983 & 1.000 & \textbf{0.997} & 1.000 & 0.996 & 1.000 & 0.998 & 1.000 \\
    5-Task Mix & 0.984 & 1.000 & 0.987 & 1.000 & 0.983 & 1.000 & 0.982 & 1.000 & \textbf{0.998} & 1.000 & 0.997 & 1.000 & 0.999 & 1.000 \\
    \midrule
    \multicolumn{15}{c}{\cellcolor{yellow!10}\textbf{rMD17}\textit{(kcal/mol/\AA{}, kcal/mol)}} \\
    Aspirin+Uracil & 0.956 & 1.000 & 0.961 & 1.000 & 0.885 & 1.000 & 0.943 & 1.000 & \textbf{0.995} & 1.000 & 0.962 & 1.000 & 0.997 & 1.000 \\
    Eth+Mal+Asp & 0.927 & 1.000 & 0.919 & 1.000 & 0.829 & 1.000 & 0.888 & 1.000 & \textbf{0.994} & 1.000 & 0.978 & 1.000 & 0.997 & 1.000 \\
    5-Task Mix & 0.909 & 1.000 & 0.863 & 1.000 & 0.900 & 1.000 & 0.928 & 1.000 & \textbf{0.994} & 1.000 & 0.991 & 1.000 & 0.998 & 1.000 \\
    \midrule
    \multicolumn{15}{c}{\cellcolor{blue!10}\textbf{MD22}\textit{(kcal/mol/\AA{}, kcal/mol)}} \\
    Ac-Ala3+AT-AT & 0.976 & 1.000 & 0.970 & 1.000 & 0.922 & 1.000 & 0.968 & 1.000 & \textbf{0.992} & 1.000 & 0.989 & 1.000 & 0.995 & 1.000 \\
    Ac-Ala3+DHA & 0.962 & 1.000 & 0.962 & 1.000 & 0.934 & 1.000 & 0.938 & 1.000 & \textbf{0.990} & 1.000 & 0.986 & 1.000 & 0.994 & 1.000 \\
    \midrule
    \multicolumn{15}{c}{\cellcolor{orange!10}\textbf{LiPS20}\textit{(eV/\AA{}, eV)}} \\
    Li$_2$S + Li$_3$P & -216.7 & -0.47 & -7421.7 & -0.02 & -41.1 & -11.2 & -9756.7 & 0.42 & \textbf{0.103} & 1.000 & -0.801 & 1.000 & 0.687 & 1.000 \\
    \bottomrule
    \end{tabular}
    }
\end{table}

\begin{table}[htbp]
    \centering
    \caption{Force $R^2$ comparison across domains on Orb.}
    \label{tab:r2_orb}
    \resizebox{\textwidth}{!}{
    \begin{tabular}{l|c|c|c|c|c|c|c}
    \toprule
    \textbf{Scenario} & \textbf{Wt Avg} & \textbf{Fisher} & \textbf{TIES} & \textbf{EMR} & \textbf{GFFMerge} & \textbf{Joint FT SB} & \textbf{Joint FT} \\
    \midrule
    \multicolumn{8}{c}{\cellcolor{green!10}\textbf{MD17}\textit{(kcal/mol/\AA{}, kcal/mol)}} \\
    Aspirin + Uracil & 0.923 & 0.851 & 0.911 & 0.824 & \textbf{0.997} & 0.993 & 0.998 \\
    Ethanol + Malonaldehyde + Aspirin & 0.800 & 0.810 & 0.867 & 0.472 & \textbf{0.994} & 0.989 & 0.997 \\
    5-Task Mix & 0.847 & 0.807 & 0.914 & -0.524 & \textbf{0.996} & 0.994 & 0.998 \\
    \midrule
    \multicolumn{8}{c}{\cellcolor{yellow!10}\textbf{rMD17}\textit{(kcal/mol/\AA{}, kcal/mol)}} \\
    Aspirin + Uracil & 0.932 & 0.835 & 0.921 & 0.740 & \textbf{0.996} & 0.993 & 0.998 \\
    Ethanol + Malonaldehyde + Aspirin & 0.779 & 0.833 & 0.863 & 0.596 & \textbf{0.995} & 0.991 & 0.997 \\
    5-Task Mix & 0.851 & 0.825 & 0.929 & -0.387 & \textbf{0.996} & 0.995 & 0.998 \\
    \midrule
    \multicolumn{8}{c}{\cellcolor{blue!10}\textbf{MD22}\textit{(kcal/mol/\AA{}, kcal/mol)}} \\
    Ac-Ala3-NHMe + AT-AT & 0.921 & 0.896 & 0.897 & 0.828 & \textbf{0.995} & 0.991 & 0.996 \\
    Ac-Ala3-NHMe + DHA + Stachyose & 0.902 & 0.839 & 0.900 & 0.027 & \textbf{0.993} & 0.991 & 0.997 \\
    \midrule
    \multicolumn{8}{c}{\cellcolor{orange!10}\textbf{LiPS20}\textit{(eV/\AA{}, eV)}} \\
    Li$_2$S + Li$_3$P & 0.722 & 0.697 & 0.706 & 0.716 & \textbf{0.973} & 0.946 & 0.965 \\
    \bottomrule
    \end{tabular}
    }
\end{table}

\subsection{Computational Efficiency and Timing Analysis}
In Tables \ref{tab:timing_m3gnet} and \ref{tab:timing_orb}, we report the total wall-clock time required to execute the merging process for all the baselines. All baselines are similarly efficient. However, GFFMerge is efficient while retaining highly competitive fidelity as shown in tables \ref{tab:M3GNet_results} and \ref{tab:Orb_results}.

\begin{table}[htbp]
    \centering
    \caption{Computational efficiency analysis on M3GNet. Times are reported in seconds.}
    \label{tab:timing_m3gnet}
    \resizebox{\textwidth}{!}{
    \begin{tabular}{l|c|c|c|c|c|c}
    \toprule
    \textbf{Scenario} & \textbf{Wt Avg (s)} & \textbf{Fisher (s)} & \textbf{TIES (s)} & \textbf{EMR (s)} & \textbf{GFFMerge (s)} & \textbf{Joint FT (s)} \\
    \midrule
    \multicolumn{7}{c}{\cellcolor{green!10}\textbf{MD17}} \\
    Aspirin + Uracil & 36.87 & 77.53 & 37.04 & 45.41 & 45.00 & 723.82 \\
    Ethanol + Malonaldehyde + Aspirin & 62.73 & 111.11 & 62.93 & 71.59 & 71.17 & 784.15 \\
    5-Task Mix & 211.10 & 277.83 & 211.37 & 220.29 & 226.74 & 1665.18 \\
    \midrule
    \multicolumn{7}{c}{\cellcolor{blue!10}\textbf{MD22}} \\
    Ac-Ala3-NHMe + AT-AT & 295.51 & 362.39 & 295.67 & 304.64 & 323.78 & 2397.11 \\
    Ac-Ala3-NHMe + DHA + Stachyose & 456.42 & 568.27 & 456.68 & 465.35 & 518.61 & 6412.42 \\
    \midrule
    \multicolumn{7}{c}{\cellcolor{orange!10}\textbf{LiPS20}} \\
    Li$_2$S + Li$_3$P & 609.07 & 678.57 & 609.28 & 618.10 & 658.13 & 3296.31 \\
    \bottomrule
    \end{tabular}
    }
\end{table}

\begin{table}[htbp]
    \centering
    \caption{Computational efficiency analysis on Orb. Times are reported in seconds.}
    \label{tab:timing_orb}
    \resizebox{\textwidth}{!}{
    \begin{tabular}{l|c|c|c|c|c|c}
    \toprule
    \textbf{Scenario} & \textbf{Wt Avg (s)} & \textbf{Fisher (s)} & \textbf{TIES (s)} & \textbf{EMR (s)} & \textbf{GFFMerge (s)} & \textbf{Joint FT (s)} \\
    \midrule
    \multicolumn{7}{c}{\cellcolor{green!10}\textbf{MD17}} \\
    Aspirin + Uracil & 146.89 & 207.10 & 284.38 & 194.18 & 154.21 & 1364.61 \\
    Ethanol + Malonaldehyde + Aspirin & 163.91 & 229.47 & 320.19 & 209.44 & 171.78 & 1354.46 \\
    5-Task Mix & 451.72 & 528.98 & 642.83 & 496.57 & 462.75 & 2893.44 \\
    \midrule
    \multicolumn{7}{c}{\cellcolor{yellow!10}\textbf{rMD17}} \\
    Aspirin + Uracil & 154.79 & 224.52 & 319.50 & 208.10 & 161.47 & 1468.70 \\
    Ethanol + Malonaldehyde + Aspirin & 180.20 & 246.36 & 335.19 & 226.13 & 188.08 & 1396.37 \\
    5-Task Mix & 444.09 & 528.47 & 644.88 & 491.15 & 455.13 & 2969.44 \\
    \midrule
    \multicolumn{7}{c}{\cellcolor{blue!10}\textbf{MD22}} \\
    Ac-Ala3-NHMe + AT-AT & 550.10 & 621.48 & 697.59 & 595.42 & 563.17 & 3526.78 \\
    Ac-Ala3-NHMe + DHA + Stachyose & 855.88 & 952.79 & 1031.15 & 900.65 & 872.48 & 6902.25 \\
    \midrule
    \multicolumn{7}{c}{\cellcolor{orange!10}\textbf{LiPS20}} \\
    Li$_2$S + Li$_3$P & 739.18 & 824.75 & 892.44 & 786.08 & 755.97 & 4257.98 \\
    \bottomrule
    \end{tabular}
    }
\end{table}

\FloatBarrier
\subsection{Extended Out-of-Distribution (rMD17) Results}
To ensure robustness to data noise and harder generalization scenarios, we present extended predictive MAE performance on the revised MD17 (rMD17) dataset. Unlike MD17, rMD17 utilizes highly accurate coupled-cluster (CCSD(T)) levels of theory for labels, presenting a stricter learning landscape. Tables \ref{tab:rmd17_mae_m3gnet} thorugh \ref{tab:timing_rmd17_orb} confirm that \nameff still vastly outperforms conventional baselines across this refined dataset.

\begin{table}[htbp]
    \centering
    \caption{rMD17 MAE on M3GNet. Force MAE (F) is reported in kcal/mol/\AA{} and Energy MAE (E) is reported in kcal/mol.}
    \label{tab:rmd17_mae_m3gnet}
    \scalebox{0.75}{
    \begin{tabular}{l|cc|cc|cc|cc|cc|cc}
    \toprule
    \multirow{2}{*}{\textbf{Dataset}} & \multicolumn{2}{c|}{\textbf{Wt Avg}} & \multicolumn{2}{c|}{\textbf{Fisher}} & \multicolumn{2}{c|}{\textbf{TIES}} & \multicolumn{2}{c|}{\textbf{EMR}} & \multicolumn{2}{c|}{\textbf{GFFMerge}} & \multicolumn{2}{c}{\textbf{Joint FT}} \\
     & F & E & F & E & F & E & F & E & F & E & F & E \\
    \midrule
    \multicolumn{13}{c}{\cellcolor{yellow!10}\textbf{rMD17}} \\
    Aspirin + Uracil & 4.39 & 4.35 & 4.14 & 5.10 & 7.42 & 13.85 & 4.95 & 1.13 & \textbf{1.54} & \textbf{0.04} & 1.22 & 0.03 \\
    Eth + Mal + Asp & 5.58 & 10.16 & 5.82 & 10.20 & 9.00 & 11.79 & 6.54 & 2.26 & \textbf{1.51} & \textbf{0.05} & 1.00 & 0.04 \\
    5-Task Mix & 5.89 & 9.72 & 7.07 & 9.83 & 6.95 & 8.63 & 5.61 & 9.66 & \textbf{1.58} & \textbf{0.06} & 0.83 & 0.03 \\
    \bottomrule
    \end{tabular}
    }
\end{table}

\begin{table}[htbp]
    \centering
    \caption{rMD17 MAE on Orb. Force MAE is reported in kcal/mol/\AA{}.}
    \label{tab:rmd17_mae_orb}
    \scalebox{0.75}{
    \begin{tabular}{l|c|c|c|c|c|c}
    \toprule
    \textbf{Dataset} & \textbf{Wt Avg} & \textbf{Fisher} & \textbf{TIES} & \textbf{EMR} & \textbf{GFFMerge} & \textbf{Joint FT} \\
    \midrule
    \multicolumn{7}{c}{\cellcolor{yellow!10}\textbf{rMD17}} \\
    Aspirin + Uracil & 5.919 & 8.821 & 6.347 & 11.553 & \textbf{1.355} & 1.013 \\
    Eth + Mal + Asp & 9.680 & 8.598 & 7.662 & 13.626 & \textbf{1.605} & 1.243 \\
    5-Task Mix & 7.820 & 8.668 & 5.889 & 26.992 & \textbf{1.363} & 0.846 \\
    \bottomrule
    \end{tabular}
    }
\end{table}

\begin{table}[!htbp]
    \centering
    \caption{Comparison of total training time and speedup between Joint FT and GFFMerge for rMD17 using M3GNet.}
    \label{tab:timing_rmd17_m3gnet}
    \scalebox{0.75}{
    \begin{tabular}{l|c|c|c}
    \toprule
    \textbf{Scenario} & \textbf{Joint FT (s)} & \textbf{GFFMerge (s)} & \textbf{Speedup} \\
    \midrule
    \multicolumn{4}{c}{\cellcolor{yellow!10}\textbf{rMD17}} \\
    Aspirin + Uracil & 732.08 & 44.66 & 16.39x \\
    Ethanol + Malonaldehyde + Aspirin & 821.09 & 79.41 & 10.34x \\
    5-Task Mix & 1650.15 & 216.55 & 7.62x \\
    \bottomrule
    \end{tabular}}
\end{table}

\begin{table}[!h]
    \centering
    \caption{Comparison of total training time and speedup between Joint FT and GFFMerge for rMD17 using Orb.}
    \label{tab:timing_rmd17_orb}
    \scalebox{0.75}{
    \begin{tabular}{l|c|c|c}
    \toprule
    \textbf{Scenario} & \textbf{Joint FT (s)} & \textbf{GFFMerge (s)} & \textbf{Speedup} \\
    \midrule
    \multicolumn{4}{c}{\cellcolor{yellow!10}\textbf{rMD17}} \\
    Aspirin + Uracil & 1468.70 & 161.47 & 9.10x \\
    Ethanol + Malonaldehyde + Aspirin & 1396.37 & 188.08 & 7.42x \\
    5-Task Mix & 2969.44 & 455.13 & 6.52x \\
    \bottomrule
    \end{tabular}}
\end{table}
\end{review}

\end{document}